\documentclass{article} 

\usepackage{amsmath,amsfonts,bm}

\def\eqref#1{equation~\ref{#1}}

\def\1{\bm{1}}

\DeclareMathAlphabet{\mathsfit}{\encodingdefault}{\sfdefault}{m}{sl}
\SetMathAlphabet{\mathsfit}{bold}{\encodingdefault}{\sfdefault}{bx}{n}

\def\gA{{\mathcal{A}}}

\def\gD{{\mathcal{D}}}

\def\gL{{\mathcal{L}}}

\def\gP{{\mathcal{P}}}

\def\gS{{\mathcal{S}}}
\def\gT{{\mathcal{T}}}

\def\gX{{\mathcal{X}}}
\def\gY{{\mathcal{Y}}}

\def\sP{{\mathbb{P}}}

\def\sR{{\mathbb{R}}}

\newcommand{\E}{\mathbb{E}}

\usepackage{hyperref}
\usepackage{url}
\usepackage{booktabs,multirow,threeparttable,siunitx,tabularx}
\usepackage{colortbl}
\usepackage[dvipsnames,table]{xcolor}
\usepackage{array}
\usepackage{algorithm}
\usepackage{algpseudocode}
\usepackage{graphicx}
\usepackage{caption}
\usepackage{subcaption}
\definecolor{pipatint}{gray}{0.94}
\usepackage{threeparttable}
\usepackage{titlesec}
\titlespacing*{\paragraph}{0pt}{0.5ex}{0.5em}
\usepackage{natbib}

\newcommand{\cX}{\mathcal X}
\newcommand{\Dtr}{\mathcal D_{\mathrm{tr}}}
\newcommand{\Dval}{\mathcal D_{\mathrm{val}}}
\newcommand{\Emp}{\widehat{\mathbb E}}
\newcommand{\sg}{\operatorname{sg}}
\newcommand{\tr}{\operatorname{tr}}
\newcommand{\Tp}{\mathcal T^{+,\pi}}

\usepackage{amsthm}

\newtheorem{theorem}{Theorem}[section]

\newtheorem{lemma}[theorem]{Lemma}
\newtheorem{corollary}[theorem]{Corollary}

\theoremstyle{definition}

\theoremstyle{remark}
\newtheorem*{remark}{Remark}

\title{SUAF: Spectral Universal Advantage Features for Zero-Shot Policy Improvement}

\author{%
  Huaze Tang\textsuperscript{1,}\thanks{Huaze Tang and Runqing Miao contributed equally.},\quad
  Runqing Miao\textsuperscript{1,\(\ast\)},\quad
  Tianren Peng\textsuperscript{1},\quad
  Lam Nguyen\textsuperscript{1},\\
  Chao Wang\textsuperscript{2},\quad
  Jairong Li\textsuperscript{3,}\thanks{Jairong Li and Wenbo Ding are the corresponding authors.},\quad
  Wenbo Ding\textsuperscript{1,\(\dagger\)}\\[0.5em]
  {\small \textsuperscript{1}Tsinghua Shenzhen International Graduate School,
    Tsinghua University}\\
  {\small \textsuperscript{2}Tencent}\\
  {\small \textsuperscript{3}Peng Cheng Laboratory}\\[0.3em]
  {\small \texttt{\{tanghz24,miaorq26,rql25\}@mails.tsinghua.edu.cn}}\\
  {\small \texttt{bobchaowang@tencent.com, lijr02@pcl.ac.cn}}\\
  {\small \texttt{ding.wenbo@sz.tsinghua.edu.cn}}%
}

\begin{document}

\maketitle

\begin{abstract}
Reward-independent representations enable reinforcement learning agents to reuse knowledge across tasks, but limited representation capacity raises three fundamental questions: what statistical dependence should these representations capture, which components of that dependence should be retained for policy improvement, and how does discarding the remaining components affect decision-making? We address these questions through canonical dependence analysis. For a reference policy, we construct an action-centered, positive-lag canonical dependence matrix (CDM) that isolates the additional predictive contribution of the current action beyond the current state. Its singular decomposition yields Spectral Universal Advantage Features (SUAF), paired features that recover delayed advantage through linear reward projections. {\color{black}We show that the spectrum quantifies mutual information between current action and future state–action visitation conditioned on current state and determines reward-specific approximation errors.} Retaining the leading modes gives optimal rank-constrained approximations under both isotropic average-case and reward-uniform minimax criteria. Separating the immediate reward contrast also avoids the spectral floor arising from direct compression of the full advantage operator. To learn SUAF from reward-free data, we develop an action-centered Soft-HGR objective with bidirectional temporal-difference estimation. We further connect spectral approximation errors to sufficient conditions for policy improvement. Finite-state experiments validate the spectral predictions and feature-learning method, while zero-shot evaluations on discrete tasks and continuous-control benchmarks demonstrate the utility of the learned representations across rewards.
\end{abstract}

\section{Introduction}

Recent advances in large language models (LLMs) and embodied artificial intelligence (AI) have triggered a growing ambition to build general-purpose agents that operate in open-ended environments \cite{xi2025rise,liu2025aligning}. Unlike traditional task-specific models, these agents are expected to handle a diverse array of goals, costs, and preferences within the same environment, often specified on the fly by human instructions or changing task contexts. Although these changes alter the reward function, knowledge of how actions influence future outcomes remains reusable. Reward-independent representations seek to capture this knowledge and support new tasks without learning a separate critic from scratch.  

Among reward-independent representations, successor features (SFs) provide a foundational framework for such reuse by separating reward weights from the discounted accumulation of features \cite{dayan1993improving,barreto2017successor}. Building on this, more recent successor-measure (SM) representations seek to encode behavior across policies using reward-free interaction data \cite{blier2021learning,touati2021learning,touati2022does}. However, a critical challenge arises: which features should be learned? To address this problem, \citet{ollivier2025featuresbestsuccessorfeatures} connects optimal features to the spectral structure of the full advantage operator through expected downstream performance under strong behavior-cloning regularization. Despite these theoretical advances, limited representation capacity still raises three fundamental questions: what statistical dependence should these representations capture, which components of that dependence should be retained for policy improvement, and how does discarding the remaining components affect decision-making?

To address these problems, we introduce \textbf{Spectral Universal Advantage Features} (SUAF), a reward-independent representation of delayed advantage with respect to the corresponding behavior policy. Our starting point is that future outcomes depend on both state and action, while advantage measures the action's contribution relative to a reference policy. We therefore isolate the dependence that remains after subtracting the policy-averaged future {\color{black}state--action} visitation. Building on canonical dependence analysis \cite{huang2024universal}, we construct an action-centered, positive-lag canonical dependence matrix (CDM) {\color{black} between current state--action pair and future visitation, where action centering is conducted by subtracting the policy-averaged visitation and positive lag excludes the current time step from the visitation count.}
The singular value decomposition of this matrix yields paired features of current and future state--action pairs{\color{black}, which defines the proposed SUAF. The full expansion {\color{black} of SUAF} recovers the advantage function by projecting a downstream reward onto the future features. {\color{black}We further show that the immediate reward contrast imposes a strictly positive lower bound on the worst-case error of directly approximating the full advantage operator at low rank. We call this lower bound the spectral floor.} When the immediate reward contrast can be evaluated directly, SUAF {\color{black}avoids this floor by} keeping it explicit and compressing only the delayed component. For this delayed component, {\color{black}SUAF} gives an optimal rank-constrained approximation under both average-case and worst-case reward objectives. Under weak dependence, we further relate the sum of squared singular values to the conditional mutual information between the current action and the future state--action pair given the current state.}
We then ask how discarding the remaining {\color{black} modes} affects decision-making {\color{black} and} derive an exact approximation-error expression via projections onto discarded modes, distinguishing {\color{black}the total dependence carried by the discarded modes from their contribution to the advantages for a given reward}, and obtain a sufficient condition under which the predicted policy-improvement gain exceeds the approximation {\color{black}{loss}}. We further bound local representation reuse under policy changes within trust region. To learn SUAF without constructing the CDM, we develop an action-centered Soft-HGR objective \cite{wang2019efficient} with bidirectional temporal-difference estimation, using forward and backward predictors from reward-free transitions. The learned representation supports advantage readout and policy extraction. Experiments examine the spectral floor, truncation errors, and feature-learning accuracy, with zero-shot evaluations on discrete and continuous-control benchmarks.

In summary, our contributions are threefold:
\begin{enumerate}
\item We introduce SUAF through an action-centered positive-lag CDM, connecting delayed advantage to the dependence between current actions and future outcomes. We identify a spectral floor in the full advantage operator that motivates retaining the immediate reward contrast explicitly and compressing only the delayed component.
\item We establish average-case and reward-uniform minimax optimality for spectral approximation of delayed advantage. We derive reward-specific truncation errors that yield sufficient conditions for policy improvement.
\item We develop an action-centered Soft-HGR learning method with bidirectional TD estimation. Experiments validate the spectral analysis and feature-learning {\color{black}performance}, and evaluate zero-shot policy improvement across rewards in discrete and continuous environments.
\end{enumerate}

\section{Preliminaries}


\subsection{Universal feature functions for high dimensional learning}
Let $X\in\gX$ and $Y\in\gY$ be two random variables with joint
distribution $P_{XY}$ and marginal distributions $P_X$ and $P_Y$.
Assuming that $P_X(x)>0$ and $P_Y(y)>0$ on their supports, the
canonical dependence matrix (CDM)
$\widetilde{\boldsymbol B}\in
\mathbb{R}^{|\gY|\times|\gX|}$ is defined as
\begin{equation*}
    \widetilde{\boldsymbol B}(y,x)
    =
    \frac{
        P_{XY}(x,y)-P_X(x)P_Y(y)
    }{
        \sqrt{P_X(x)P_Y(y)}
    }.
\end{equation*}
The CDM is the normalized residual between the joint distribution
$P_{XY}$ and the independent distribution $P_XP_Y$. In particular,
$\widetilde{\boldsymbol B}=\boldsymbol 0$ if and only if $X$ and $Y$
are independent. When $X$ and $Y$ are weakly dependent, $\|\widetilde{\boldsymbol B}\|_{\mathrm F}^2 \approx 2I\!\left(X;Y\right)$, where $I(X;Y)$ is mutual information. Let the singular value decomposition (SVD) of the CDM be $\widetilde{\boldsymbol B}= \sum_{i=1}^{R} \sigma_i \psi_i^Y \left(\psi_i^X\right)^\text{T}$ with $\sigma_1\geq \sigma_2\geq\cdots\geq\sigma_R>0$, 
where
$R\leq\min\{|\gX|-1,|\gY|-1\}$.
The corresponding feature functions are defined as
\begin{equation*}
    f_i(x)
    =
    \frac{\psi_i^X(x)}{\sqrt{P_X(x)}},
    \qquad
    g_i(y)
    =
    \frac{\psi_i^Y(y)}{\sqrt{P_Y(y)}}.
\end{equation*}
These feature functions are centered and orthonormal under their respective marginal distributions.
For any $k\leq R$, the first $k$ pairs of $\{f,g\}$ solve the
generalized canonical correlation problem
\begin{equation}
\begin{aligned}
    \max_{\{f_i,g_i\}_{i=1}^{k}}
    \quad&
    \sum_{i=1}^{k}
    \E[f_i(X)g_i(Y)]\\
    \mathrm{s.t.}\quad&
    \E[f_i(X)]
    =
    \E[g_i(Y)]
    =
    0,\\
    &\E[f_i(X)f_j(X)]
    =
    \E[g_i(Y)g_j(Y)]
    =
    \bm{1}(i=j),
\end{aligned}
\end{equation}
whose optimal value is $\sum_{i=1}^{k}\sigma_i$. The case $k=1$ reduces to the Hirschfeld-Gebelein-R\'enyi (HGR) maximal correlation. Under the local information-geometric formulation of \cite{huang2024universal}, these leading feature functions are universal in the
sense that they optimize the average local inference performance for unknown weak attributes of $Y$.

\subsection{Markov Decision Processes}

An MDP is defined as a tuple $\langle \gS, \gA, T, r, \gamma, \mu\rangle$, where $\gS$ denotes the state space, $\gA$ denotes the action space, $T:\gS\times\gA\to \gP(\gS)$ denotes the transition probability with $\gP(\cdot)$ denotes a probability distribution over a set, $r:\gS\times\gA\to\sR$ denotes the reward function with $|r(s,a)|\leq R_{\max}$ for all $(s,a)$, $\gamma\in[0,1)$ denotes the discount factor and $\mu$ denotes the distribution of initial states. The goal of RL is to learn a policy $\pi_\theta:\gS\to\gP(\gA)$ with parameters $\theta$ that maximize the expected return $J_r(\pi_\theta) = \E_{s_0\sim\mu}\E_{\pi_\theta}\left[\sum_{t=0}^\infty \gamma^t r(s_t,a_t)\right]$ where $s_t$ and $a_t$ denotes the state and action at timestep $t$.

The marginal occupancy measure, $d^\pi(s,a)$ is defined as the normalized probability of visiting state-action pair $(s,a)$ following the policy $\pi$ from a state sampled from the initial state distribution, which is $d^\pi(s,a) = (1-\gamma) \sum_{t=0}^\infty \gamma^t \sP_\pi(s_t=s,a_t=a)$. The successor measure, $M^\pi(s,a,s^+,a^+)$, is defined as the probability of being in state $s^+$ and taking action $a^+$ when starting from the state-action pair $(s,a)$ and following the policy $\pi$, which is $M^\pi(s^+,a^+|s,a) = (1-\gamma)\sum_{t=0}^\infty \gamma^t \sP_\pi(s_t=s^+,a_t=a^+ | s_0=s,a_0=a)$. The marginal occupancy measure is the expectation of successor measure, namely, $d^\pi(s,a) = \E_{s_0\sim\mu,a_0\sim \pi(\cdot|s_0)}[M^\pi(s,a|s_0,a_0)]$. 
The $Q$-function defined using successor measure is $Q^\pi(s,a) = \E_{(s^+,a^+)\sim M^\pi}[r(s^+,a^+)|s,a]$ or $Q^\pi=(1-\gamma)^{-1}M^\pi r$. The $V$-function is defined as $V^\pi(s) = \sum_{a} \pi(a|s) Q^\pi(s,a)$ and advantage function is defined as $A^\pi(s,a) = Q^\pi(s,a) - V^\pi(s)$.

\section{Spectral Universal Advantage Feature Functions}
\label{sec:universal_adv_feature}

{\color{black} This section introduces SUAF and establishes its theoretical foundations of universality. We first define SUAF by dependent CDM and derive an exact spectral representation of the advantage function. We then establish its universality through average-case and minimax optimality under a rank constraint. Finally, we show that directly compressing the full advantage introduces a spectral floor due to the immediate reward contrast, providing the theoretical justification for retaining immediate reward explicitly and compressing only the delayed component.}

\paragraph{Definition and Properties of SUAF} We establish the SUAF with dependent CDM. To begin with, we define the positive-lag successor measure $M^{+,\pi}(s,a,s^+,a^+)$ as
\begin{equation*}
    M^{+,\pi}(s^+,a^+|s,a) = (1-\gamma) \sum_{t=1}^\infty \gamma^{t-1} \sP_\pi(s_t=s^+,a_t=a^+|s_0=s,a_0=a).
\end{equation*}
This measure represents the normalized discounted distribution of future state–action pairs $(s^+,a^+)$ reached under policy $\pi$ from $(s,a)$, excluding the current time step. Unlike the marginal occupancy measure $d^\pi$, it remains dependent on the initial state–action pair and therefore characterizes how the current action shapes the strictly future visitation distribution. Based on the positive-lag successor measure, we consider the positive-lag dependent CDM (DCDM)\footnote{We remove the $t=0$ term to control the spectrum. See details in Theorem \ref{thm:removing_spectral_floor} and Appendix \ref{sec:immediate_spectral_floor}.} as
\begin{equation}
\begin{aligned}
&\widetilde{\boldsymbol B}^{+,\pi}((s,a),(s^+,a^+))\\
&\quad= \sqrt{d^\pi(s,a)}\frac{M^{+,\pi}(s^+,a^+|s,a) - \E_{a^\prime\sim\pi(\cdot|s)}\left[M^{+,\pi}(s^+,a^+|s,a^\prime)\right]}{\sqrt{\nu^{+,\pi}(s^+,a^+)}},
\end{aligned}
\end{equation}
where $\nu^{+,\pi}(\cdot)=\sum_{s,a}d^\pi(s,a)
M^{+,\pi}(\cdot\mid s,a)$ is the marginal distribution of $(s^+,a^+)$.
The DCDM is the normalized residual between successor measure and the corresponding distribution conditioned on the state alone. In particular, conditioning on $s$ removes the predictive information already contained in the current state, so that $\widetilde{\boldsymbol B}^{+,\pi}$ captures the additional influence of the current action $a$ on the strictly future state-action distribution.
Especially, when $(S^+,A^+)$ is weakly dependent on $(S,A)$, $\|
\widetilde{\boldsymbol B}^{+,\pi}
\|_{\mathrm F}^2
\approx 2I(A;S^+,A^+|S)$ holds\footnote{$I(A;S^+,A^+|S)$ denotes the conditional mutual information. See in Appendix \ref{sec:preperties_dcdm} for details.}. The weak dependence happens when the effective discount horizon $1/(1-\gamma)$ is substantially longer than the mixing time of the Markov chain induced by $\pi$. A sufficiently exploratory policy and overlapping stochastic transitions will further weaken the dependence. With the SVD of DCDM as 
$$
\widetilde{\boldsymbol B}^{+,\pi} = \sum_{i=1}^R \sigma_i^{+,\pi} \psi_{i,+,\pi}^{S,A}( \psi_{i,+,\pi}^{S^+,A^+})^\text{T}
$$
with $\sigma_1^{+,\pi} \geq \dots \geq \sigma_R^{+,\pi} > 0$, where $R \leq |\gS| - 1$ indicating a low-rank matrix\footnote{The rank is less than the state-action joint space size. See proof in Appendix \ref{sec:preperties_dcdm}.}, we define the SUAF as
\begin{equation}
    f_i(s,a)
    =
    \frac{\psi_{i,+,\pi}^{S,A}(s,a)}{\sqrt{d^\pi(s,a)}},
    \quad
    g_i(s^+,a^+)
    =
    \frac{\psi_{i,+,\pi}^{S^+,A^+}(s^+,a^+)}{\sqrt{\nu^{+,\pi}(s^+,a^+)}}.
\end{equation}
The proposed SUAF shares the properties with the universal features proposed by \citet{huang2024universal}, and see Appendix \ref{sec:preperties_suaf} for details.

\paragraph{Reward-Independent Universal Optimality of SUAF} In this paragraph, we propose that the SUAF is universal for downstream tasks. Furthermore, when the advantage operator is approximated with rank at most $k$, we prove that SUAF is min-max optimal for downstream tasks.
\begin{theorem}[Universal advantage function representation]
\label{thm:advantage-representation}
For any reward function \(r:\mathcal S\times\mathcal A\to\mathbb R\), and for every \((s,a)\) in the support of \(d^\pi\), the advantage function admits the exact representation
\begin{equation}
\label{eq:advantage-spectral-representation}
A_r^\pi(s,a)
=
C_\pi r(s,a)
+
\frac{\gamma}{1-\gamma}
\sum_{i=1}^{R} \sigma_i^{+,\pi} f_i(s,a) \left\langle g_i,r\right\rangle_{\nu^{+,\pi}},
\end{equation}
where $C_\pi r(s,a):=r(s,a)-\mathbb E_{a\sim\pi(\cdot\mid s)}\left[r(s,a)\right]$ is the immediate centered reward and $\left\langle g_i,r\right\rangle_{\nu^{+,\pi}}\!=\!\sum_{s,a}\!\nu^{+,\pi}(s,a)g_i(s,a)r(s,a)$ is the inner product on Hilbert space $L^2(\nu^{+,\pi})$.
\end{theorem}
Denote the positive-lag future operator as 
\begin{equation*}
\begin{aligned}
    & \gT^{+,\pi}r(s,a)\\
    &\quad = \sum_{s^+,a^+}\left(
M^{+,\pi}(s^+,a^+\mid s,a) - \mathbb E_{a'\sim\pi(\cdot\mid s)}
\left[ M^{+,\pi}(s^+,a^+\mid s,a') \right]
\right)
r(s^+,a^+),
\end{aligned}
\end{equation*}
and the advantage function can be written as $A_r^{\pi} = C_\pi r + \frac{\gamma}{1-\gamma} \gT^{+,\pi}r$ with $\gT^{+,\pi}=\sum_i \sigma_{i}^{+,\pi}f_i\otimes g_i$.
Consequently, the DCDM provides a reward-independent representation of the delayed component of the advantage function. The functions $\{g_i\}_{i=1}^R$ form an orthonormal basis for the reward subspace relevant to the delayed advantage. In particular, every reward can be decomposed as $r=\sum_{i=1}^R \langle g_i,r\rangle_{\nu^{+,\pi}}g_i + r_\perp$, where the residual $r_\perp$ satisfies
$
\gT^{+,\pi}r_\perp(s,a)=0
$
for every $(s,a)$. Hence, $r_\perp$ has no effect on the delayed advantage, and the coefficients
$\{\langle g_i,r\rangle_{\nu^{+,\pi}}\}_{i=1}^R$
completely characterize the reward information relevant to future credit assignment. 

When only $k<R$ features can be used, the approximation is denoted as $\widehat{\gT}^{+,\pi}$ with $\text{rank}(\widehat{\gT}^{+,\pi})\leq k$. The corresponding advantage function is $\hat{A}_r^\pi=C_\pi r + \frac{\gamma}{1-\gamma}\widehat{\gT}^{+,\pi}r$. Theorem~\ref{thm:average_case_optimality} then shows that, conditional on the environment and policy, SUAF provides universal, reward-independent advantage features with optimal rank-\(k\) approximation risk across downstream tasks.
\begin{theorem}[Average-case optimality of the rank-$k$ representations]
\label{thm:average_case_optimality}
Let reward function $r$ be sampled from uniform distribution over the unit sphere of $L^2(\nu^{+,\pi})$, and let $m=|\operatorname{supp}(\nu^{+,\pi})|$. Among all reward-independent rank-$k$ approximations $\widehat{\gT}^{+,\pi}$
for $\hat{A}_r^\pi$, the top-$k$ truncated representation
\begin{equation}
\label{eq:truncated_advantage_function}
{A}_{r,k}^\pi
=
C_\pi r
+
\frac{\gamma}{1-\gamma}
\sum_{i=1}^k
\sigma_i^{+,\pi}f_i
\langle g_i,r\rangle_{\nu^{+,\pi}}
\end{equation}
minimizes the expected advantage approximation error, i.e.,
$$
\mathbb E_r
\left[
\left\|A_r^\pi-{A}_{r,k}^\pi\right\|_{d^\pi}^2
\right]
= \!\!\!\!\!\!\inf_{\text{rank}(\widehat{\gT}^{+,\pi})\leq k}
\!\!\!\!\mathbb E_r
\left[
\left\|A_r^\pi-\hat{A}_{r}^\pi\right\|_{d^\pi}^2
\right]
= \frac{1}{m}
\left(\frac{\gamma}{1-\gamma}\right)^2
\sum_{i>k}\left(\sigma_i^{+,\pi}\right)^2,
$$
and top-k truncation is also optimal under minimax case, i.e.,
\begin{equation*}
    \sup_{\| r \|_{\nu^{+,\pi}}\leq 1}\!\!\left\|A_r^\pi-{A}_{r,k}^\pi\right\|_{d^\pi}^2 
    = \!\!\!\!\!\!\inf_{\text{rank}(\widehat{\gT}^{+,\pi})\leq k}\sup_{\| r \|_{\nu^{+,\pi}}\leq 1}\!\! \left\|A_r^\pi-\hat{A}_{r}^\pi\right\|_{d^\pi}^2
    \!\!= \!\!\left(\frac{\gamma}{1-\gamma}\right)^2 \!\!\left(\sigma^{+,\pi}_{k+1}\right)^2.
\end{equation*}
\end{theorem}
The uniform distribution over the unit sphere of $L^2(\nu^{+,\pi})$ serves as a task-agnostic reward model, with each reward defining a downstream task. 

Furthermore, we claim that the exact immediate term also removes a spectral floor that arises when approximating the full advantage operator.
\begin{theorem}[Removing the immediate-term spectral floor]
\label{thm:removing_spectral_floor}
Consider a finite MDP with $0<\gamma<1$ and strictly positive
$d^\pi$ and $\nu^{+,\pi}$.
Let
$
\alpha_\pi
=
\min_s\frac{d_S^\pi(s)}{\nu_S^{+,\pi}(s)},
q=|\mathcal S|(|\mathcal A|-1).
$
For every $0\le k<q$,
\begin{align}
&\inf_{\operatorname{rank}(K)\le k}
\sup_{\|r\|_{\nu^{+,\pi}}\le1}
\|A_r^\pi-Kr\|_{d^\pi}^2
\ge
\alpha_\pi+
\left(\frac{\gamma}{1-\gamma}\right)^2
\left(\sigma_{k+1}^{+,\pi}\right)^2,
\label{eq:full_advantage_floor}
\end{align}
where singular values beyond the rank of
$\gT^{+,\pi}$ are defined to be zero.
In contrast, retaining $C_\pi r$ exactly and truncating
only the future term incurs worst-case error without $\alpha_\pi$ as shown in Theorem \ref{thm:average_case_optimality}.
If $\nu^{+,\pi}=d^\pi$, then $\alpha_\pi=1$
and \eqref{eq:full_advantage_floor} holds with equality.
\end{theorem}

Therefore, directly compressing the full advantage operator retains a nonzero error floor for $k<q$, whereas the SUAF approximation error follows only the delayed spectrum and vanishes once all nonzero future modes are retained. The improvement comes from computing the known immediate term exactly, rather than spending the learned rank budget on representing it. For per-reward analysis, please refer to Appendix \ref{sec:reward_shaping}.

\section{Methods}
\label{sec:algo}


In this section, we use the truncated SUAF to achieve policy improvement. We propose two classes of improvements, including one-step update and trust region update. After that, we introduce methods for estimating SUAF.

\paragraph{One Step Policy Update} 
Following One-step FB \cite{zheng2026can}, we use the SUAF of a fixed behavior policy $\pi$ to perform reward-conditioned policy improvement. Given a downstream reward $r$, we compute its coefficients $\beta_i(r)=\langle g_i^\pi,r\rangle_{\nu^{+,\pi}}$
and construct $A_{r,k}^{\pi}(s,a)$ following \eqref{eq:truncated_advantage_function}.
The improved policy selects actions maximizing this score:
$
\pi_r^+(s)\in \arg\max_a A_{r,k}^{\pi}(s,a).
$
For continuous actions, this maximization can be approximated by training an actor (which will be frozen when deploy) while keeping the SUAF fixed.
A soft variant can replace the argmax with a Boltzmann update as $\pi_r^+(a|s) \propto \pi(a|s)\exp(\tau A_{r,k}^\pi(s,a))$. Equivalently, $\pi_r^+$ solves the KL-regularized policy improvement problem $\max_{\pi^\prime}\E_{\pi^\prime}[A_{r,k}^\pi(s,a)] - \tau^{-1}\gD_{\text{KL}}(\pi^\prime\|\pi)$. As $\tau\to\infty$, this recovers the greedy update and as $\tau\to 0$, it recovers $\pi$.
For state-only rewards, $C_\pi r=0$, so an actor conditioned on $\beta(r)$ can be pretrained across coefficient vectors and reused for new rewards without further parameter updates. With exact full-rank SUAF and exact maximization, $\pi_r^+$ is greedy with respect to $Q_r^\pi$ and hence satisfies
$V_r^{\pi_r^+}\ge V_r^\pi$.
This provides one policy-improvement step. Note that no training steps when deploy to downstream tasks and hence it forms a zero-shot deployment.
\begin{theorem}[Boltzmann policy improvement with truncated SUAF]
\label{thm:suaf_policy_improvement_one_step}
Let $A_{r,k}^{\pi}$ be the exact rank-$k$ SUAF advantage
approximation, and let $\pi_\tau(a| s) \propto \pi(a| s)e^{\tau A_{r,k}^{\pi}(s,a)}$ with $\tau>0$. Define $B_k:=\sup_{s,a,b}|A_{r,k}^{\pi}(s,a)-A_{r,k}^{\pi}(s,b)|$, $\varepsilon_k(r)=\frac{\gamma}{1-\gamma} \left[\sum_{i>k} (\sigma_i^{+,\pi})^2\beta_i(r)^2\right]^{1/2}$ and $\Delta_k(r):=\|A_{r,k}^{\pi}\|_{d^\pi}^2-\|C^\pi r\|_{d^\pi}\varepsilon_k(r)$
If $\Delta_k(r)>0$, then it guarantees $J_r(\pi_\tau)>J_r(\pi)$ if temperature satisfies
$
0<\tau<
\frac{4(1-\gamma)^2\Delta_k(r)}
     {R_{\max}B_k^2}.
$
For state-only rewards, $C^\pi r=0$ and
$$
\Delta_k(r)
=\left(\frac{\gamma}{1-\gamma}\right)^2
  \sum_{i\le k}(\sigma_i^{+,\pi})^2\beta_i(r)^2.
$$
In this case, if $\Delta_k(r)>0$, the larger sufficient range
\[
    0<\tau<
    \frac{4\gamma}{R_{\max}B_k^2}
    \sum_{i\le k}(\sigma_i^{+,\pi})^2\beta_i(r)^2
\]
also guarantees strict improvement.
\end{theorem}

\paragraph{Policy Gradient with Trust Region}

Beyond one-step policy improvement, SUAF can also be reused under local policy changes. Trust-region methods constrain policy updates to prevent destructive changes \cite{schulman2015trust,schulman2017proximal}. We show that the error of reusing SUAF from an old policy is controlled within the trust region.

Consider two policies $\pi$ and $\pi_\theta$ in the same finite MDP with a common initial distribution and $0<\gamma<1$.
Assume that $d^\pi$ and $\nu^{+,\pi}$ have full support. Using the SUAF $f^\pi$ and $g^\pi$ learned under $\pi$, we approximate the advantage under $\pi'$ by
\begin{equation}
\label{eq:transferred_advantage}
\begin{aligned}
    \widehat A_r
    &=
    C_{\pi_\theta}r
    +
    \frac{\gamma}{1-\gamma}
    \sum_{i=1}^{R}
    \sigma_i^{+,\pi}
    \bigl(C_{\pi_\theta}f_i^\pi\bigr)
    \beta_i(r), \\
    \widehat A_r^k
    & =
    C_{\pi_\theta}r
    +
    \frac{\gamma}{1-\gamma}
    \sum_{i=1}^{k}
    \sigma_i^{+,\pi}
    \bigl(C_{\pi_\theta}f_i^\pi\bigr)
    \beta_i(r)
\end{aligned}
\end{equation}
where
$C_{\pi_\theta}u(s,a)
=u(s,a)-\mathbb E_{a\sim\pi_\theta(\cdot\mid s)}[u(s,a)]$ and $\beta_i(r)=\langle g_i^\pi,r\rangle_{\nu^{+,\pi}}$.
This approximation retains the old features and reward coefficients, while recentering the action-dependent features under the new policy.
The discrepancy from $A_r^{\pi_\theta}$ isolates the error caused by reusing the old policy's future information.
We measure policy shift by the KL divergence as:
\begin{align}
    \gD_{\mathrm{KL}}(\pi_\theta|\pi)
    &:=
    \mathbb E_{d_S^{\pi_\theta}}
    \left[
        D_{\mathrm{KL}}\bigl(
            \pi_\theta(\cdot| s)\|\pi(\cdot| s)
        \bigr)
    \right],\\
    \gD_{\mathrm{KL}}(\pi\|\pi_\theta)
    &:=
    \mathbb E_{d_S^{\pi}}
    \left[
        D_{\mathrm{KL}}\bigl(
            \pi(\cdot| s)\|\pi_\theta(\!\cdot| s)
        \bigr)
    \right],
\end{align}
where $d_S^\pi(s)=\sum_a d^\pi(s,a)$ is marginal state distribution\footnote{$\gD_{\mathrm{KL}}(\pi\|\pi_\theta)$ and $\gD_{\mathrm{KL}}(\pi_\theta\|\pi)$ are usually referred as the forward and reverse KL divergence respectively.}.
Define the action sensitivity
\begin{equation}
    L_{\max}
    :=
    \sup_{\|r\|_{\nu^{+,\pi}}\leq 1}
    \max\{
    \max_{s,a,b} \left|A_r^\pi(s,a)-A_r^\pi(s,b)\right|,
    \max_{s,a,b} \left|A_r^{\pi_\theta}(s,a)-A_r^{\pi_\theta}(s,b)\right|
    \}.
\end{equation}
For the policy gradient $\nabla_\theta \bar{J}_r(\pi_\theta) = (1-\gamma)\nabla_\theta J_r(\pi_\theta)$, we approximate the gradient as $\widehat{G}_r = \E_{d^{\pi_\theta}}[\nabla_\theta \log \pi_\theta \widehat A_r]$ and rank-$k$ approximation 
$$
\widehat G_\theta^k(r) = \mathbb E_{d^{\pi_\theta}}[\nabla_\theta\log\pi_\theta\,\widehat A_r^k].
$$

\begin{theorem}[Policy improvement using transferred SUAF]
\label{thm:suaf_policy_improvement}
Given the policy $\pi_{\theta}$ in the trust region of $\pi_{\text{ref}}$, i.e.,
$
\gD(\pi_\theta \| \pi_{\text{ref}}) < \delta,
$, fix a reward $r$ with $\|r\|_{\nu^{+,\pi_{\text{ref}}}}\le1$ and define
$
    u_r
    :=
    F_\theta^\dagger\widehat G_r,
    s_r
    :=
    \|\widehat G_r\|_{F_\theta^\dagger},
    \varepsilon_\delta
    :=
    (
        \frac{\gamma L_{\max}^2}{2(1-\gamma)}
        \delta
    )^{1/2}
$ with $F_\theta=\E_\theta[\nabla_\theta\log\pi_\theta(\nabla_\theta\log\pi_\theta)^\text{T}]$ being the Fisher matrix.
Then
$
    \nabla_\theta \bar{J}_r(\pi_\theta)^\top u_r
    \ge
    s_r(s_r-\varepsilon_\delta).
$
In particular, if $s_r>\varepsilon_\delta$, then $u_r$ is a strict ascent direction, and there exists $\eta_0>0$ such that
$
    J_r(\pi_{\theta+\eta u_r})
    >
    J_r(\pi_\theta),
    0<\eta<\eta_0.
$
Moreover, if $\theta\mapsto \bar{J}_r(\pi_\theta)$ has an $L$-Lipschitz gradient on a convex neighborhood containing the update segment, strict improvement holds whenever 
\begin{equation*}
    0<\eta<
    \frac{2s_r(s_r-\varepsilon_\delta)}
         {L\|u_r\|_2^2}.
\end{equation*}
\end{theorem}
Theorem \ref{thm:suaf_policy_improvement} indicates that, when the estimated gradient signal exceeds the approximation error, the resulting natural-gradient direction is a strict ascent direction, and sufficiently small steps improve policy performance. These results align with the trust-region principle and motivate integrating SUAF as a reusable critic into existing policy optimization frameworks, such as TRPO \cite{schulman2015trust} and PPO \cite{schulman2017proximal}. After training, SUAF and the reward-conditioned actor are frozen. For a new reward, only the linear coefficients are computed, and the policy is obtained by a single forward pass without any actor update.
\begin{theorem}[Policy-gradient error with truncated SUAF]
\label{cor:truncated_suaf_pg}
Under the assumption in Theorem \ref{thm:suaf_policy_improvement}.
For
$\eta_\delta=\min\{1,\sqrt{\delta/[2(1-\gamma)]}\}$, $B(\delta)=\min\{R^2_{\max}, 1+ R^2_{\max}C_\gamma\sqrt{\delta}\}$ with $C_\gamma = \sqrt{(4+\gamma^{-1})/(2(1-\gamma)^2)}$,
\begin{align*}
&\sup_{\|r\|_{\nu^{+,\pi_\theta}}\le1}
\left\|
\nabla_\theta\bar J_r(\pi_\theta)
-\widehat G_\theta^k(r)
\right\|_{F_\theta^\dagger}^2\\
&\quad\quad \le
B(\delta)\left[
\sqrt{\frac{\gamma\delta L_{\max}^2}{2(1-\gamma)}}
+
\sqrt{
\left(\frac{\gamma}{1-\gamma}\right)^2
\left(\sigma_{k+1}^{+,\pi}\right)^2
+
L_{\max}^2\eta_\delta
}
\right]^2.
\end{align*}
\end{theorem}

{\color{black} Theorem \ref{cor:truncated_suaf_pg} shows that trust-region control remains effective with truncated SUAF. The policy-gradient error bound accounts for both policy shift and spectral truncation. A smaller trust region controls the error from reusing reference-policy features, while retaining more spectral modes reduces the truncation error. When the policy is unchanged, only the truncation term remains. In contrast, when all nonzero modes are retained, the bound tends to zero as the policy shift vanishes.}


\paragraph{Soft-HGR}  Soft-HGR provides an approach to learning feature functions by replacing hard orthogonality constraints with a covariance penalty \cite{wang2019efficient}. We adapt this objective to the DCDM and use bidirectional temporal-difference (TD) learning to estimate the long-horizon conditional expectations. The detailed process of algorithm can be found in Appendix \ref{sec:leanring_suaf}.
We parameterize two $k$-dimensional feature functions,
$f_\theta$ and $g_\omega$, with neural networks parameters $\theta$ and $\omega$ respectively.
enforcing
$
\mathbb E_{\pi}[f_\theta|s]=0,
\mathbb E_{\nu^{+,\pi}}[g_\omega]=0.
$
These conditions are implemented by subtracting the corresponding conditional and marginal means. In continuous spaces, the means are approximated using samples.
Furthermore, defining $C_f=\mathbb E_{d^\pi}
[f_\theta f_\theta^\text{T}]$ and $C_g=\mathbb E_{\nu^{+,\pi}}
[g_\omega g_\omega^\text{T}]$, the feature-learning objective of soft-HGR is
\begin{equation}
\label{eq:suaf_soft_hgr}
\max_{\theta,\omega}
\mathcal J(\theta,\omega) = \E_{(s,a)\sim d^\pi}\E_{(s^+,a^+)\sim M^{+,\pi}(\cdot|s,a)}
[f_\theta^\text{T}(s,a) g_\omega(s^+,a^+)] - \frac12\operatorname{tr}(C_fC_g).
\end{equation}
The conditional centering of $f_\theta$ removes dependence explained solely by the current state, so the objective targets the action-dependent successor residual represented by
$\widetilde{\boldsymbol B}^{+,\pi}$. Soft-HGR learns low-rank factors rather than individually normalized singular functions. Weighted whitening is then applied to obtain the canonical feature pairs and estimated singular values using  sample moments.

\paragraph{Bidirectional conditional prediction.}
The original soft-HGR method requires full trajectories, which can be a burden. We instead introduce bidirectional Bellman conditional prediction. The correlation term in \eqref{eq:suaf_soft_hgr} admits two equivalent expressions:
\begin{equation}
\label{eq:suaf_bidirectional}
\mathbb E_{(s,a)\sim d^\pi}\E_{(s^+,a^+)\sim M^{+,\pi}(\cdot|s,a)}[f_\theta^\text{T}(s,a) g_\omega(s^+,a^+)]
= \mathbb E_{d^\pi}[f_\theta^\text{T}h]
= \mathbb E_{\nu^{+,\pi}}[b^\text{T} g_\omega],
\end{equation}
where
$
h(s,a)=\mathbb E[g_\omega(s^+,a^+)| s,a]
$
and
$
b(s^+,a^+)=\mathbb E[f_\theta(s,a)| s^+,a^+].
$
We estimate these quantities with auxiliary prediction networks. For one-step transitions $(s,a)\sim d^\pi$ and $(s^\prime,a^\prime)\sim P_\pi(\cdot|s,a)$, the forward predictor and backward recursion satisfy
\begin{align*}
h(s,a)
&=
\E_{P_\pi}\!\left[
(1-\gamma)g_\omega(s^\prime,a^\prime)+\gamma h(s^\prime,a^\prime)
\mid s, a
\right], \\
v(s^\prime, a^\prime)
& =
w_1(s^\prime, a^\prime)\,
\mathbb E\!\left[
(1-\gamma)f_\theta(s,a)+\gamma v(s,a)
\mid s^\prime, a^\prime
\right],
\end{align*}
where $\nu^{+,\pi}(s,a)b(s,a) = d^\pi(s,a) v(s,a)$ and $w_1(s,a) = \nu_1^\pi(s,a)/d^\pi(s,a)$. Accordingly, networks $h_\xi$ and $v_\eta$ are trained with TD objectives
\begin{equation}
\label{eq:suaf_forward_td}
\gL_{\mathrm F}(\xi) = \E_{(s,a)\sim d^\pi}\E_{(s^\prime,a^\prime)\sim P_\pi}
\left[
\left\|
h_\xi(s,a)
-
\operatorname{sg}\!\left[
(1-\gamma)g_\omega(s^\prime,a^\prime)
+\gamma h_{\bar\xi}(s^\prime,a^\prime)
\right]
\right\|_2^2
\right],
\end{equation}
\begin{equation*}
\begin{aligned}
&\gL_{\mathrm B}(\eta) = \mathbb E_{(s^\prime,a^\prime)\sim \nu_1^\pi}\E_{(s,a)\sim p_\pi}
\left[
\left\|
v_\eta(s^\prime, a^\prime)\right.)\right. \\
&\quad \left.\left. - \operatorname{sg}\!\left[
w_1(s^\prime, a^\prime)
\bigl((1-\gamma)f_\theta(s,a)
\!+\!\gamma v_{\bar\eta}(s,a)\bigr)
\right]
\right\|_2^2
\right],
\end{aligned}
\end{equation*}
where $\operatorname{sg}$ denotes stop-gradient and $\bar\xi,\bar\eta$ are target-network parameters. The density ratio $w_1$ can be eliminated by a surrogate loss $\widehat{\gL}_{\mathrm B}(\eta)$ sharing the same minimum as $\gL_{\mathrm B}(\eta)$. We defer its derivation to Appendix~\ref{sec:leanring_suaf}.

\paragraph{Alternating soft-HGR updates.}
With the prediction networks held fixed, we alternate between minimizing
\begin{equation}
\label{eq:suaf_alternate_update}
\gL_f(\theta)
\!=\! 
\E_{d^\pi}\!\!
\left[
\!-f_\theta^\text{T}\operatorname{sg}[h_\xi]
\!+\!\frac12f_\theta^\text{T}\operatorname{sg}[C_g]f_\theta
\right],
\gL_g(\omega)
\!=\!
\E_{\nu^{+,\pi}}\!\!
\left[
\!-g_\omega^\text{T}\operatorname{sg}[\widehat b_\theta]
\!+\!\frac12g_\omega^\text{T}\operatorname{sg}[C_f]g_\omega
\right].
\end{equation}
With exact conditional predictors and population moments, these updates are precisely block gradient ascent on \eqref{eq:suaf_soft_hgr}. The forward and backward predictors, $h_\xi$ and $v_\eta$ are held fixed during each feature update. With exact conditional predictions and population moments, these updates recover the corresponding soft-HGR gradients without differentiating through the temporal-difference (TD) solutions. In practice, prediction and moment-estimation errors introduce errors into these gradient estimates. 
Expectations are estimated using samples. In particular, $\nu^{+,\pi}$ samples are obtained by selecting an anchor from $d^\pi$ and a future lag $L$ with $\Pr(L=\ell)=(1-\gamma)\gamma^{\ell-1}$.

\section{Experiments}
\label{sec:experiments}

{\color{black} This section evaluates whether SUAF provides accurate advantage representations and effective zero-shot control across reward tasks. We first describe the experimental setup and evaluation protocol. We then use finite-state environments to examine representation accuracy, the benefit of separating immediate and delayed effects, and the resulting policy improvement. Finally, we evaluate zero-shot performance on continuous locomotion, navigation, and manipulation tasks against feature-learning and successor-based methods.}

\subsection{Experimental Setup}
\paragraph{Discrete Finite Environments}
We evaluate on Grid World \cite{MinigridMiniworld23}, Four-room \cite{sutton1999between}, and Taxi \cite{dietterich2000hierarchical} to cover basic navigation, narrow passages with long-range credit assignment, and compositional tasks (navigation, pickup, delivery), respectively. These domains vary in state-space size and temporal structure. For each, we partition reward tasks into in-distribution {\color{black}(ID)} and out-of-distribution {\color{black}(OOD)} sets while fixing dynamics. Representations are learned from reward-free transitions under a behavior policy and rewards are used only for policy extraction. Since states and transitions are enumerable, we compute the optimal return \(J(\pi^*)\) via dynamic programming.
\paragraph{Continuous State and Action Environments.}
We further use four state-based continuous-control domains: walker and cheetah from ExORL \cite{yarats2022don}, and antmaze-large-navigate (denoted as antmaze-large for short) and cube-single-play from OGBench \cite{park2025ogbench}. These jointly cover locomotion under varied rewards, goal navigation, and robotic pick-and-place. Representations are trained on reward-free offline transitions and reused across downstream tasks. As tabular dynamic programming is infeasible, we follow benchmark metrics: undiscounted episodic return for ExORL and task success rate for OGBench.
{\color{black}Futhermore, we evaluate generalization to held-out reward tasks by retraining all methods under predefined ID/OOD splits. For Walker, stand and walk are ID tasks, while run and flip are OOD tasks. For Cheetah, run and run backward are ID tasks, while walk and walk backward are OOD tasks. For both Antmaze-large and Cube-single-play, Tasks 1–2 form the ID set and Tasks 3–4 form the OOD set.}
\paragraph{Baselines} For baselines, we compare against methods that are commonly used and are the state of the art in reward-independent features, including Laplacian features \cite{wu2019laplacian}, BYOL-$\gamma$ \cite{lawson2026self}, TD-JEPA \cite{bagatella2026td}, ICVF \cite{ghosh2023reinforcement}, HILP \cite{park2024foundation}, Forward-Backward \cite{touati2022does} and One-Step FB \cite{zheng2026can}.

\subsection{Results and Analysis}
\label{sec:results}

\begin{figure}[t]
    \centering
    \includegraphics[width=\linewidth]{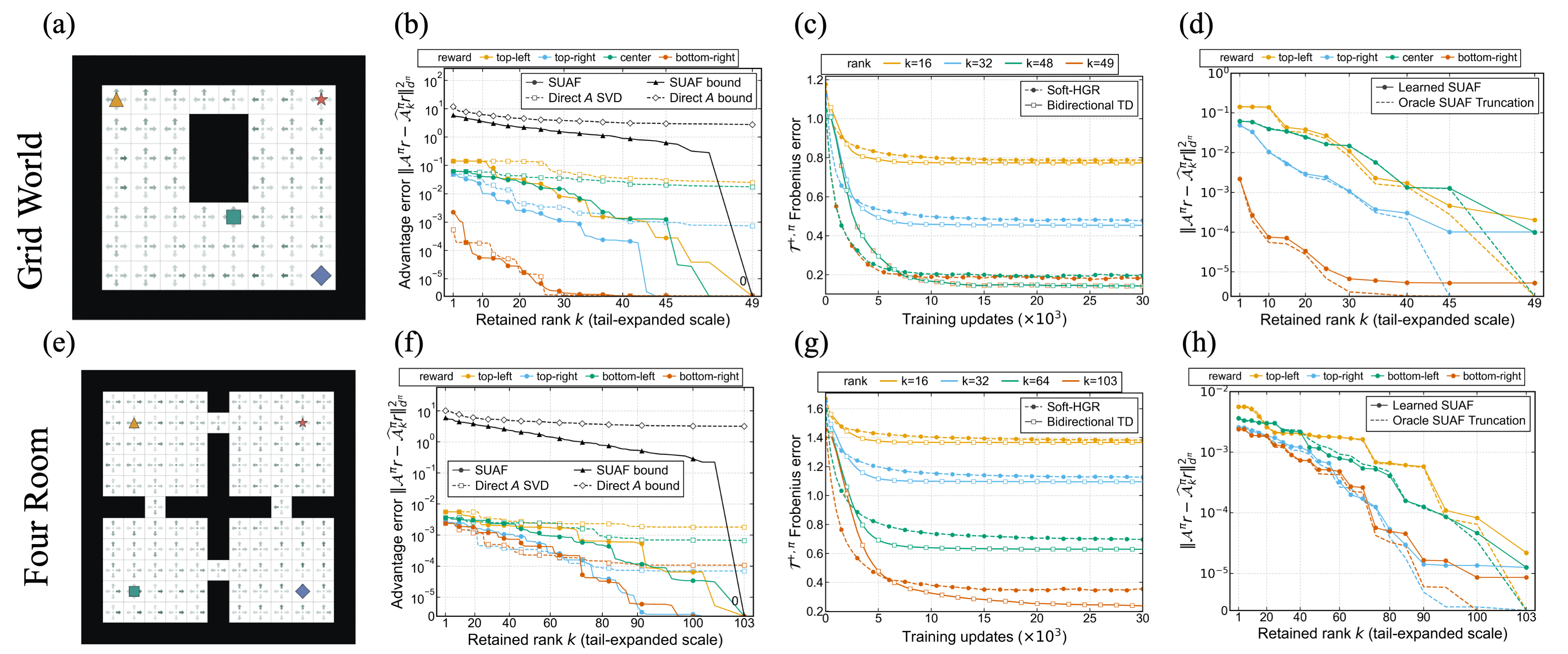}
    \caption{\textbf{Spectral approximation and learning of SUAF in Grid World (top) and Four-room (bottom).} \textbf{(a, e)} Environments with reward locations. Background arrows indicate the policy. \textbf{(b, f)} Advantage approximation errors for SUAF truncation and direct SVD truncation, together with theoretical worst-case error bounds. \textbf{(c, g)} Weighted Frobenius errors for approximating \(\mathcal{T}^{+,\pi}\) using Soft-HGR and Bidirectional TD at different ranks. \textbf{(d, h)} Reward-specific advantage approximation errors of SUAF learned using Bidirectional TD, compared with exact SUAF truncation computed from the exact operator. Rank axes in (b, d, f, h) use a nonlinear scale that expands the spectral tail.
}
    \label{fig:env_theorem}
\end{figure}

\begin{table*}[t]
\centering
\caption{\textbf{Zero-shot evaluation on finite state environments.} The performance is measured using the improvement rate defined as $\frac{J(\pi) - J(\pi_\beta)}{J(\pi^\ast) - J(\pi_\beta)}$, where $\pi^\ast$ is the optimal policy, $\pi$ is the generated policy and $\pi_\beta$ is the sample policy. Each result is measured as \textit{mean} with \textit{95\% confidence interval} and reports with 5 random sample policies with 5 seeds for each policy.}
\label{tab:ood}
\resizebox{\textwidth}{!}{
\begin{tabular}{l|ccccccc|c}
\toprule
Domain (Task Num.) & Laplacian & BYOL-\(\gamma\) & TD-JEPA & ICVF & HILP & FB & One-Step FB & Ours \\
\midrule
\rowcolor{gray!15} \multicolumn{9}{l}{\textbf{In-Distribution Tasks}} \\
Grid World (4)
& $0.006_{\pm0.030}$ 
& $0.690_{\pm0.113}$ 
& {$\mathbf{0.993}_{\pm0.015}$} 
& $0.804_{\pm0.014}$ 
& $0.770_{\pm0.040}$ 
& \underline{$0.929_{\pm0.048}$} 
& $0.599_{\pm0.036}$ 
& $0.768_{\pm0.040}$ \\
Four-room (4)
& $0.000_{\pm0.007}$
& $0.042_{\pm0.016}$
& {$\mathbf{0.900}_{\pm0.063}$}
& $0.293_{\pm0.040}$
& $0.450_{\pm0.167}$
& \underline{$0.773_{\pm0.110}$}
& $0.175_{\pm0.102}$
& $0.510_{\pm0.008}$ \\
Taxi (4)
& $0.029_{\pm0.256}$
& $0.064_{\pm0.009}$
& $-0.037_{\pm0.085}$
& \underline{$0.071_{\pm0.016}$}
& $0.042_{\pm0.113}$
& $-0.226_{\pm0.015}$
& $0.067_{\pm0.040}$
& {$\mathbf{0.264}_{\pm0.018}$} \\
\midrule
\rowcolor{gray!15} \multicolumn{9}{l}{\textbf{OOD Tasks}} \\
Grid World (2)
& $0.061_{\pm0.059}$
& $0.443_{\pm0.055}$
& $0.446_{\pm0.061}$
& $0.437_{\pm0.021}$
& $0.405_{\pm0.044}$
& \underline{$0.474_{\pm0.087}$}
& $0.424_{\pm0.038}$
& {$\mathbf{0.873}_{\pm0.016}$} \\
{Four-room (2)}
& $0.070_{\pm0.028}$
& $0.169_{\pm0.103}$
& $0.347_{\pm0.020}$
& $0.283_{\pm0.085}$
& $0.150_{\pm0.082}$
& \underline{$0.385_{\pm0.073}$}
& $0.180_{\pm0.012}$
& {$\mathbf{0.636}_{\pm0.009}$} \\
Taxi (3)
& $-0.014_{\pm0.111}$
& $-0.027_{\pm0.002}$
& $0.009_{\pm0.025}$
& $-0.026_{\pm0.015}$
& $-0.021_{\pm0.048}$
& \underline{$0.085_{\pm0.013}$}
& $-0.016_{\pm0.019}$
& {$\mathbf{0.300}_{\pm0.010}$} \\
\midrule
\bottomrule
\end{tabular}
}
\end{table*}

\begin{table*}[htbp]
\centering
\caption{\textbf{Zero-shot evaluation on state-based ExORL and OGBench benchmarks.} The full zero-shot evaluation results on 8 ExORL and 10 OGBench tasks. ExORL is measured with returns and OGBench is measured with success rate. The results are reported as \textit{mean} over 8 seeds with \textit{std}.}
\label{tab:zero_shot}
\resizebox{\textwidth}{!}{
\begin{tabular}{l l|c c c c c c c | c}
\toprule
Domain & Task & Laplacian & BYOL-\(\gamma\) & TD-JEPA & ICVF & HILP & FB & One-Step FB & Ours \\
\midrule
\rowcolor{gray!15} \multicolumn{10}{l}{\textbf{ExORL}: Locomotion}\\
\multirow{5}{*}{walker} 
& overall
& 228 \(\pm\) 2 
& 227 \(\pm\) 2 
& 316 \(\pm\) 12 
& \textbf{619 \(\pm\) 23} 
& 393 \(\pm\) 108 
& 400 \(\pm\) 40 
& 379 \(\pm\) 26 
& \underline{445 \(\pm\) 39} \\

& flip 
& 243 \(\pm\) 5 
& 242 \(\pm\) 4 
& 381 \(\pm\) 34 
& \textbf{538 \(\pm\) 19} 
& 332 \(\pm\) 135 
& 277 \(\pm\) 100 
& \underline{388 \(\pm\) 30} 
& 357 \(\pm\) 42 \\

& run & 89 \(\pm\) 1 
& 89 \(\pm\) 2 
& 112 \(\pm\) 22 
& \textbf{258 \(\pm\) 20} 
& 136 \(\pm\) 36 
& 194 \(\pm\) 20 
& 198 \(\pm\) 30 
& \underline{210 \(\pm\) 20} \\

& stand 
& 389 \(\pm\) 3 
& 387 \(\pm\) 5 
& 387 \(\pm\) 11 
& \textbf{858 \(\pm\) 31} 
& \underline{691 \(\pm\) 126}
& 621 \(\pm\) 93 
& 524 \(\pm\) 60 
& 670 \(\pm\) 100 \\

& walk 
& 192 \(\pm\) 4 
& 192 \(\pm\) 6
& 384 \(\pm\) 47 
& \textbf{821 \(\pm\) 40} 
& 413 \(\pm\) 195 
& 506 \(\pm\) 34
& 407 \(\pm\) 63 
& \underline{542 \(\pm\) 41}\\

\midrule
\multirow{5}{*}{cheetah} 
& overall 
& 125 \(\pm\) 41 
& 127 \(\pm\) 39 
& 185 \(\pm\) 7 
& 187 \(\pm\) 13 
& 116 \(\pm\) 78 
& 271 \(\pm\) 46 
& \textbf{378 \(\pm\) 56} 
& \underline{375 \(\pm\) 21}\\

& run 
& 40 \(\pm\) 13 &
42 \(\pm\) 13 
& 66 \(\pm\) 3 
& {89 \(\pm\) 7} 
& 38 \(\pm\) 32 
& 43 \(\pm\) 36 
& \underline{97 \(\pm\) 19} 
& \textbf{100} \(\pm\) 12 \\

& run backward 
& 50 \(\pm\) 17 
& 48 \(\pm\) 15 
& 69 \(\pm\) 6 
& 48 \(\pm\) 9 
& 36 \(\pm\) 40 
& 125 \(\pm\) 30 
& \textbf{181 \(\pm\) 60} 
& \underline{159 \(\pm\) 70}\\

& walk 
& 185 \(\pm\) 61 
& 199 \(\pm\) 61 
& 293 \(\pm\) 15 
& 385 \(\pm\) 21 
& 195 \(\pm\) 121 
& 251 \(\pm\) 166 
& \textbf{424 \(\pm\) 44} 
& \underline{420 \(\pm\) 32}\\

& walk backward 
& 226 \(\pm\) 77 
& 220 \(\pm\) 70 
& 312 \(\pm\) 26 
& 226 \(\pm\) 38 
& 194 \(\pm\) 202 
& 663 \(\pm\) 143 
& \underline{811 \(\pm\) 156} 
& \textbf{820 \(\pm\) 232}\\
\midrule
\rowcolor{gray!15} \multicolumn{10}{l}{\textbf{OGBench}: Navigation and manipulation}\\
\multirow{6}{*}{antmaze large} 
& overall 
& 9 \(\pm\) 1 
& 21 \(\pm\) 2 
& 14 \(\pm\) 9 
& 23 \(\pm\) 3 
& \underline{34 \(\pm\) 2} 
& 25 \(\pm\) 5 
& 30 \(\pm\) 9 
& \textbf{35 \(\pm\) 5}\\

& task 1 
& 2 \(\pm\) 1 
& 6 \(\pm\) 3 
& 15 \(\pm\) 11 
& 4 \(\pm\) 2 
& 13 \(\pm\) 8 
& \textbf{46 \(\pm\) 9} 
& 21 \(\pm\) 9 
& \underline{36 \(\pm\) 8}\\

& task 2 
& 2 \(\pm\) 1 
& 11 \(\pm\) 4 
& 1 \(\pm\) 2 
& 9 \(\pm\) 3 
& {16 \(\pm\) 6} 
& 2 \(\pm\) 3 
& \textbf{41 \(\pm\) 12} 
& \underline{24 \(\pm\) 1}\\

& task 3 
& 29 \(\pm\) 3 
& 57 \(\pm\) 8 
& 35 \(\pm\) 23 
& \underline{67 \(\pm\) 8 }
& \textbf{75 \(\pm\) 6} 
& 31 \(\pm\) 10 
& 15 \(\pm\) 4 
& 57 \(\pm\) 5\\

& task 4 
& 6 \(\pm\) 2 
& 14 \(\pm\) 5 
& 7 \(\pm\) 7 
& 18 \(\pm\) 6 
& \underline{27 \(\pm\) 10}
& 3 \(\pm\) 2 
& \textbf{33 \(\pm\) 15} 
& 12 \(\pm\) 6 \\

& task 5 
& 6 \(\pm\) 3 
& 16 \(\pm\) 4 
& 11 \(\pm\) 8 
& 18 \(\pm\) 4 
& \underline{40 \(\pm\) 8 }
& \textbf{44 \(\pm\) 19 }
& 37 \(\pm\) 20 
& \textbf{44 \(\pm\) 9} \\

\midrule
\multirow{6}{*}{cube single play} 
& overall 
& 6 \(\pm\) 2 
& 13 \(\pm\) 2 
& 8 \(\pm\) 4 
& 13 \(\pm\) 2 
& \textbf{30 \(\pm\) 8} 
& 2 \(\pm\) 1 
& 3 \(\pm\) 2 
& \underline{21 \(\pm\) 10}\\

& task 1 
& 5 \(\pm\) 2 
& 12 \(\pm\) 4 
& 6 \(\pm\) 4 
& 13 \(\pm\) 2 
& \textbf{27 \(\pm\) 7} 
& 1 \(\pm\) 1 
& 3 \(\pm\) 4 
& \underline{20 \(\pm\) 7} \\

& task 2 
& 5 \(\pm\) 2 
& 13 \(\pm\) 3 
& 8 \(\pm\) 6 
& 13 \(\pm\) 3 
& \textbf{30 \(\pm\) 10} 
& 3 \(\pm\) 2 
& 4 \(\pm\) 4 
& \underline{17 \(\pm\) 12}\\

& task 3 
& 7 \(\pm\) 3 
& 13 \(\pm\) 4 
& 14 \(\pm\) 9 
& 14 \(\pm\) 4 
& \textbf{23 \(\pm\) 13} 
& 3 \(\pm\) 3 & 4 \(\pm\) 4 
& \underline{18 \(\pm\) 9} \\

& task 4 
& 6 \(\pm\) 2 
& 15 \(\pm\) 6 
& 6 \(\pm\) 4 
& 11 \(\pm\) 3 
& \textbf{37 \(\pm\) 19} 
& 2 \(\pm\) 2 
& 2 \(\pm\) 2 
& \underline{21 \(\pm\) 12}\\

& task 5 
& 5 \(\pm\) 3 
& 11 \(\pm\) 3 
& 5 \(\pm\) 4 
& 13 \(\pm\) 4 
& \textbf{30 \(\pm\) 22} 
& 1 \(\pm\) 2 
& 1 \(\pm\) 1 
& \underline{28 \(\pm\) 20} \\
\bottomrule
\end{tabular}
}
\end{table*}

{\color{black}
\begin{table*}[htbp]
\centering
\caption{{\color{black}\textbf{Generalization ability evaluation on state-based ExORL and OGVench benchmarks.}
Walker is measured by episodic return and Antmaze-large by success rate (\%).
Results are reported over 8 seeds.
The best and second-best means in each row are shown in bold and underlined, respectively.}}
\label{tab:zero_shot_id_ood_details_continuous}
\resizebox{\textwidth}{!}{
\begin{tabular}{l l|c c c c c c c|c}
\toprule
Domain & Task
& Laplacian & BYOL-\(\gamma\) & TD-JEPA
& ICVF & HILP & FB & One-Step FB & SUAF \\
\midrule

\rowcolor{gray!15}
\multicolumn{10}{l}{\textbf{In-distribution (ID) tasks}} \\

\multirow{3}{*}{walker}

& overall
& 290.5 \(\pm\) 4
& 289.5 \(\pm\) 6
& 385.5 \(\pm\) 34
& \textbf{839.5} \(\pm\) 36
& 552.0 \(\pm\) 164
& 563.5 \(\pm\) 70
& 465.5 \(\pm\) 62
& \underline{606.0} \(\pm\) 76 \\

& stand
& 389 \(\pm\) 3
& 387 \(\pm\) 5
& 387 \(\pm\) 11
& \textbf{858} \(\pm\) 31
& \underline{691} \(\pm\) 126
& 621 \(\pm\) 93
& 524 \(\pm\) 60
& 670 \(\pm\) 100 \\

& walk
& 192 \(\pm\) 4
& 192 \(\pm\) 6
& 384 \(\pm\) 47
& \textbf{821} \(\pm\) 40
& 413 \(\pm\) 195
& 506 \(\pm\) 34
& 407 \(\pm\) 63
& \underline{542} \(\pm\) 41 \\

\midrule
\multirow{3}{*}{cheetah}

& overall
& 45 \(\pm\) 15
& 45 \(\pm\) 14
& 67.5 \(\pm\) 5
& 68.5 \(\pm\) 8
& 37 \(\pm\) 36
& 84 \(\pm\) 33
& \textbf{139} \(\pm\) 45
& \underline{129.5} \(\pm\) 50 \\

& run 
& 40 \(\pm\) 13 &
42 \(\pm\) 13 
& 66 \(\pm\) 3 
& {89 \(\pm\) 7} 
& 38 \(\pm\) 32 
& 43 \(\pm\) 36 
& \underline{97} \(\pm\) 19 
& \textbf{100} \(\pm\) 12 \\

& run backward 
& 50 \(\pm\) 17 
& 48 \(\pm\) 15 
& 69 \(\pm\) 6 
& 48 \(\pm\) 9 
& 36 \(\pm\) 40 
& 125 \(\pm\) 30 
& \textbf{181} \(\pm\) 60 
& \underline{159} \(\pm\) 70\\

\midrule
\multirow{3}{*}{antmaze-large}
& overall 
& 2 \(\pm\) 1 
& 8.5 \(\pm\) 4 
& 8 \(\pm\) 8 
& 6.5 \(\pm\) 3 
& 14.5 \(\pm\) 7
& 24 \(\pm\) 7 
& \textbf{31} \(\pm\) 11
& \underline{30} \(\pm\) 6\\

& task 1
& 2 \(\pm\) 1
& 6 \(\pm\) 3
& 15 \(\pm\) 11
& 4 \(\pm\) 2
& 13 \(\pm\) 8
& \textbf{46} \(\pm\) 9
& 21 \(\pm\) 9
& \underline{36} \(\pm\) 8 \\

& task 2
& 2 \(\pm\) 1
& 11 \(\pm\) 4
& 1 \(\pm\) 2
& 9 \(\pm\) 3
& 16 \(\pm\) 6
& 2 \(\pm\) 3
& \textbf{41} \(\pm\) 12
& \underline{24} \(\pm\) 1 \\

\midrule
\multirow{3}{*}{cube single play} 
& overall 
& 5 \(\pm\) 2 
& 12.5 \(\pm\) 4 
& 7 \(\pm\) 5 
& 13 \(\pm\) 3 
& \textbf{28.5} \(\pm\) 9
& 2 \(\pm\) 2 
& 3.5 \(\pm\) 4 
& \underline{18.5} \(\pm\) 10\\

& task 1 
& 5 \(\pm\) 2 
& 12 \(\pm\) 4 
& 6 \(\pm\) 4 
& 13 \(\pm\) 2 
& \textbf{27} \(\pm\) 7
& 1 \(\pm\) 1 
& 3 \(\pm\) 4 
& \underline{20} \(\pm\) 7 \\

& task 2 
& 5 \(\pm\) 2 
& 13 \(\pm\) 3 
& 8 \(\pm\) 6 
& 13 \(\pm\) 3 
& \textbf{30} \(\pm\) 10
& 3 \(\pm\) 2 
& 4 \(\pm\) 4 
& \underline{17} \(\pm\) 12\\

\midrule
\rowcolor{gray!15}
\multicolumn{10}{l}{\textbf{Out-of-distribution (OOD) tasks}} \\

\multirow{3}{*}{walker}

& overall
& 115 \(\pm\) 6
& 84.5 \(\pm\) 14
& 119.5 \(\pm\) 29
& 258 \(\pm\) 20 
& 203 \(\pm\) 70
& 177 \(\pm\) 60
& \underline{270} \(\pm\) 31
& \textbf{271.5} \(\pm\) 25 \\

& run
& 48 \(\pm\) 4 %
& 37 \(\pm\) 6 %
& 63 \(\pm\) 34 %
& 147 \(\pm\) 25 %
& 105 \(\pm\) 32 %
& 91 \(\pm\) 18 %
& \underline{176} \(\pm\) 28 %
& \textbf{198 \(\pm\) 16} \\ %

& flip
& 182 \(\pm\) 8 %
& 132 \(\pm\) 19 %
& 176 \(\pm\) 23 %
& \textbf{369} \(\pm\) 13 %
& 301 \(\pm\) 93 %
& 263 \(\pm\) 82 %
& \underline{364} \(\pm\) 34 %
& 345 \(\pm\) 32 \\ %

\midrule
\multirow{3}{*}{cheetah}

& overall
& 132.5 \(\pm\) 54 %
& 115.5 \(\pm\) 66 %
& 204 \(\pm\) 28 %
& 182.5 \(\pm\) 53 %
& 87.5 \(\pm\) 47 %
& 388 \(\pm\) 167 %
& \underline{530} \(\pm\) 98 %
& \textbf{573} \(\pm\) 222 \\ %

& walk 
& 75 \(\pm\) 34 %
& 89 \(\pm\) 56 %
& 153 \(\pm\) 32 %
& 276 \(\pm\) 52 %
& 69 \(\pm\) 62 %
& 189 \(\pm\) 174 %
& \underline{375} \(\pm\) 39 %
& \textbf{398} \(\pm\) 64\\ 

& walk backward 
& 190\(\pm\) 69 %
& 142 \(\pm\) 74 %
& 255 \(\pm\) 23 %
& 89 \(\pm\) 53 %
& 106 \(\pm\) 25 %
& 587 \(\pm\) 159 %
& \underline{685} \(\pm\) 134 %
& \textbf{748}\(\pm\) 308\\ %

\midrule
\multirow{3}{*}{antmaze large}
& overall
& 9 \(\pm\) 3
& 19.5 \(\pm\) 7
& 12 \(\pm\) 7
& 27.5 \(\pm\) 6
& \underline{28.5} \(\pm\) 11
& 14 \(\pm\) 9
& 20.5 \(\pm\) 14
& \textbf{30.5} \(\pm\) 7 \\

& task 3
& 17 \(\pm\) 4 %
& 37 \(\pm\) 10 %
& 23 \(\pm\) 10 %
& \underline{45} \(\pm\) 6 %
& 38 \(\pm\) 7 %
& 27 \(\pm\) 12 %
& 11 \(\pm\) 6 %
& \textbf{52} \(\pm\) 8 \\ %

& task 4
& 1 \(\pm\) 1 %
& 2 \(\pm\) 1 %
& 1 \(\pm\) 1 %
& 10 \(\pm\) 4 %
& \underline{19} \(\pm\) 13 %
& 1 \(\pm\) 1 %
& \textbf{30} \(\pm\) 19 %
& 9 \(\pm\) 6 \\ %

\midrule
\multirow{3}{*}{cube single play}

& overall
& 2.5 \(\pm\) 1 %
& 4 \(\pm\) 4 %
& 4.5 \(\pm\) 3 %
& 7.5 \(\pm\) 5 %
& \underline{15.5} \(\pm\) 10 %
& 1 \(\pm\) 1 %
& 1.5 \(\pm\) 1 %
& \textbf{17} \(\pm\) 11 \\ %

& task 3 
& 3 \(\pm\) 1 %
& 3 \(\pm\) 1 %
& 5 \(\pm\) 3 %
& 10 \(\pm\) 7 %
& \underline{14} \(\pm\) 8 %
& 1 \(\pm\) 1 %
& 2 \(\pm\) 1 %
& \textbf{16} \(\pm\) 10 \\ %

& task 4 
& 2 \(\pm\) 1 %
& 5 \(\pm\) 5 %
& 4 \(\pm\) 2 %
& 5 \(\pm\) 2 %
& \underline{17} \(\pm\) 12 %
& 1 \(\pm\) 1 %
& 1 \(\pm\) 1 %
& \textbf{18} \(\pm\) 11 \\ %

\bottomrule
\end{tabular}
}
\end{table*}
}


\paragraph{Spectral approximation and learning of SUAF.}
Figure \ref{fig:env_theorem} evaluates SUAF in Grid World and Four-room. SUAF represents the reward-to-advantage map by computing the immediate action contrast \(C_\pi r\) exactly and using learned spectral features only for the delayed, action-dependent component. This separation matters because directly compressing the full advantage operator incurs an immediate-term spectral floor when \(k<|\mathcal S|(|\mathcal A|-1)\). In contrast, the squared error of exact SUAF truncation is bounded, for \(\|r\|_{\nu^{+,\pi}}\leq1\), by \((\gamma/(1-\gamma))^2(\sigma_{k+1}^{+,\pi})^2\), which vanishes when all nonzero delayed modes are retained. Figures \ref{fig:env_theorem}(b,f) show that the reward-specific errors decrease with retained rank and remain below this worst-case bound. Direct SVD of the full advantage operator has substantially larger errors at the same budget of learned modes. We then test whether the SUAF can be learned from transitions. Across the ranks shown in Figures \ref{fig:env_theorem}(c,g), Bidirectional TD achieves lower weighted Frobenius error for the delayed operator than Soft-HGR. Finally, Figures \ref{fig:env_theorem}(d,h) show that the learned SUAF yields decreasing reward-specific advantage errors as rank increases. Its remaining gap from exact SUAF truncation, particularly at higher ranks, reflects feature-learning error in addition to the spectral truncation error.

\paragraph{Zero-shot policy generation ability.} (1) \textbf{Zero-shot policy extraction in finite-state environments.} Table \ref{tab:ood} evaluates the fraction of the improvement from the behavior policy to the optimal policy recovered by each method. Our method achieves the highest mean improvement rate on all three OOD task sets. Performance on in-distribution tasks is more mixed. TD-JEPA and FB outperform SUAF on Grid World and Four-room, whereas SUAF achieves the best result on Taxi. These results support the utility of SUAF for extracting policies across different reward tasks, with particularly favorable performance on the evaluated OOD tasks, while showing that it does not uniformly dominate existing representations. (2) \textbf{Evaluation on ExORL and OGBench.} Table \ref{tab:zero_shot} examines whether these benefits extend to continuous-state locomotion, navigation, and manipulation tasks. SUAF achieves the second-highest mean return on both ExORL domains. On OGBench, SUAF obtains the highest mean success rate on antmaze large, closely followed by HILP, and ranks second on cube single play, behind HILP. Given the reported variability, the small differences on cheetah and antmaze should be interpreted as comparable performance rather than decisive advantages. Overall, SUAF remains competitive across these domains, while the remaining gaps on walker and cube manipulation indicate room for improvement beyond the finite-state setting. {\color{black}(3) \textbf{OOD generalization in continuous-state domains.} Table \ref{tab:zero_shot_id_ood_details_continuous} further examines this capability on continuous-state tasks. SUAF consistently achieves the highest overall mean return across all four evaluated OOD continuous domains. The consistent top-ranking performance on the aggregated overall metrics across diverse environments suggests that SUAF's task-independent representation generalizes robustly when extrapolating to unseen reward functions, even in complex continuous control settings.}

\paragraph{Why SUAF shows strong OOD generalization?} {\color{black} SUAF achieves the highest OOD improvement rates across all three finite-state domains and all four continuous-state domains. SUAF performs well across both regimes, supporting its ability to combine the strengths of successor-based and feature-learning representations when transferring to held-out reward tasks.} A potential contributor is its explicit treatment of reward dependence: new rewards enter through linear projections onto shared features, without requiring a task-conditioned advantage predictor. Methods such as FB and TD-JEPA additionally rely on neural predictors conditioned on the task embedding, whose generalization across task embeddings can introduce approximation errors. One-Step FB shares SUAF’s task-independent representation and exhibits smaller ID-to-OOD performance changes on the navigation domains, consistent with this interpretation.  


\section{Conclusion}

In this work, we introduced SUAF, a reward-independent representation that preserves the immediate reward contrast and allocates learned capacity to delayed action effects. This decomposition avoids the immediate-term spectral floor of full advantage compression and yields optimal spectral approximations of the delayed operator. We connected representation errors to sufficient conditions for policy improvement and characterized local reuse under policy shifts. Bidirectional TD provides a practical approach to learning the paired features from reward-free data. Experiments support their utility for zero-shot policy extraction across diverse reward tasks. The results highlight the value of separating directly computable reward information from learned temporal effects, while leaving improved estimation of weak spectral modes and reliable reuse over larger policy changes as directions for future work.

\subsection*{AI use statement}

In this work, we used generative AI tools for
(i) language editing, i.e., grammar and clarity revisions of author-written text;
(ii) literature search, to help identify potentially relevant prior work;
(iii) coding and formatting assistance, including plotting scripts for figures and LaTeX formatting of equations;
(iv) drafting sections of the paper, with all AI-generated drafts treated as starting points and subsequently reviewed, edited, and verified by the authors

We have not used generative AI tools to generate, modify, or select experimental data or results, or to fabricate citations.

We have reviewed all AI-assisted work. Every reference surfaced with AI assistance was located and checked against the original source by the authors before inclusion. AI-suggested text edits were reviewed and revised by the authors. AI-assisted plotting code was reviewed by all authors, and each figure was checked against the raw exported results. We take responsibility for the final content of this work, including text, claims, and artifacts produced with the aid of generative AI.

\subsection*{Reproducibility statement}

To facilitate reproducibility, we provide a complete account of the assumptions and proofs underlying our theoretical results in Appendix~\ref{app:assumptions} and Appendix~\ref{app:proofs}. Detailed experimental settings, including dataset preprocessing, model configurations, training procedures, evaluation protocols, and hyperparameters, are provided in Appendix~\ref{app:exp_details}. An anonymized implementation of our method, together with the scripts used to reproduce the main experiments, is available at \url{https://anonymous.4open.science/r/suaf-C01C}. We refer readers to these materials for the exact details needed to reproduce our results.







\bibliography{iclr2027_conference}
\bibliographystyle{iclr2027_conference}
\newpage

\appendix


\section{Related Works and Discussion}

\subsection{Related Works}

\noindent\textbf{Successor features and reward factorization.}
Classical successor features~\cite{barreto2017successor,borsa2018universal} model $r_w=\phi^\text{T} w$, yielding $Q_{r_w}^\pi=(\psi^\pi)^\text{T} w$, with $\psi^\pi$ accumulating discounted features.
SUAF instead derives $g_i$ from delayed credit, with readouts $\beta_i^{+,\pi}(r)=\langle g_i,r\rangle_{\nu^{+,\pi}}$.
The residual $r_\perp=r-\sum_{i=1}^{R}\beta_i^{+,\pi}(r)g_i$, orthogonal to all $R$ nonzero modes, contributes no delayed advantage, while $C_\pi r$ remains its immediate effect.
Truncating to $k<R$ can discard decision-relevant modes, with error $\|A_r^\pi-A_{r,k}^\pi\|_{d^\pi}^2=(\frac{\gamma}{1-\gamma})^2\sum_{i=k+1}^{R}(\sigma_i^{+,\pi})^2[\beta_i^{+,\pi}(r)]^2$.
Therefore, SUAF optimizes advantage reconstruction, not reward reconstruction.

\noindent\textbf{Optimal successor features.}
\citet{ollivier2025featuresbestsuccessorfeatures} characterizes the leading right singular functions of $\mathcal A_\pi=C_\pi(I-\gamma P_\pi)^{-1}$ through asymptotic downstream performance under strong behavior-cloning regularization, without providing a learning algorithm.
If under matching stationary measures and $L^2$ geometry, the full DCDM has the same right singular subspaces.
However, our $g_i$ instead diagonalize $(\mathcal D_\pi^+)^*\mathcal D_\pi^+$, where $\mathcal D_\pi^+=\mathcal A_\pi-C_\pi=\gamma C_\pi P_\pi(I-\gamma P_\pi)^{-1}$, and generally differ.
In doing this, we are able to get a possibly close to $0$ approximation loss to $\gA_\pi$, which enables the control of policy gradient error. See detailed discussions in Appendix \ref{sec:immediate_spectral_floor}.
Our rank-$k$ optimality is exact for the fixed-policy delayed operator.
We jointly estimate paired action-credit and reward features using Soft-HGR~\cite{wang2019efficient} and bidirectional TD. 
We emphasize that \(C_\pi r=0\) on the subspace of state-only rewards does not change the full-space operator underlying SUAF. SUAF decomposes the action-centered positive-lag DCDM, whereas the full successor measure, the full advantage operator, and the one-step transition kernel define different spectral problems and generally yield different features.

\noindent\textbf{Proto successor measures and scope.}
Proto Successor Measures~\cite{pmlr-v267-agarwal25e} represent feasible successor measures across policies in a policy-independent affine basis, $M^\pi=b+\sum_\ell w_\ell^\pi\Phi_\ell$.
For $0<\gamma<1$, $M^\pi=(1-\gamma)I+\gamma M^{+,\pi}$ links PSM to $\widetilde{\boldsymbol B}^{+,\pi}=D_{d^\pi}^{1/2}C_\pi M^{+,\pi}D_{\nu^{+,\pi}}^{-1/2}$, where $D_\mu=\operatorname{diag}(\mu)$.
It removes action-invariant occupancy and instead encodes rewards through $\beta^{+,\pi}(r)$ at fixed $\pi$.
Exact PSM determines DCDM, but truncated DCDM cannot generally reconstruct the full successor measure.
Policy-dependent centering and marginals prevent automatically inheriting PSM's cross-policy affine structure.
Our scope is fixed-policy reward reuse, not policy-independent optimal control. For zero-shot learning with SUAF, please refer to Appendix \ref{app:suaf_oracle_control}.

\noindent\textbf{Spectral Reinforcement Learning.}
Existing spectral representations differ in the operator whose structure they retain (Table~\ref{tab:operator_comparison}). Gao et al.~\cite{gao2025spectral} study spectral representations of the one-step transition operator, while Diff-SR~\cite{shribak2024diffusion} extracts transition-based features through diffusion and energy-based models. FaStR~\cite{wu2026factorized} further factorizes the transition kernel across states, actions, and next states. Moving beyond one-step dynamics, One-Step FB~\cite{zheng2026can} represents the successor measure of a fixed behavior policy for downstream policy improvement. Ollivier~\cite{ollivier2025featuresbestsuccessorfeatures} selects successor features using a spectral criterion derived from downstream performance under strong behavior-cloning regularization. SUAF instead applies action centering to the positive-lag successor measure before spectral decomposition. This yields a reward-independent representation of \emph{delayed advantage}, while the immediate reward contrast is computed exactly; the rank budget is therefore devoted to the component that must be approximated.

\begin{table}[t]
\centering
\caption{\textbf{Comparison of the operators underlying related representations.} Transition-based methods represent one-step dynamics. SUAF decomposes the action-centered positive-lag successor operator with immediate reward.}
\label{tab:operator_comparison}
\footnotesize
\begin{tabularx}{\linewidth}{@{}lX@{}}
\toprule
Representation & Represented or decomposed object \\
\midrule
Gao et al. \cite{gao2025spectral}
& One-step transition kernel $P(s'| s,a)$; spectral factorization. \\
Diff-SR \cite{shribak2024diffusion}
& One-step $P(s'| s,a)$; diffusion/energy-based spectral features. \\
FaStR \cite{wu2026factorized}
& Transition tensor:
  $P(s'| s,a)\!\!\approx\!\!
  \sum_i \phi_i^S(s)\phi_i^A(a)m_i(s')$ (CP factorization). \\
One-Step FB \cite{zheng2026can}
& Successor-measure ratio $M^{\pi_\beta}(y| x)/\rho(y)$
  for a fixed behavior policy. \\
Ollivier \cite{ollivier2025featuresbestsuccessorfeatures}
& Spectral criterion based on downstream performance;
  involves the full advantage operator
  $C^\pi(I-\gamma P^\pi)^{-1}$ \\
\textbf{SUAF (ours)}
& $\operatorname{svd}\!\left(
  D_{d^\pi}^{1/2}C^\pi M^{+,\pi}
  D_{\nu^{+,\pi}}^{-1/2}\right)$;
  $A_r^\pi=C^\pi r+
  \frac{\gamma}{1-\gamma}C^\pi M^{+,\pi}r$. \\
\bottomrule
\end{tabularx}
\end{table}

\subsection{Difference and relations between SUAF and existing operators}

\begin{figure}
    \centering
    \includegraphics[width=\linewidth]{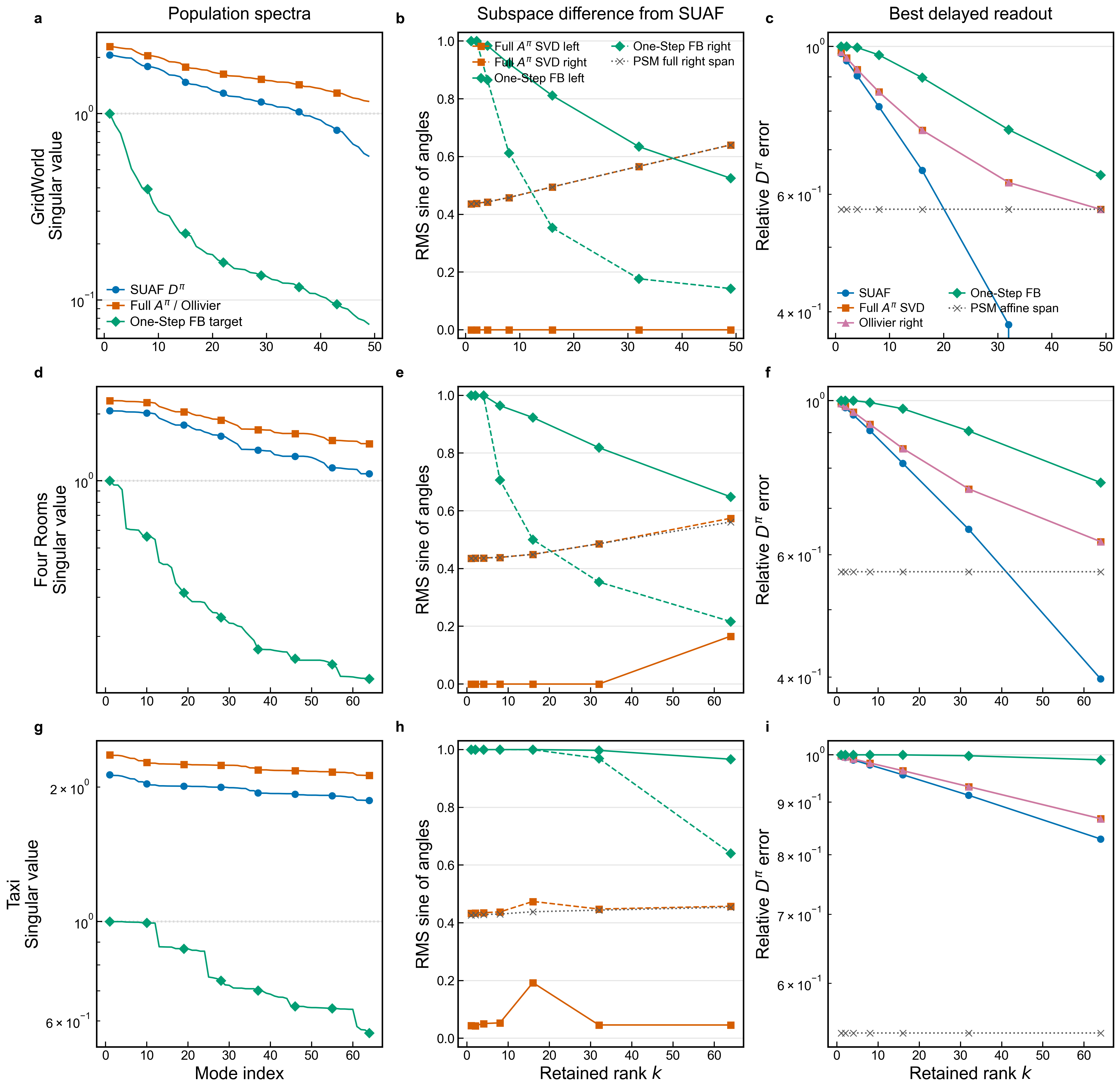}
    \caption{\textbf{The population feature subspaces induced by SUAF, the full advantage operator, One-Step FB, Ollivier, and PSM.} Since all the space are solved explicitly, hence no seeds are needed here.}
    \label{fig:spectral_subspace}
\end{figure}

Figure~\ref{fig:spectral_subspace} compares the population feature subspaces induced by SUAF, the full advantage operator, One-Step FB, Ollivier, and PSM. All operators are constructed exactly under the same reference policy, so the differences do not arise from optimization noise.

The left column shows the singular spectra. For SUAF, we plot the singular values of the delayed contribution
\(
D^\pi:=\frac{\gamma}{1-\gamma}\mathcal T^{+,\pi},
\)
whereas the full-advantage curve is obtained from
\(
A^\pi=C_\pi+D^\pi.
\)
The dotted horizontal line marks the unit spectral floor contributed by the immediate action-centering operator \(C_\pi\). In GridWorld and Four Rooms, it recovers
\[
\sigma_i(A^\pi)
=
\sqrt{1+\sigma_i(D^\pi)^2}.
\]
Consequently, the spectrum of the full advantage cannot decay below one even when the delayed singular values are small. Directly truncating \(A^\pi\) therefore spends rank on the analytically known immediate term, whereas SUAF assigns its entire rank budget to delayed credit assignment. The One-Step FB spectrum is that of the uncentered positive-lag successor kernel; its absolute scale depends on the successor-measure normalization, so this curve is primarily used to compare spectral decay rather than magnitude.

The middle column measures the difference between feature subspaces. Let
\(U_k^{D},V_k^{D}\) denote the leading SUAF left and right singular
vectors and let \(U_k^{m},V_k^{m}\) denote those of method \(m\). We compute
the principal angles through
\[
\cos\theta_i^{L}
=
\sigma_i\!\left((U_k^{D})^\top U_k^{m}\right),
\qquad
\cos\theta_i^{R}
=
\sigma_i\!\left((V_k^{D})^\top V_k^{m}\right),
\]
and report
\[
\operatorname{RMS-sin}
=
\left(\frac{1}{k}\sum_{i=1}^{k}\sin^2\theta_i\right)^{1/2}.
\]
A value of zero means that the two subspaces coincide, whereas a value
near one means that they are nearly orthogonal. Although the left
singular subspaces of \(D^\pi\) and \(A^\pi\) coincide in the stationary
GridWorld and Four Rooms constructions, their right subspaces differ
substantially. At \(k=8\), the right-subspace RMS sine is \(0.458\) in
GridWorld, \(0.439\) in Four Rooms, and \(0.438\) in Taxi. Since the
features proposed by Ollivier are the right singular functions of
\(A^\pi\), the Ollivier and full-\(A^\pi\) right subspaces are identical
by construction.

The discrepancy with One-Step FB is larger. One-Step FB factorizes an
uncentered successor measure and therefore retains future information
that is predictable from the current state alone. SUAF instead subtracts
this state-predictable component and retains only the incremental effect
of the current action. Accordingly, at \(k=8\), the right-subspace RMS
sine between SUAF and One-Step FB is \(0.612\), \(0.707\), and \(1.000\)
in the three environments, respectively. These principal angles provide
direct spectral evidence that SUAF does not merely learn a rotated
version of the One-Step FB representation.

The right column evaluates how well each subspace reconstructs the
delayed operator. For a method providing left and right subspaces, we
report
\[
\operatorname{Err}_{D}^{m}(k)
=
\frac{
\left\|
D^\pi-
P_{U_k^{m}}D^\pi P_{V_k^{m}}
\right\|_{\mathrm F}}
{\|D^\pi\|_{\mathrm F}},
\]
where \(P_U=UU^\top\). For Ollivier, only the right subspace is prescribed,
so we use the best unrestricted left readout,
\[
\operatorname{Err}_{D}^{\mathrm{Oll}}(k)
=
\frac{\|D^\pi-D^\pi P_{V_k^{A}}\|_{\mathrm F}}
{\|D^\pi\|_{\mathrm F}}.
\]
For SUAF, this error reduces to
\[
\operatorname{Err}_{D}^{\mathrm{SUAF}}(k)
=
\left(
\frac{\sum_{i>k}\sigma_i(D^\pi)^2}
     {\sum_i\sigma_i(D^\pi)^2}
\right)^{1/2},
\]
which is the Eckart-Young optimal rank-\(k\) error. SUAF consequently
achieves the lowest delayed-operator error among all rank-matched methods
at every evaluated rank. For example, at \(k=8\), the errors of
SUAF, full-\(A^\pi\), and One-Step FB are respectively
\(0.812/0.855/0.971\) in GridWorld,
\(0.906/0.925/0.994\) in Four Rooms, and
\(0.978/0.982/1.000\) in Taxi.

The PSM curve is not a rank-matched baseline. It projects onto the
complete policy-independent affine span, whose dimensions are \(200\),
\(416\), and \(2600\) in the three environments. It is therefore shown
only as a containment reference and should not be interpreted as
outperforming a rank-\(k\) representation.

\subsection{Properties of DCDM}
\label{sec:preperties_dcdm}

\subsubsection{Rank bound of the DCDM}
For a finite MDP, the positive-lag future advantage operator satisfies
$R:=\operatorname{rank}(\mathcal T^{+,\pi})\leq|\mathcal S|-1$,
regardless of the number of actions.
To see this, define the policy-averaged reward
$\bar r_\pi(s):=\sum_a\pi(a\mid s)r(s,a)$.
Since all future actions are sampled from $\pi$, the future term
depends on $r$ only through $\bar r_\pi$.
More explicitly, let
$(Tv)(s,a):=\sum_{s'}T(s'\mid s,a)v(s')$
and let $P_S^\pi$ be the state transition operator under $\pi$.
Then
\[
    \mathcal T^{+,\pi}r
    =
    (1-\gamma)C_\pi T
    (I-\gamma P_S^\pi)^{-1}\bar r_\pi.
\]
Thus, $\mathcal T^{+,\pi}$ factors through the
$|\mathcal S|$-dimensional space of state functions.
Moreover, a constant $\bar r_\pi$ produces no action contrast,
since $(I-\gamma P_S^\pi)^{-1}\mathbf 1
=(1-\gamma)^{-1}\mathbf 1$ and $C_\pi T\mathbf 1=0$.
Consequently, at most $|\mathcal S|-1$ nonzero SUAF pairs
are needed to represent the future advantage term exactly
for every reward.

\subsubsection{Information Background of DCDM}
We show that the property $\left\|
\widetilde{\boldsymbol B}^{+,\pi}
\right\|_{\mathrm F}^2
\approx 2I(A;S^+,A^+|S)$ holds under the weak dependency condition. For simplicity, denote 
$\delta(s^+,a^+,s,a)=\frac{M^{+,\pi}(s^+,a^+\mid s,a)}{\nu^{+,\pi}(s^+,a^+)}-1,$
i.e., $M^{+,\pi}(s^+,a^+\mid s,a)=\nu^{+,\pi}(s^+,a^+)(1+\delta(s^+,a^+,s,a)).$
It follows that $$\bar M^{+,\pi}(s^+,a^+\mid s)=\E_{a^\prime\sim\pi(\cdot|s)}\left[M^{+,\pi}(s^+,a^+|s,a^\prime)\right]=\nu^{+,\pi}(s^+,a^+)(1+\bar\delta(s^+,a^+,s)),$$
where $\bar\delta(s^+,a^+,s)=\sum_{a^\prime}\pi(a^\prime\mid s)\delta(s^+,a^+,s,a^\prime)$. 
We also have that 
\begin{align}
    \sum_{s^+,a^+,s,a}d^\pi(s,a)\nu^{+,\pi}(s^+,a^+)\delta(s^+,a^+,s,a)=0,\label{eq:delta_mean_zero_1}\\\sum_{s^+,a^+,s}d^\pi(s,a)\nu^{+,\pi}(s^+,a^+)\delta(s^+,a^+,s)=0.\label{eq:delta_mean_zero_2}
\end{align}
Assume that $\vert\delta(s^+,a^+,s,a)\vert<\epsilon, \forall s^+,a^+,s,a$ follows from the weak dependency condition, then $\vert\bar\delta(s^+,a^+,s)\vert<\epsilon, \forall s^+,a^+,s$. Computation gives that $I(A;S^+,A^+\mid S)=\frac12\left\|
\widetilde{\boldsymbol B}^{+,\pi}
\right\|_{\mathrm F}^2+O(\epsilon^3)$ as follows:
\begin{align}
    &\qquad I(A;S^+,A^+\mid S)\nonumber\\
    &=\sum_{s^+,a^+,s,a}d^\pi(s,a)M^{+,\pi}(s^+,a^+\mid s,a)\ln\frac{M^{+,\pi}(s^+,a^+\mid s,a)}{\bar M^{+,\pi}(s^+,a^+\mid s)}\\
    &=\sum_{s^+,a^+,s,a}d^\pi(s,a)\nu^{+,\pi}(s^+,a^+)(1+\delta(s^+,a^+,s,a))\ln\frac{1+\delta(s^+,a^+,s,a)}{1+\bar\delta(s^+,a^+,s)}\\
    &=\sum_{s^+,a^+,s,a}d^\pi(s,a)\nu^{+,\pi}(s^+,a^+)\cdot\nonumber\\
    &\quad\left(\delta(s^+,a^+,s,a)-\bar\delta(s^+,a^+,s)+\frac12(\delta(s^+,a^+,s,a)-\bar\delta(s^+,a^+,s))^2+O\left(\epsilon^3\right)\right)\label{eq:mut_expansion}\\
    &=\sum_{s^+,a^+,s,a}d^\pi(s,a)\nu^{+,\pi}(s^+,a^+)\cdot\frac12(\delta(s^+,a^+,s,a)-\bar\delta(s^+,a^+,s))^2+O\left(\epsilon^3\right)\label{eq:chi2_expansion}\\
    &=\frac12\sum_{s^+,a^+,s,a}d^\pi(s,a)\frac{(M^{+\pi}(s^+,a^+\mid s,a)-\bar M^{+\pi}(s^+,a^+\mid s))^2}{\nu^{+,\pi}(s^+,a^+)}+O\left(\epsilon^3\right)\\
    &=\frac12\left\|\widetilde{\boldsymbol B}^{+,\pi}\right\|_{\mathrm F}^2+O(\epsilon^3),
\end{align}
where from to \eqref{eq:mut_expansion} to \eqref{eq:chi2_expansion} we have used \eqref{eq:delta_mean_zero_1} and \eqref{eq:delta_mean_zero_2}.

\subsection{Properties of the proposed SUAF}
\label{sec:preperties_suaf}

The proposed SUAF satisfies the following properties: (1) \textbf{Weighted orthonormality.} For all $i,j\leq R$, $\E_{d^\pi}[f_if_j]=\E_{\nu^{+,\pi}}[g_ig_j]=\bm{1}(i=j)$; (2) \textbf{Centering.} $\E_{\pi(\cdot|s)}[f_i]=\mathbb{E}_{\nu^{+,\pi}}[g]=0$; (3) \textbf{Diagonal canonical dependence.} $\mathbb E_{(s,a)\sim d^\pi,(s^+,a^+)\sim M^{+,\pi}(\cdot\mid s,a)}\left[f_i(s,a)g_j(s^+,a^+)\right]=\sigma_i^{+,\pi}\bm{1}(i=j)$; (4) \textbf{Variational optimality.} The first feature pair solves $\sigma_1^{+,\pi} = \max_{f,g} \E\left[f(s,a)g(s^+,a^+)\right]$, subject to $\E_\pi[f(s,a)|s]=\E_{\nu^{+,\pi}}[g]=0, \|f\|_{d^\pi}=\|g\|_{\nu^{+,\pi}}=1$. Each subsequent pair $(f_i,g_i)$ solves the same problem subject additionally to weighted orthogonality with all preceding feature pairs. Furthermore, since both feature families are centered and normalized, $\sigma_i^{+,\pi}$ is the correlation carried by the $i$-th feature pair, and hence $0\leq\sigma_i^{+,\pi}\leq1$.

\subsection{Reward Shaping in the Sense of SUAF}
\label{sec:reward_shaping}

The accuracy for an individual reward depends on the weighted spectral coefficients:
\begin{equation}
    \left\|A_r^\pi-A_{r,k}^\pi\right\|_{d^\pi}^2
    =
    \left(\frac{\gamma}{1-\gamma}\right)^2
    \sum_{i=k+1}^{R}
    \left(\sigma_i^{+,\pi}\right)^2
    \left|\langle g_i,r\rangle_{\nu^{+,\pi}}\right|^2.
\end{equation}
Consequently, rewards with substantial energy in the discarded modes can be poorly approximated. To address this problem, we consider potential-based reward shaping that redistributes credit between the exact immediate term and the approximated future term to reduce this error while preserving the exact advantage function. For a bounded state potential $\Phi$, define
\begin{equation}
    r_\Phi(s,a,s')
    =
    r(s,a)+\gamma\Phi(s')-\Phi(s),
\end{equation}
In the discounted infinite-horizon setting, this transformation gives $Q_{r_\Phi}^\pi(s,a)=Q_r^\pi(s,a)-\Phi(s)$ and $V_{r_\Phi}^\pi(s)=V_r^\pi(s)-\Phi(s)$, hence $A_{r_\Phi}^\pi=A_r^\pi$. To apply the state-action reward representation, let $\bar r_\Phi = r + \gamma T\Phi - \Phi$, where $(T\Phi)(s,a)=\E[\Phi(s^\prime)\mid s,a]$.
Let $\Pi_k$ denote the $L^2(d^\pi)$-orthogonal projection onto $\operatorname{span}\{f_1,\ldots,f_k\}$.

\begin{theorem}[Potential-dependent approximation error]
\label{thm:potential_error}
Let $0<\gamma<1$ and let $r$ be bounded. For $i=k+1,\ldots,R$, define the tail vectors
$
b_i=\sigma_i\langle g_i,r\rangle_\nu,
$
and
$
z_i(\Phi)=
\sigma_i\langle g_i,\gamma T\Phi-\Phi\rangle_\nu.
$
Let $c=\gamma/(1-\gamma)$. The approximation
$A_{\bar r_\Phi,k}^\pi
=C_\pi\bar r_\Phi+c\Pi_k\gT^{+,\pi}\bar r_\Phi$
satisfies
\begin{align}
E_k(\Phi_w;r):=\|A_r^\pi-A_{\bar r_\Phi,k}^\pi\|_{d^\pi}^2=c^2\|b+z(\Phi)\|_2^2
=\gamma^2
\|(I-\Pi_k)C_\pi T(V_r^\pi-\Phi)\|_{d^\pi}^2.
\label{eq:shaping_value_error}
\end{align}
Consequently, shaping does not increase the error if and only if
$
2\langle b,z(\Phi)\rangle+\|z(\Phi)\|_2^2\leq 0.
$
\end{theorem}
Therefore, potential quality is determined by the unremained action-credit component of its value error, rather than by ordinary value prediction error alone.
\begin{corollary}[Optimal linear combination of potentials]
\label{cor:optimal_potential}
Fix bounded state features $h_1,\ldots,h_p$ and set
$\Phi_w=\sum_{\ell=1}^p w_\ell h_\ell$.
Let $D=[z(h_1),\ldots,z(h_p)]$.
A minimum-norm optimal coefficient vector and the resulting
error are
$$
w^\star=-D^\dagger b,
\quad
\min_w E_k(\Phi_w;r)
=c^2\|(I-DD^\dagger)b\|_2^2.
$$
In particular, zero error is attainable within this class
if and only if $b\in\operatorname{col}(D)$.
\end{corollary}
Without restricting the potential class, the oracle choice $\Phi=V_r^\pi$ yields zero error, although obtaining this potential requires policy evaluation and will cancel out all the inner product terms $\langle g_i,r_\Phi\rangle$. In contrast, we propose that move the reward to the head of spectrum also help SUAF algorithms estimate advantage better. For a specific example, please refer to \ref{sec:taxi_reward_shaping_exp}.

\section{Extra Experiments}

\subsection{Full result}

In this subsection, we report our complete results, including the mean and statistical results (e.g., 95\% confidence intervals) across all tasks in both discrete and continuous environments. The results are summarized in Table \ref{tab:ood_full} and Table \ref{tab:zero_shot_full}.

\begin{table*}[t]
\centering
\caption{\textbf{Zero-shot evaluation on finite state environments.} The performance is measured using the improvement rate defined as $\frac{J(\pi) - J(\pi_\beta)}{J(\pi^\ast) - J(\pi_\beta)}$, where $\pi^\ast$ is the optimal policy solved by linear program, $\pi$ is the generated policy and $\pi_\beta$ is the sample policy. Each result is measured as \textit{mean} with \textit{95\% confidence interval} and reports with 5 random sample policies with 5 seeds for each policy.}
\label{tab:ood_full}
\resizebox{\textwidth}{!}{
\begin{tabular}{ll|ccccccc|c}
\toprule
Domain & Task & Laplacian & BYOL-\(\gamma\) & TD-JEPA & ICVF & HILP & FB & One-Step FB & Ours \\
\midrule
\rowcolor{gray!15} \multicolumn{10}{l}{\textbf{In-Distribution Tasks}} \\
\multirow{5}{*}{Grid World} & \textbf{overall} 
& $0.006_{\pm0.030}$ 
& $0.690_{\pm0.113}$ 
& {$\mathbf{0.993}_{\pm0.015}$} 
& $0.804_{\pm0.014}$ 
& $0.770_{\pm0.040}$ 
& \underline{$0.929_{\pm0.048}$} 
& $0.599_{\pm0.036}$ 
& $0.768_{\pm0.040}$ \\

& top-left 
& $-0.016_{\pm0.124}$ 
& $0.967_{\pm0.033}$ 
& $0.984_{\pm0.045}$ 
& {$\mathbf{1.000}_{\pm0.000}$} 
& {$\mathbf{1.000}_{\pm0.000}$} 
& \underline{$0.979_{\pm0.038}$} 
& $0.767_{\pm0.126}$ 
& $0.893_{\pm0.057}$ \\

& top-right 
& $-0.002_{\pm0.052}$ 
& $0.060_{\pm0.143}$ 
& \underline{$0.998_{\pm0.006}$} 
& {$\mathbf{1.000}_{\pm0.000}$} 
& $0.931_{\pm0.186}$ 
& $0.868_{\pm0.095}$ 
& $0.517_{\pm0.033}$ 
& $0.695_{\pm0.023}$ \\

& center 
& $-0.001_{\pm0.069}$ 
& $0.734_{\pm0.438}$ 
& $0.994_{\pm0.016}$ 
& \underline{$0.997_{\pm0.009}$} 
& {$\mathbf{1.000}_{\pm0.000}$} 
& $0.895_{\pm0.153}$ 
& $0.760_{\pm0.051}$ 
& $0.849_{\pm0.040}$ \\

& bottom-right 
& $0.042_{\pm0.103}$ 
& {$\mathbf{1.000}_{\pm0.000}$} 
& \underline{$0.996_{\pm0.012}$} 
& $0.218_{\pm0.060}$ 
& $0.150_{\pm0.030}$ 
& $0.974_{\pm0.045}$ 
& $0.355_{\pm0.063}$ 
& $0.635_{\pm0.178}$ \\ 
\midrule
\multirow{5}{*}{Four-room} & \textbf{overall}
& $0.000_{\pm0.007}$
& $0.042_{\pm0.016}$
& {$\mathbf{0.900}_{\pm0.063}$}
& $0.293_{\pm0.040}$
& $0.450_{\pm0.167}$
& \underline{$0.773_{\pm0.110}$}
& $0.175_{\pm0.102}$
& $0.510_{\pm0.008}$ \\

& top-left
& $-0.003_{\pm0.010}$
& $0.042_{\pm0.028}$
& {$\mathbf{0.919}_{\pm0.045}$}
& $0.278_{\pm0.206}$
& {$0.642_{\pm0.207}$}
& $0.508_{\pm0.006}$
& $0.022_{\pm0.003}$
& \underline{$0.671_{\pm0.031}$} \\

& top-right
& $-0.001_{\pm0.010}$
& $0.056_{\pm0.012}$
& {$\mathbf{0.962}_{\pm0.037}$}
& $0.261_{\pm0.096}$
& $0.236_{\pm0.155}$
& \underline{$0.961_{\pm0.042}$}
& $0.071_{\pm0.149}$
& $0.415_{\pm0.000}$ \\

& bottom-left
& $0.003_{\pm0.016}$
& $0.054_{\pm0.072}$
& \underline{$0.825_{\pm0.220}$}
& $0.348_{\pm0.150}$
& $0.582_{\pm0.468}$
& {$\mathbf{0.849}_{\pm0.212}$}
& $0.250_{\pm0.260}$
& $0.528_{\pm0.011}$ \\

& bottom-right
& $0.000_{\pm0.010}$
& $0.018_{\pm0.037}$
& {$\mathbf{0.894}_{\pm0.040}$}
& $0.286_{\pm0.146}$
& $0.340_{\pm0.227}$
& \underline{$0.772_{\pm0.324}$}
& $0.357_{\pm0.063}$
& $0.425_{\pm0.000}$ \\
\midrule
\multirow{5}{*}{Taxi} & \textbf{overall}
& $0.029_{\pm0.256}$
& $0.064_{\pm0.009}$
& $-0.037_{\pm0.085}$
& \underline{$0.071_{\pm0.016}$}
& $0.042_{\pm0.113}$
& $-0.226_{\pm0.015}$
& $0.067_{\pm0.040}$
& {$\mathbf{0.264}_{\pm0.018}$} \\

& delivery\footnotemark[1]
& {$0.000_{\pm0.000}$}
& $0.000_{\pm0.000}$
& $0.000_{\pm0.000}$
& $0.000_{\pm0.000}$
& $0.000_{\pm0.000}$
& $0.000_{\pm0.000}$
& $0.000_{\pm0.000}$
& {$0.000_{\pm0.000}$} \\

& fast delivery\footnotemark[2]
& {$0.000_{\pm0.000}$}
& $0.000_{\pm0.000}$
& $0.000_{\pm0.000}$
& $0.000_{\pm0.000}$
& $0.000_{\pm0.000}$
& {$0.000_{\pm0.000}$}
& $0.000_{\pm0.000}$
& $0.000_{\pm0.000}$ \\

& safe delivery
& $0.128_{\pm1.033}$
& $0.271_{\pm0.035}$
& $-0.158_{\pm0.323}$
& \underline{$0.297_{\pm0.067}$}
& $0.179_{\pm0.459}$
& $-0.904_{\pm0.051}$
& $0.262_{\pm0.158}$
& {$\mathbf{0.453}_{\pm0.000}$} \\

& staged service
& $-0.012_{\pm0.009}$
& $-0.015_{\pm0.000}$
& \underline{$0.010_{\pm0.033}$}
& $-0.012_{\pm0.004}$
& $-0.012_{\pm0.008}$
& $0.000_{\pm0.007}$
& $0.007_{\pm0.016}$
& {$\mathbf{0.605}_{\pm0.073}$} \\
\midrule
\rowcolor{gray!15} \multicolumn{10}{l}{\textbf{OOD Tasks}} \\
\multirow{3}{*}{Grid World} & \textbf{overall}
& $0.061_{\pm0.059}$
& $0.443_{\pm0.055}$
& $0.446_{\pm0.061}$
& $0.437_{\pm0.021}$
& $0.405_{\pm0.044}$
& \underline{$0.474_{\pm0.087}$}
& $0.424_{\pm0.038}$
& {$\mathbf{0.873}_{\pm0.016}$} \\

& state composition
& $0.078_{\pm0.069}$
& {$\mathbf{0.923}_{\pm0.061}$}
& $0.839_{\pm0.110}$
& $0.811_{\pm0.043}$
& $0.743_{\pm0.104}$
& \underline{$0.890_{\pm0.106}$}
& $0.777_{\pm0.044}$
& $0.771_{\pm0.040}$ \\

& state-action circulation
& $0.045_{\pm0.074}$
& $-0.036_{\pm0.060}$
& $0.053_{\pm0.031}$
& $0.063_{\pm0.020}$
& $0.067_{\pm0.028}$
& $0.058_{\pm0.073}$
& \underline{$0.072_{\pm0.041}$}
& $\mathbf{0.975}_{\pm0.021}$ \\
\midrule
\multirow{3}{*}{Four-room} & \textbf{overall}
& $0.070_{\pm0.028}$
& $0.169_{\pm0.103}$
& $0.347_{\pm0.020}$
& $0.283_{\pm0.085}$
& $0.150_{\pm0.082}$
& \underline{$0.385_{\pm0.073}$}
& $0.180_{\pm0.012}$
& {$\mathbf{0.636}_{\pm0.009}$} \\

& state composition
& $0.068_{\pm0.029}$
& $0.258_{\pm0.201}$
& {$0.602_{\pm0.045}$}
& $0.509_{\pm0.209}$
& $0.337_{\pm0.155}$
& {$\mathbf{0.660}_{\pm0.117}$}
& $0.321_{\pm0.026}$
& \underline{$0.640_{\pm0.017}$} \\

& state-action circulation
& $0.071_{\pm0.028}$
& $0.081_{\pm0.014}$
& $0.092_{\pm0.065}$
& $0.057_{\pm0.042}$
& $-0.037_{\pm0.049}$
& \underline{$0.111_{\pm0.068}$}
& $0.038_{\pm0.017}$
& {$\mathbf{0.632}_{\pm0.007}$} \\
\midrule
\multirow{4}{*}{Taxi} & \textbf{overall}
& $-0.014_{\pm0.111}$
& $-0.027_{\pm0.002}$
& $0.009_{\pm0.025}$
& $-0.026_{\pm0.015}$
& $-0.021_{\pm0.048}$
& \underline{$0.085_{\pm0.013}$}
& $-0.016_{\pm0.019}$
& {$\mathbf{0.300}_{\pm0.010}$} \\

& pickup priority
& $-0.012_{\pm0.009}$
& $-0.015_{\pm0.000}$
& $0.011_{\pm0.023}$
& $-0.012_{\pm0.002}$
& $-0.012_{\pm0.003}$
& $0.025_{\pm0.025}$
& \underline{$0.051_{\pm0.022}$}
& {$\mathbf{0.642}_{\pm0.029}$} \\

& center hazard
& {$\mathbf{0.005}_{\pm0.000}$}
& {$\mathbf{0.005}_{\pm0.000}$}
& $-0.025_{\pm0.041}$
& $\mathbf{0.005}_{\pm0.000}$
& \underline{$0.004_{\pm0.004}$}
& $-0.019_{\pm0.027}$
& $-0.032_{\pm0.023}$
& $\mathbf{0.005}_{\pm0.000}$ \\

& action fuel
& $-0.036_{\pm0.324}$
& $-0.072_{\pm0.007}$
& $0.041_{\pm0.056}$
& $-0.071_{\pm0.044}$
& $-0.054_{\pm0.140}$
& \underline{$0.248_{\pm0.013}$}
& $-0.066_{\pm0.045}$
& {$\mathbf{0.253}_{\pm0.001}$} \\
\midrule
\bottomrule
\end{tabular}
}
\vspace{1pt}
\begin{minipage}{\textwidth}
\footnotesize
\textsuperscript{1} {For solution to sparse reward, please refer to Appendix \ref{sec:taxi_reward_shaping_exp} with reward shaping. For fair comparison, we will not show the result here.}

\textsuperscript{2} {Fast delivery is the same with delivery environment.}
\end{minipage}
\end{table*}

\begin{table*}[htbp]
\centering
\caption{\textbf{Zero-shot evaluation on state-based ExORL and OGBench benchmarks.} The full zero-shot evaluation results on 8 ExORL tasks and 10 OGBench tasks. The results are reported as \textit{mean} over 8 seeds with \textit{std}.}
\label{tab:zero_shot_full}
\resizebox{\textwidth}{!}{
\begin{tabular}{l l|c c c c c c c | c}
\toprule
Domain & Task & Laplacian & BYOL-\(\gamma\) & TD-JEPA & ICVF & HILP & FB & One-Step FB & Ours \\
\midrule
\rowcolor{gray!15} \multicolumn{10}{l}{\textbf{ExORL}: Locomotion}\\
\multirow{5}{*}{walker} & overall
& 228 \(\pm\) 2 
& 227 \(\pm\) 2 
& 316 \(\pm\) 12 
& \textbf{619 \(\pm\) 23} 
& 393 \(\pm\) 108 
& 400 \(\pm\) 40 
& 379 \(\pm\) 26 
& \underline{445 \(\pm\) 39} \\

& flip 
& 243 \(\pm\) 5 
& 242 \(\pm\) 4 
& 381 \(\pm\) 34 
& \textbf{538 \(\pm\) 19} 
& 332 \(\pm\) 135 
& 277 \(\pm\) 100 
& \underline{388 \(\pm\) 30} 
& 357 \(\pm\) 42 \\

& run & 89 \(\pm\) 1 
& 89 \(\pm\) 2 
& 112 \(\pm\) 22 
& \textbf{258 \(\pm\) 20} 
& 136 \(\pm\) 36 
& 194 \(\pm\) 20 
& 198 \(\pm\) 30 
& \underline{210 \(\pm\) 20} \\

& stand 
& 389 \(\pm\) 3 
& 387 \(\pm\) 5 
& 387 \(\pm\) 11 
& \textbf{858 \(\pm\) 31} 
& \underline{691 \(\pm\) 126}
& 621 \(\pm\) 93 
& 524 \(\pm\) 60 
& 670 \(\pm\) 100 \\

& walk 
& 192 \(\pm\) 4 
& 192 \(\pm\) 6
& 384 \(\pm\) 47 
& \textbf{821 \(\pm\) 40} 
& 413 \(\pm\) 195 
& 506 \(\pm\) 34
& 407 \(\pm\) 63 
& \underline{542 \(\pm\) 41}\\

\midrule
\multirow{5}{*}{cheetah} 
& overall 
& 125 \(\pm\) 41 
& 127 \(\pm\) 39 
& 185 \(\pm\) 7 
& 187 \(\pm\) 13 
& 116 \(\pm\) 78 
& 271 \(\pm\) 46 
& \textbf{378 \(\pm\) 56} 
& \underline{375 \(\pm\) 21}\\

& run 
& 40 \(\pm\) 13 &
42 \(\pm\) 13 
& 66 \(\pm\) 3 
& {89 \(\pm\) 7} 
& 38 \(\pm\) 32 
& 43 \(\pm\) 36 
& \underline{97 \(\pm\) 19} 
& \textbf{100} \(\pm\) 12 \\

& run backward 
& 50 \(\pm\) 17 
& 48 \(\pm\) 15 
& 69 \(\pm\) 6 
& 48 \(\pm\) 9 
& 36 \(\pm\) 40 
& 125 \(\pm\) 30 
& \textbf{181 \(\pm\) 60} 
& \underline{159 \(\pm\) 70}\\

& walk 
& 185 \(\pm\) 61 
& 199 \(\pm\) 61 
& 293 \(\pm\) 15 
& 385 \(\pm\) 21 
& 195 \(\pm\) 121 
& 251 \(\pm\) 166 
& \textbf{424 \(\pm\) 44} 
& \underline{420 \(\pm\) 32}\\

& walk backward 
& 226 \(\pm\) 77 
& 220 \(\pm\) 70 
& 312 \(\pm\) 26 
& 226 \(\pm\) 38 
& 194 \(\pm\) 202 
& 663 \(\pm\) 143 
& \underline{811 \(\pm\) 156} 
& \textbf{820 \(\pm\) 232}\\
\midrule
\rowcolor{gray!15} \multicolumn{10}{l}{\textbf{OGBench}: Navigation and manipulation}\\
\multirow{6}{*}{antmaze large} 
& overall 
& 9 \(\pm\) 1 
& 21 \(\pm\) 2 
& 14 \(\pm\) 9 
& 23 \(\pm\) 3 
& \underline{34 \(\pm\) 2} 
& 25 \(\pm\) 5 
& 30 \(\pm\) 9 
& \textbf{35 \(\pm\) 5}\\

& task 1 
& 2 \(\pm\) 1 
& 6 \(\pm\) 3 
& 15 \(\pm\) 11 
& 4 \(\pm\) 2 
& 13 \(\pm\) 8 
& \textbf{46 \(\pm\) 9} 
& 21 \(\pm\) 9 
& \underline{36 \(\pm\) 8}\\

& task 2 
& 2 \(\pm\) 1 
& 11 \(\pm\) 4 
& 1 \(\pm\) 2 
& 9 \(\pm\) 3 
& {16 \(\pm\) 6} 
& 2 \(\pm\) 3 
& \textbf{41 \(\pm\) 12} 
& \underline{24 \(\pm\) 1}\\

& task 3 
& 29 \(\pm\) 3 
& 57 \(\pm\) 8 
& 35 \(\pm\) 23 
& \underline{67 \(\pm\) 8 }
& \textbf{75 \(\pm\) 6} 
& 31 \(\pm\) 10 
& 15 \(\pm\) 4 
& 57 \(\pm\) 5\\

& task 4 
& 6 \(\pm\) 2 
& 14 \(\pm\) 5 
& 7 \(\pm\) 7 
& 18 \(\pm\) 6 
& \underline{27 \(\pm\) 10}
& 3 \(\pm\) 2 
& \textbf{33 \(\pm\) 15} 
& 12 \(\pm\) 6 \\

& task 5 
& 6 \(\pm\) 3 
& 16 \(\pm\) 4 
& 11 \(\pm\) 8 
& 18 \(\pm\) 4 
& \underline{40 \(\pm\) 8 }
& \textbf{44 \(\pm\) 19 }
& 37 \(\pm\) 20 
& \textbf{44 \(\pm\) 9} \\

\midrule
\multirow{6}{*}{cube single play} 
& overall 
& 6 \(\pm\) 2 
& 13 \(\pm\) 2 
& 8 \(\pm\) 4 
& 13 \(\pm\) 2 
& \textbf{30 \(\pm\) 8} 
& 2 \(\pm\) 1 
& 3 \(\pm\) 2 
& \underline{21 \(\pm\) 10}\\

& task 1 
& 5 \(\pm\) 2 
& 12 \(\pm\) 4 
& 6 \(\pm\) 4 
& 13 \(\pm\) 2 
& \textbf{27 \(\pm\) 7} 
& 1 \(\pm\) 1 
& 3 \(\pm\) 4 
& \underline{20 \(\pm\) 7} \\

& task 2 
& 5 \(\pm\) 2 
& 13 \(\pm\) 3 
& 8 \(\pm\) 6 
& 13 \(\pm\) 3 
& \textbf{30 \(\pm\) 10} 
& 3 \(\pm\) 2 
& 4 \(\pm\) 4 
& \underline{17 \(\pm\) 12}\\

& task 3 
& 7 \(\pm\) 3 
& 13 \(\pm\) 4 
& 14 \(\pm\) 9 
& 14 \(\pm\) 4 
& \textbf{23 \(\pm\) 13} 
& 3 \(\pm\) 3 & 4 \(\pm\) 4 
& \underline{18 \(\pm\) 9} \\

& task 4 
& 6 \(\pm\) 2 
& 15 \(\pm\) 6 
& 6 \(\pm\) 4 
& 11 \(\pm\) 3 
& \textbf{37 \(\pm\) 19} 
& 2 \(\pm\) 2 
& 2 \(\pm\) 2 
& \underline{21 \(\pm\) 12}\\

& task 5 
& 5 \(\pm\) 3 
& 11 \(\pm\) 3 
& 5 \(\pm\) 4 
& 13 \(\pm\) 4 
& \textbf{30 \(\pm\) 22} 
& 1 \(\pm\) 2 
& 1 \(\pm\) 1 
& \underline{28 \(\pm\) 20} \\
\bottomrule
\end{tabular}
}
\end{table*}

\subsection{One-Step Policy Improvement with SUAF}
\label{app:one_step_control_protocol}

For reward $r$ and retained rank $k$, the learned SUAF critic is
\begin{equation}
 \widehat A_{r,k}^{\pi_0}
 =C_{\pi_0}r
 +\frac{\gamma}{1-\gamma}
   \sum_{i=1}^{k}\widehat\sigma_i\widehat f_i
   \langle\widehat g_i,r\rangle_{\widehat\nu}.
 \label{eq:appendix_rank_k_critic}
\end{equation}
The immediate contrast $C_{\pi_0}r$ is therefore retained exactly at every
rank, including $k=0$.  We perform exactly one policy update, without actor
iteration or feature refresh.  The Boltzmann policies are
\begin{equation}
 \pi_{k,\tau}(a\mid s)
 =\frac{\pi_0(a\mid s)
   \exp\{\tau\widehat A_{r,k}^{\pi_0}(s,a)\}}
  {\sum_b\pi_0(b\mid s)
   \exp\{\tau\widehat A_{r,k}^{\pi_0}(s,b)\}},
 \qquad \tau\in\{4,32,256\},
 \label{eq:appendix_boltzmann_update}
\end{equation}
and the greedy policy selects
$\arg\max_a\widehat A_{r,k}^{\pi_0}(s,a)$ with deterministic tie breaking.
Grid World is evaluated at $k=0,5,\ldots,45,49$ and Four-room at
$k=0,10,\ldots,100,103$.

All returns are computed exactly by solving the finite Bellman system from the
same initial law $\mu$:
\begin{equation}
 J_r(\pi)=\mu^\top(I-\gamma P_\pi)^{-1}r_\pi.
\end{equation}
The plotted quantity is the fraction of the available improvement achieved by
the one-step policy,
\begin{equation}
 \operatorname{AI}(\pi;r)
 =\frac{J_r(\pi)-J_r(\pi_0)}
 {J_r(\pi^*)-J_r(\pi_0)},
 \label{eq:appendix_achievable_improvement}
\end{equation}
where $\pi^*$ and $J_r(\pi^*)$ are obtained by exact value iteration.  Hence
$0$ denotes the reference-policy return and $1$ denotes the optimal return and the score is not clipped.

\paragraph{Ablation: Learned representations used in the rank and Boltzmann temperature sweeps.}
The displayed learned-SUAF curves use the reward-free checkpoints selected by
held-out operator error. Raw neural features are recalibrated on an independent positive-lag split using discounted feature-return calibration,
separate left/right ridge penalties equal to $10^{-3}$ times the mean
eigenvalue of the corresponding feature covariance,
and a small-matrix SVD.  The validation sweep selected zero additional
singular-value shrinkage for both checkpoints.  Checkpoint selection and
calibration are reward-free. Downstream rewards are first used only in the
linear coefficients $\langle\widehat g_i,r\rangle_{\widehat\nu}$ and the
subsequent one-step update.

\begin{figure}
    \centering
    \includegraphics[width=\linewidth]{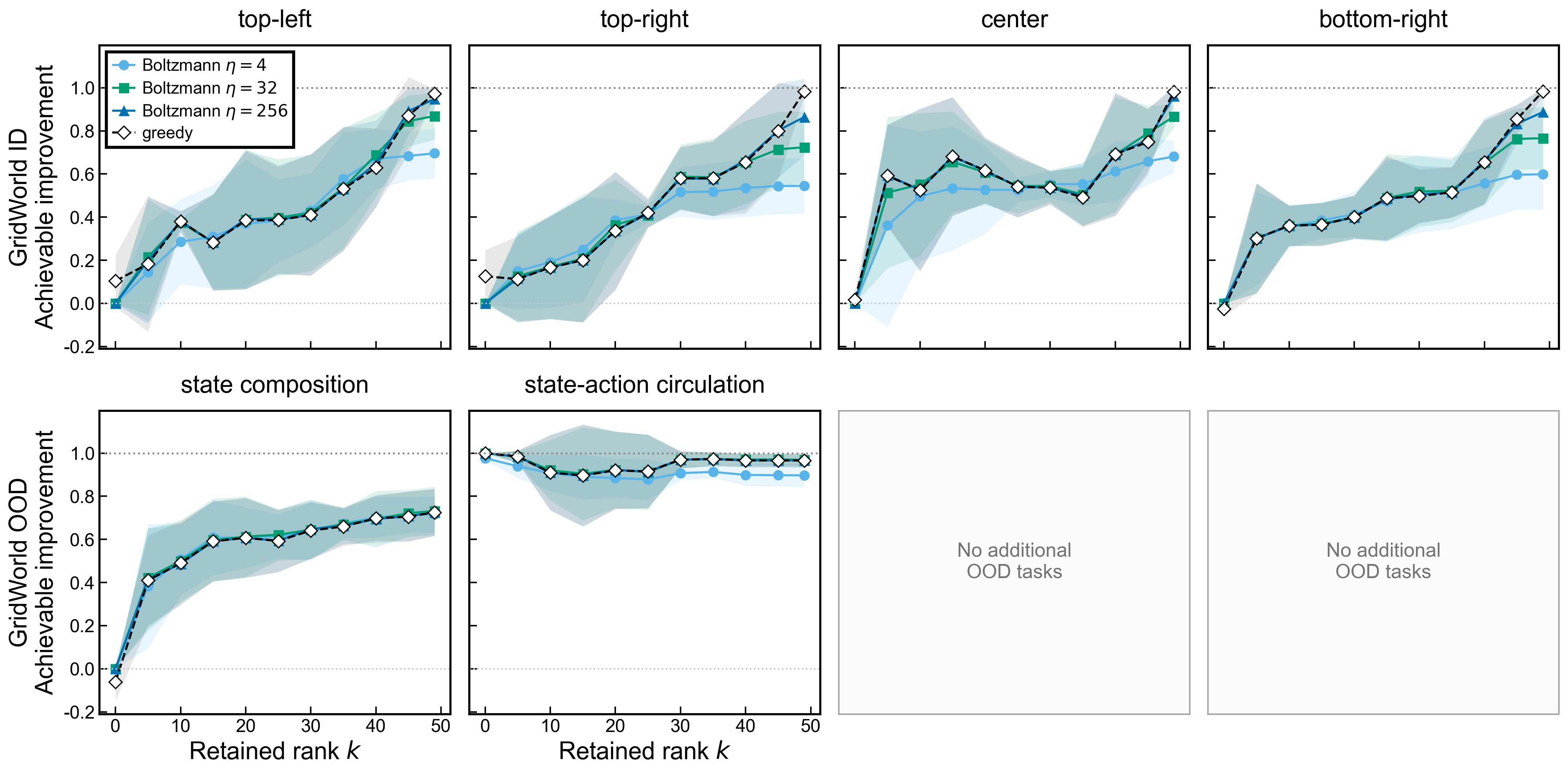}
    \caption{\textbf{One step policy update with exact SUAF in Grid World environment.} The results is calculated with 5 full-coverage random policies with 95\% CI. This also works as an ablation of Boltzmann update with different temperature.}
    \label{fig:gridworld_exact_suaf}
\end{figure}
\begin{figure}
    \centering
    \includegraphics[width=\linewidth]{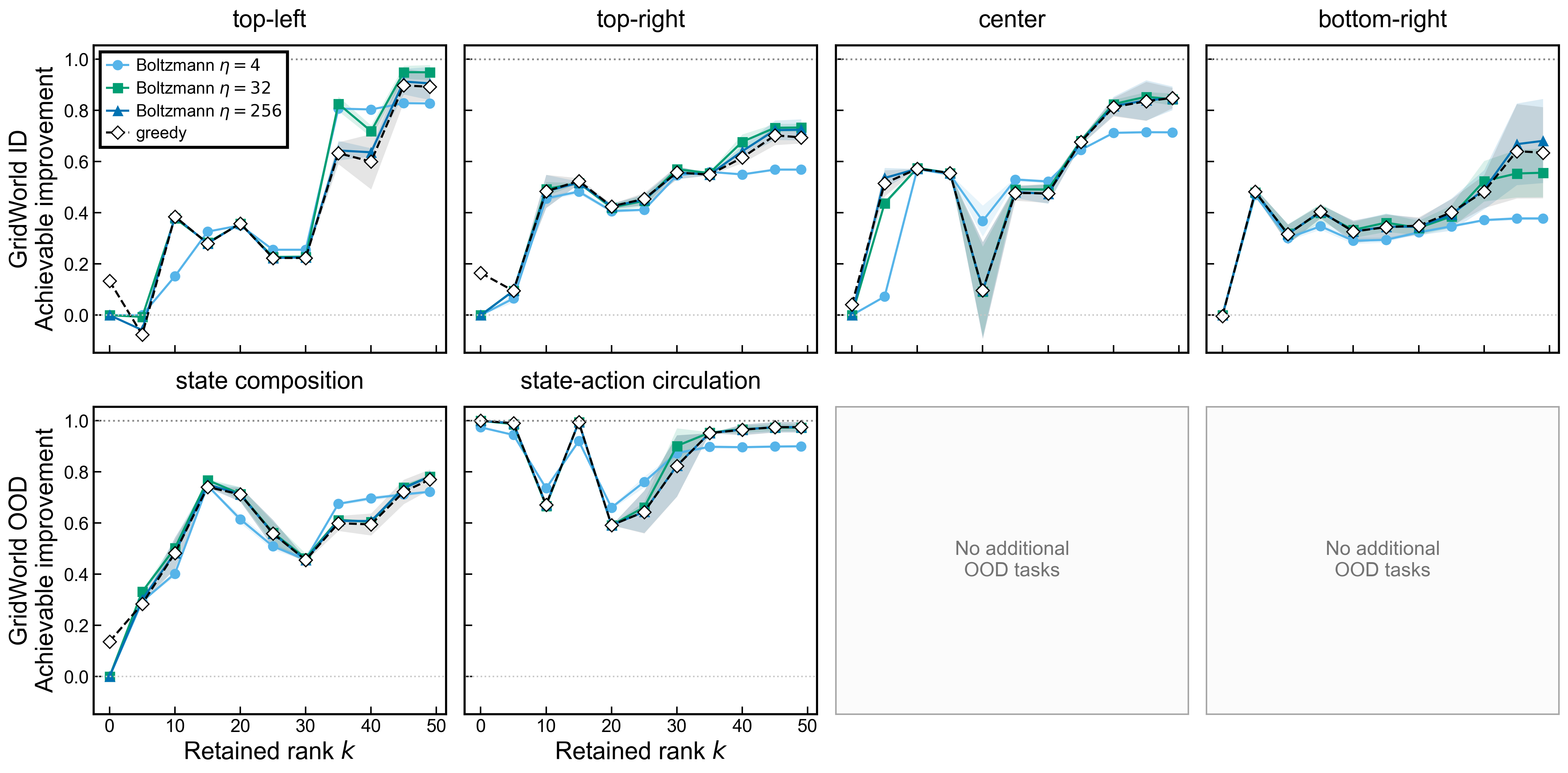}
    \caption{\textbf{One step policy update with learned SUAF in Grid World environment.} The results is calculated with 5 full-coverage random policies with 5 seeds learned SUAF with 95\% CI. This also works as an ablation of Boltzmann update with different temperature.}
    \label{fig:grid_world_learnt_suaf_ci}
\end{figure}

\begin{figure}
    \centering
    \includegraphics[width=\linewidth]{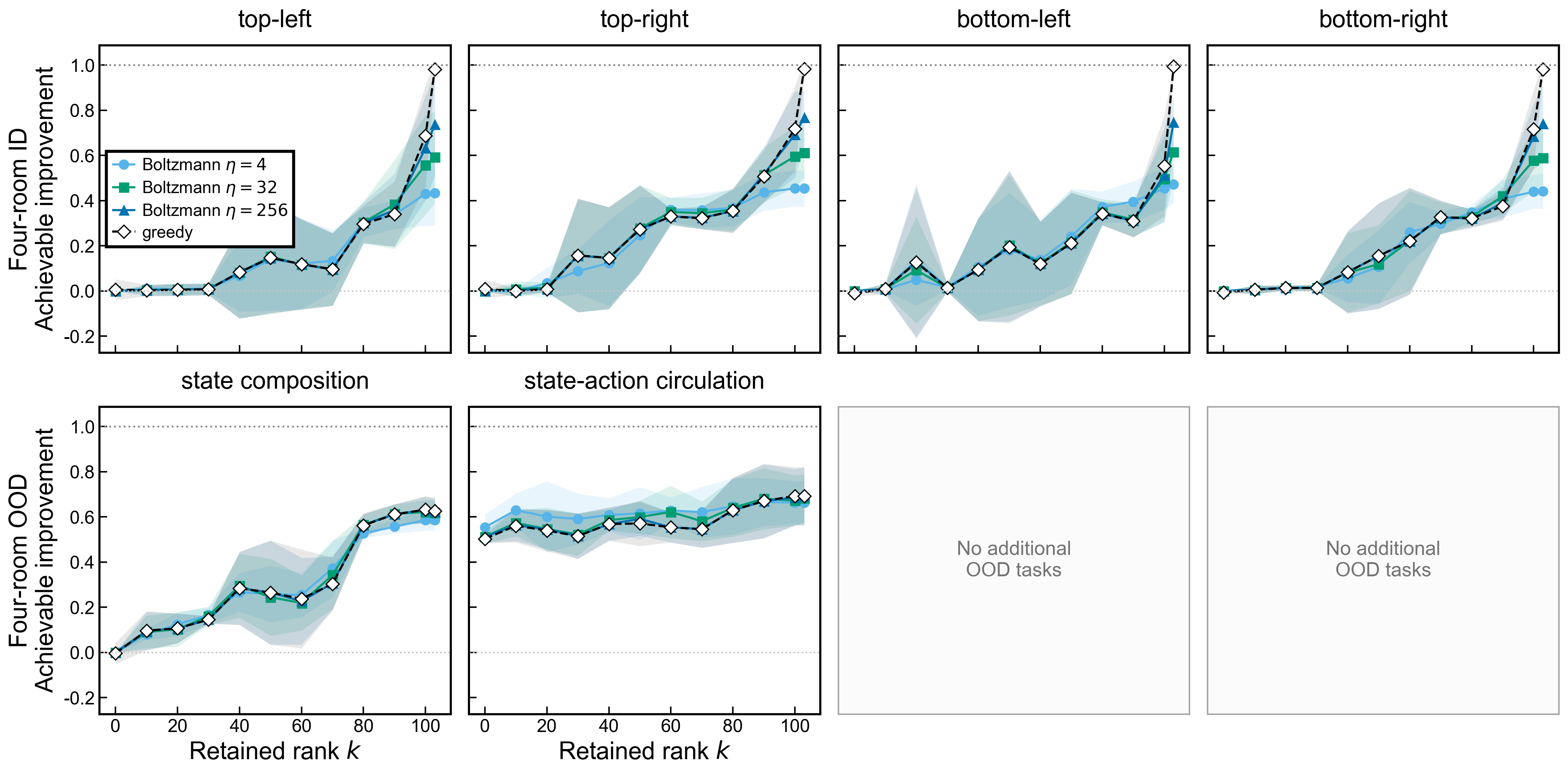}
    \caption{\textbf{One step policy update with exact SUAF in Four-room environment.} The results is calculated with 5 full-coverage random policies with 95\% CI. This also works as an ablation of Boltzmann update with different temperature.}
    \label{fig:four_room_exact_suaf}
\end{figure}
\begin{figure}
    \centering
    \includegraphics[width=\linewidth]{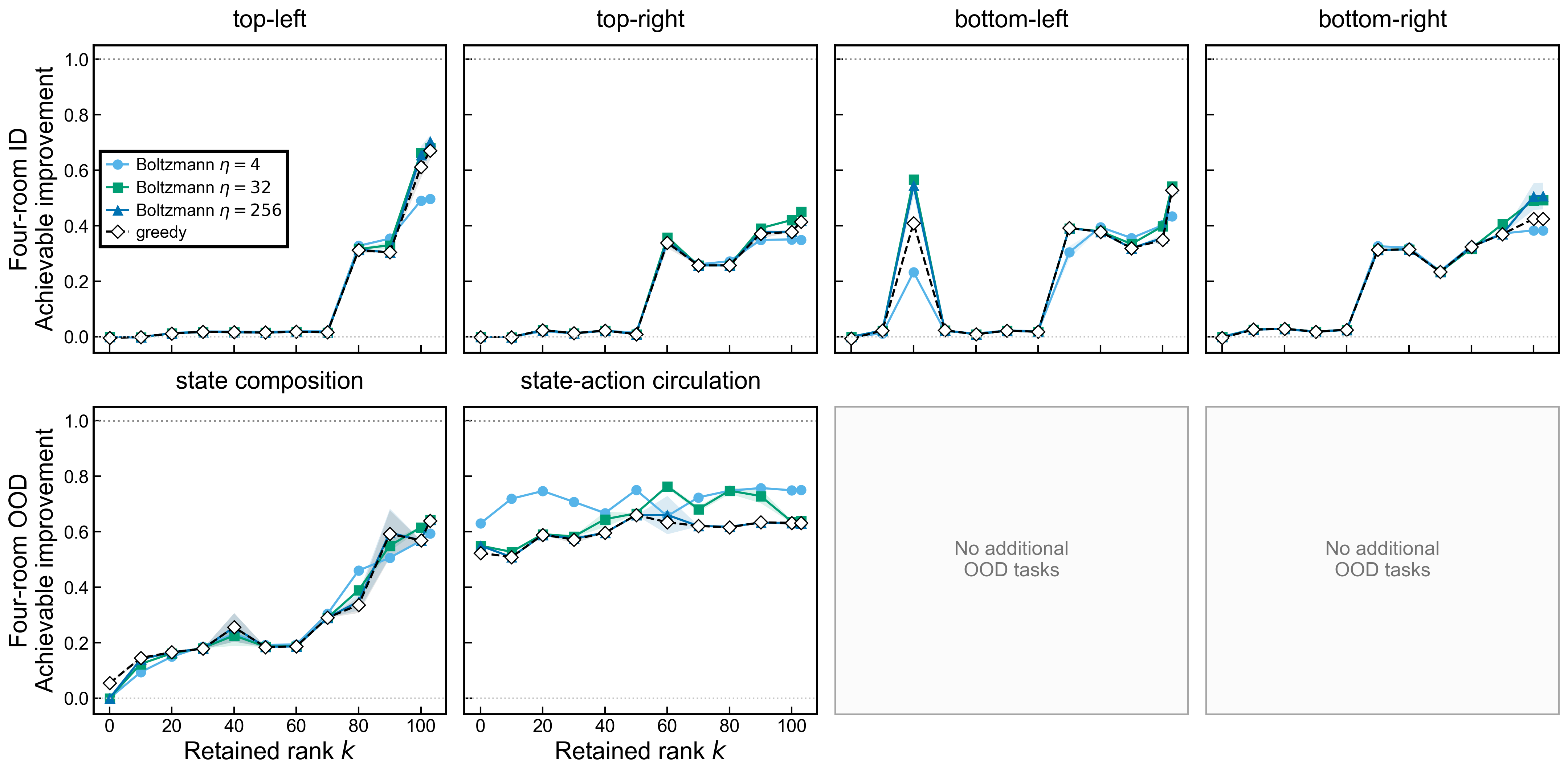}
    \caption{\textbf{One step policy update with learned SUAF in Four-room environment.} The results is calculated with 5 full-coverage random policies with 5 seeds learned SUAF with 95\% CI. This also works as an ablation of Boltzmann update with different temperature.}
    \label{fig:four_room_learnt_suaf_ci}
\end{figure}

\begin{figure}
    \centering
    \includegraphics[width=\linewidth]{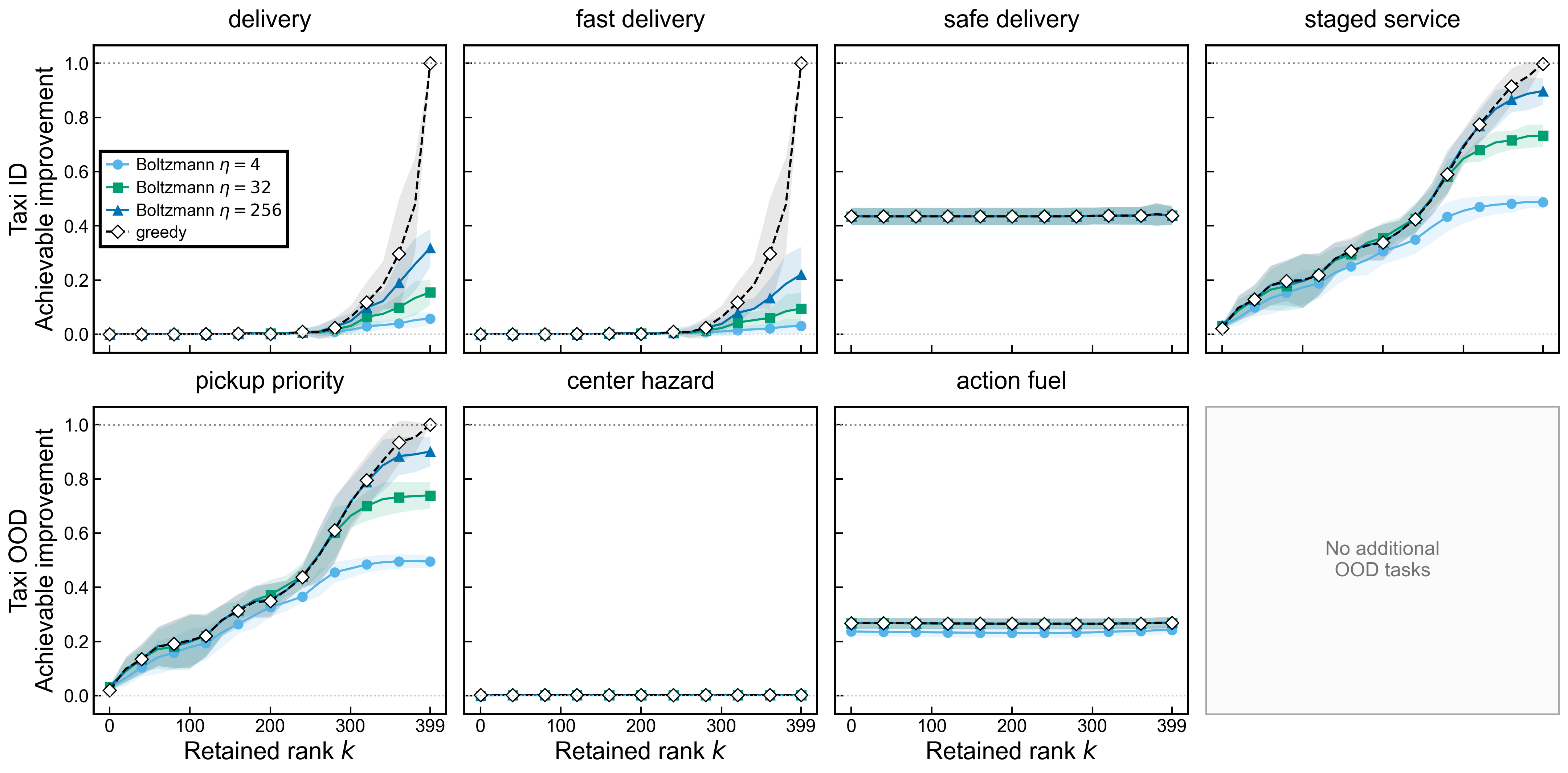}
    \caption{\textbf{One step policy update with exact SUAF in Taxi environment.} The results is calculated with 5 full-coverage random policies with 95\% CI. This also works as an ablation of Boltzmann update with different temperature. The Taxi model outputs 499 coordinates as an overparameterization diagnostic, while the population DCDM has rank 399. Coordinates beyond rank 399 do not correspond to additional nonzero population SUAF modes.}
    \label{fig:taxi_exact_suaf}
\end{figure}
\begin{figure}
    \centering
    \includegraphics[width=\linewidth]{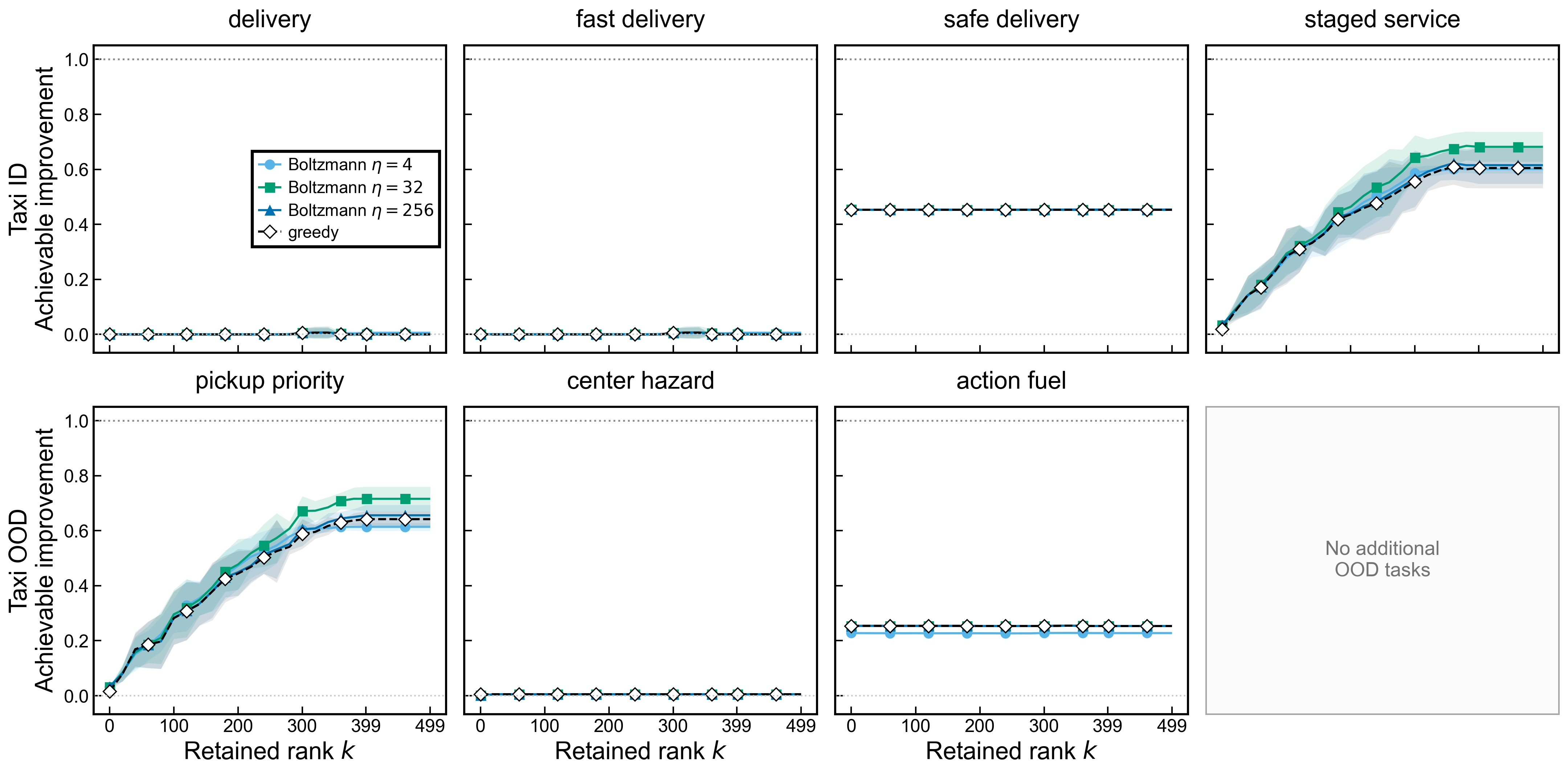}
    \caption{\textbf{One step policy update with learned SUAF in Taxi environment.} The results is calculated with 5 full-coverage random policies with 5 seeds with 95\% CI. This also works as an ablation of Boltzmann update with different temperature.}
    \label{fig:taxi_learnt_suaf_ci}
\end{figure}

\subsection{Spectral reward shaping in Taxi}
\label{sec:taxi_reward_shaping_exp}

\paragraph{Taxi as a sparse-reward stress test.}
We use the continuing Taxi environment as a deliberately difficult control problem. The state records the taxi position, passenger status, and destination, and the delivery reward is nonzero on only $4$ of the $2400$ reachable state-action pairs. Under the fixed behavior policy $\pi_0$, the positive delivery event has discounted occupancy mass only
$1.48\times 10^{-5}$. Consequently, a reward-free representation must retain weak, long-horizon distinctions involving pickup, navigation, and drop-off in order to recover a useful delivery policy from a one-step readout.

This behavior is consistent with the discussion in
Section~\ref{sec:reward_shaping}. Write the state reward averaged under the behavior policy as
\begin{equation}
  r^{\pi_0}(s) := \sum_a \pi_0(a\mid s)r(s,a),
\end{equation}
and let $\{(f_i,g_i,\sigma_i)\}_{i=1}^{R}$ denote the SUAF modes of $\pi_0$,
where the right modes are orthonormal in
$L^2(\nu^{+,\pi_0})$. Define
\begin{equation}
  \beta_i(r)
  :=
  \left\langle g_i,r^{\pi_0}\right\rangle_{\nu^{+,\pi_0}}.
\end{equation}
For the unshaped delivery reward, only $0.88\%$ of the delayed spectral energy
$\sum_i \sigma_i^2\beta_i(r)^2$ lies in the first $64$ modes, whereas
$66.0\%$ lies after mode $256$. Thus, low-rank or imperfectly learned
representations can miss most of the signal even though the downstream reward
itself has a simple semantic description.

\paragraph{An oracle spectral potential.}
To isolate the effect of spectral reward allocation, define the weighted
projector onto the complete right singular subspace as
$\Pi_G=GG^\top W_\nu$, where
$W_\nu=\operatorname{diag}(\nu^{+,\pi_0})$, and let
\begin{align}
  r_{\perp}^{\pi_0}
  &:=
  (I-\Pi_G)r^{\pi_0}.
  \label{eq:taxi_reward_orthogonal}
\end{align}
In continuing Taxi there are $400$ reachable states and the population SUAF
has rank $R=399$.  Because the right singular functions are centered and
orthonormal under $\nu^{+,\pi_0}$, they span the complete subspace of
$\nu^{+,\pi_0}$-centered state functions.  The orthogonal complement is
therefore only the constant direction, and
\begin{equation}
  r_{\perp}^{\pi_0}
  =
  \mathbb E_{S\sim\nu^{+,\pi_0}}
  \left[r^{\pi_0}(S)\right]\mathbf 1.
  \label{eq:taxi_reward_orthogonal_constant}
\end{equation}
We can consequently define the head-$m$ target reward as
\begin{align}
  r_{\mathrm{head}\text{-}m}^{\pi_0}(s)
  &:=
  \mathbb E_{\nu^{+,\pi_0}}
  \left[r^{\pi_0}(S)\right]
  +\sum_{i=1}^{m}g_i(s)\beta_i(r).
  \label{eq:taxi_head_m_reward}
\end{align}
Importantly, \eqref{eq:taxi_head_m_reward} requires neither the tail coefficients $\{\beta_i\}_{i>m}$ nor the tail singular functions. Given the reward and $\pi_0$, the state reward $r^{\pi_0}$ is known and its weighted mean and the leading coefficients
$\beta_i(r)=\langle g_i,r^{\pi_0}\rangle_{\nu^{+,\pi_0}}$ can be estimated
directly from samples of $\nu^{+,\pi_0}$.  The constant orthogonal component is
retained, while only the representable spectral tail is removed.

We then construct the state-only potential
\begin{equation}
  \Phi_m
  =
  (I-\gamma P_{\pi_0})^{-1}
  \left(r^{\pi_0}-r_{\mathrm{head}\text{-}m}^{\pi_0}\right)
  \label{eq:taxi_spectral_potential}
\end{equation}
and the shaped transition reward
\begin{equation}
  r_{\Phi_m}(s,a,s')
  =r(s,a,s')+\gamma\Phi_m(s')-\Phi_m(s).
  \label{eq:taxi_shaped_reward}
\end{equation}
By construction,
\begin{equation}
  r_{\Phi_m}^{\pi_0}
  =r_{\mathrm{head}\text{-}m}^{\pi_0},
  \qquad
  \left\langle g_i,r_{\Phi_m}^{\pi_0}\right\rangle_{\nu^{+,\pi_0}}=0
  \quad(i>m).
\end{equation}
Potential shaping leaves the advantage and the set of optimal policies
unchanged. Operationally, \eqref{eq:taxi_spectral_potential} does not move
tail coefficients into head coefficients. Instead, it removes the tail from
the delayed component and transfers its exact contribution to the immediate
action-centered term.

\begin{figure}[t]
  \centering
  \includegraphics[width=\linewidth]{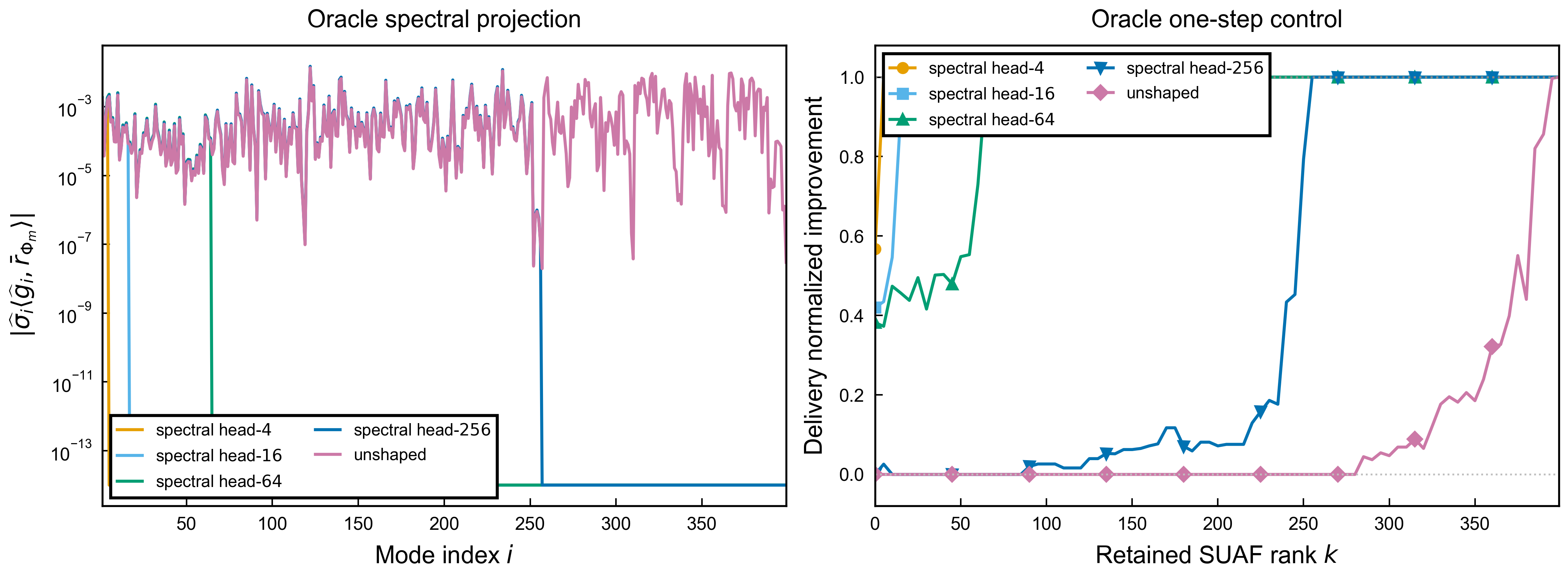}
  \caption{\textbf{Oracle spectral reward shaping in Taxi.} The left panel compares the spectral allocation of the sparse delivery reward and shaped variants for exact SUAF. The right panel reports one-step control on the original delivery objective.
  Curves or bars show the mean over four independently learned representations.}
  \label{fig:taxi_orcale_reward_shaping_exact_suaf}
\end{figure}

\begin{figure}[t]
  \centering
  \includegraphics[width=\linewidth]{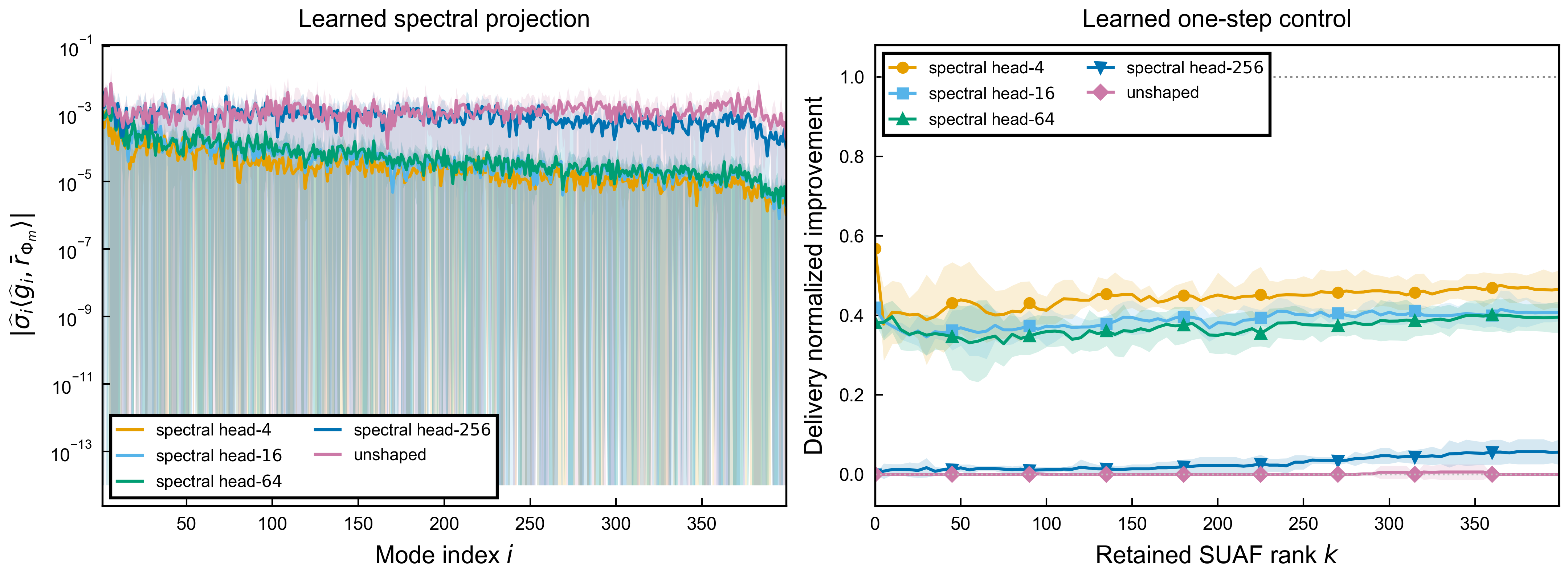}
  \caption{\textbf{Oracle spectral reward shaping in Taxi.} The left panel compares the spectral allocation of the sparse delivery reward and shaped variants for learned SUAF. The right panel reports one-step control on the original delivery objective.
  Curves or bars show the mean over four independently learned representations.}
  \label{fig:taxi_learned_reward_shaping_learned_suaf}
\end{figure}

\begin{figure}[t]
  \centering
  \includegraphics[width=\linewidth]{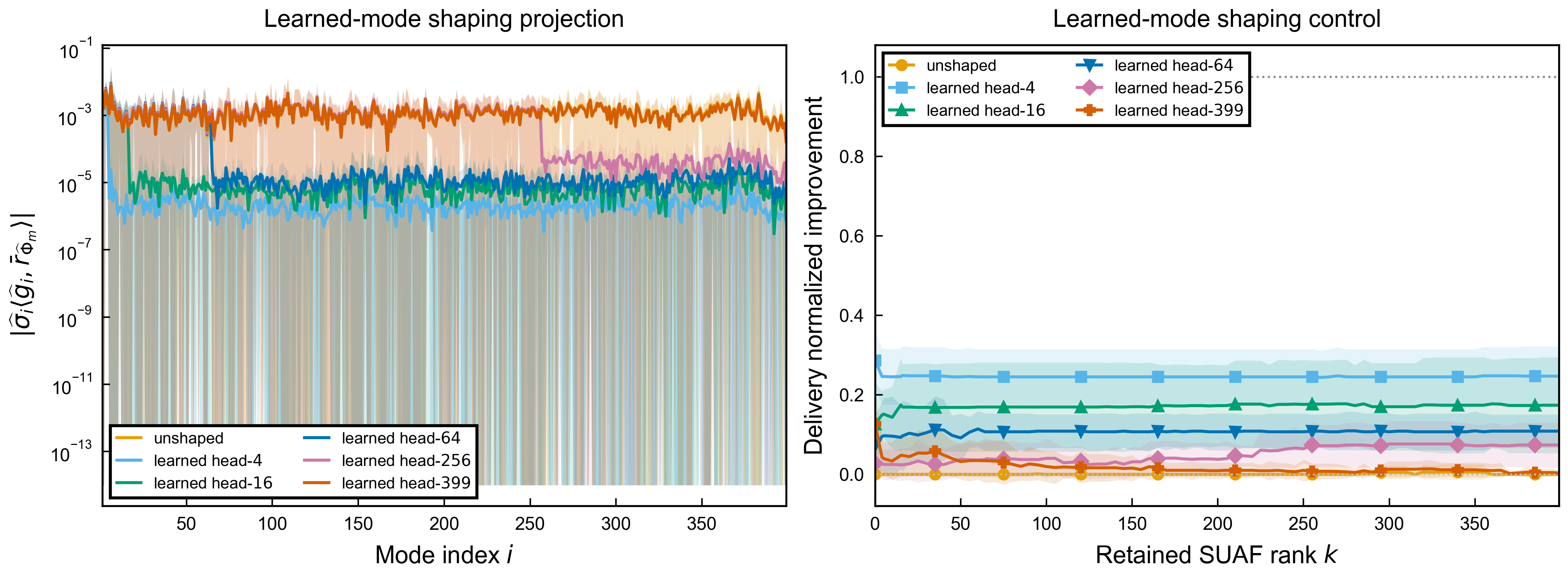}
  \caption{\textbf{Learned spectral reward shaping in Taxi.} The left panel compares the spectral allocation of the sparse delivery reward and shaped variants for learned SUAF. The right panel reports one-step control on the original delivery objective.
  Curves or bars show the mean over four independently learned representations.}
  \label{fig:taxi_orcale_reward_shaping_learned_suaf}
\end{figure}

\paragraph{Oracle-mode results.}
The population SUAF has rank $399$. With the original sparse reward, oracle
greedy one-step control remains near zero through rank $280$ and reaches the
optimal policy only at rank $399$. The four learned full-rank SUAF models also
obtain approximately zero improvement on the unshaped reward. In contrast,
the oracle spectral potentials make the task accessible at the prescribed
head rank for exact SUAF to recover the optimal policy, as is shown in Figure \ref{fig:taxi_orcale_reward_shaping_exact_suaf}. Furthermore, using the learned representations, head-$4$, head-$16$, and head-$64$ shaping obtains normalized improvements $0.466\pm0.047$, $0.407\pm0.021$, and $0.396\pm0.039$, respectively (mean and
$95\%$ confidence interval), compared with approximately zero without
shaping.

\paragraph{A learned spectral potential.}
We next replace the oracle right singular functions by the calibrated modes
from each learned checkpoint and remove the constant component altogether:
\begin{equation}
  \widehat r_{\mathrm{head}\text{-}m}^{\pi_0}(s)
  =
  \sum_{i=1}^{m}
  \widehat g_i(s)\widehat\beta_i(r),
  \qquad
  \widehat\beta_i(r)
  =
  \left\langle
  \widehat g_i,r^{\pi_0}
  \right\rangle_{\nu^{+,\pi_0}}.
\end{equation}
For this diagnostic we still use the known Taxi transition kernel to compute
\begin{equation}
  \widehat\Phi_m
  =
  (I-\gamma P_{\pi_0})^{-1}
  \left(
  r^{\pi_0}-\widehat r_{\mathrm{head}\text{-}m}^{\pi_0}
  \right).
\end{equation}
Thus, the construction uses no oracle singular functions and no tail
coefficients, but it continues to use oracle dynamics.  At the matched
truncation rank $k=m$, learned head-$4$, head-$16$, head-$64$, and head-$256$
shaping obtains normalized delivery improvements
$0.246\pm0.071$, $0.169\pm0.109$, $0.107\pm0.046$, and
$0.074\pm0.056$, respectively.  The unshaped learned SUAF remains near zero.
The smallest learned head therefore provides a clear but partial recovery,
although it remains below the corresponding oracle-mode construction.  Using
all $399$ learned modes again gives approximately zero improvement, showing
that the gain comes from removing unreliable learned directions rather than
from potential shaping alone.  Correcting the learned Gram matrix before
projection produces nearly identical results, so the remaining gap is not
explained solely by imperfect orthogonality of the calibrated
$\widehat g_i$.

\paragraph{A fully learned spectral potential.}
The preceding experiment still computes the potential with the population
transition matrix.  We remove this final oracle component using a fixed offline
replay
$\mathcal D=\{(S_n,A_n,S'_n)\}_{n=1}^{N}$.  In the finite Taxi environment we
first estimate
\begin{align}
 \widehat P_{\mathcal D}(s'\mid s,a)
 & =
 \frac{
 \sum_{n=1}^{N}
 \mathbf 1\{S_n=s,A_n=a,S'_n=s'\}
 }{
 \sum_{n=1}^{N}
 \mathbf 1\{S_n=s,A_n=a\}
 },\nonumber\\
 \widehat P_{\pi_0,\mathcal D}(s,s')
 & =
 \sum_a\pi_0(a\mid s)\widehat P_{\mathcal D}(s'\mid s,a).
\end{align}
No population transition probabilities are used in this estimator. Since $(I-\gamma\widehat P_{\pi_0,\mathcal D})^{-1} = I+\gamma\widehat P_{\pi_0,\mathcal D} + O(\gamma^2)$, we then
compute the fully learned potential
\begin{equation}
 \widetilde\Phi_{m,\mathcal D} = (I+\gamma\widehat P_{\pi_0,\mathcal D})\left(
 r^{\pi_0}
 -
 \widehat r_{\mathrm{head}\text{-}m}^{\pi_0}
 \right).
 \label{eq:taxi_fully_learned_potential}
\end{equation}
This operation is equivalent to tabular fitted policy evaluation of the
residual reward.  In large or continuous state spaces,
\eqref{eq:taxi_fully_learned_potential} can therefore be replaced by an
offline FQE or TD critic without explicitly materializing a transition matrix.

\begin{figure}
    \centering
    \includegraphics[width=0.5\linewidth]{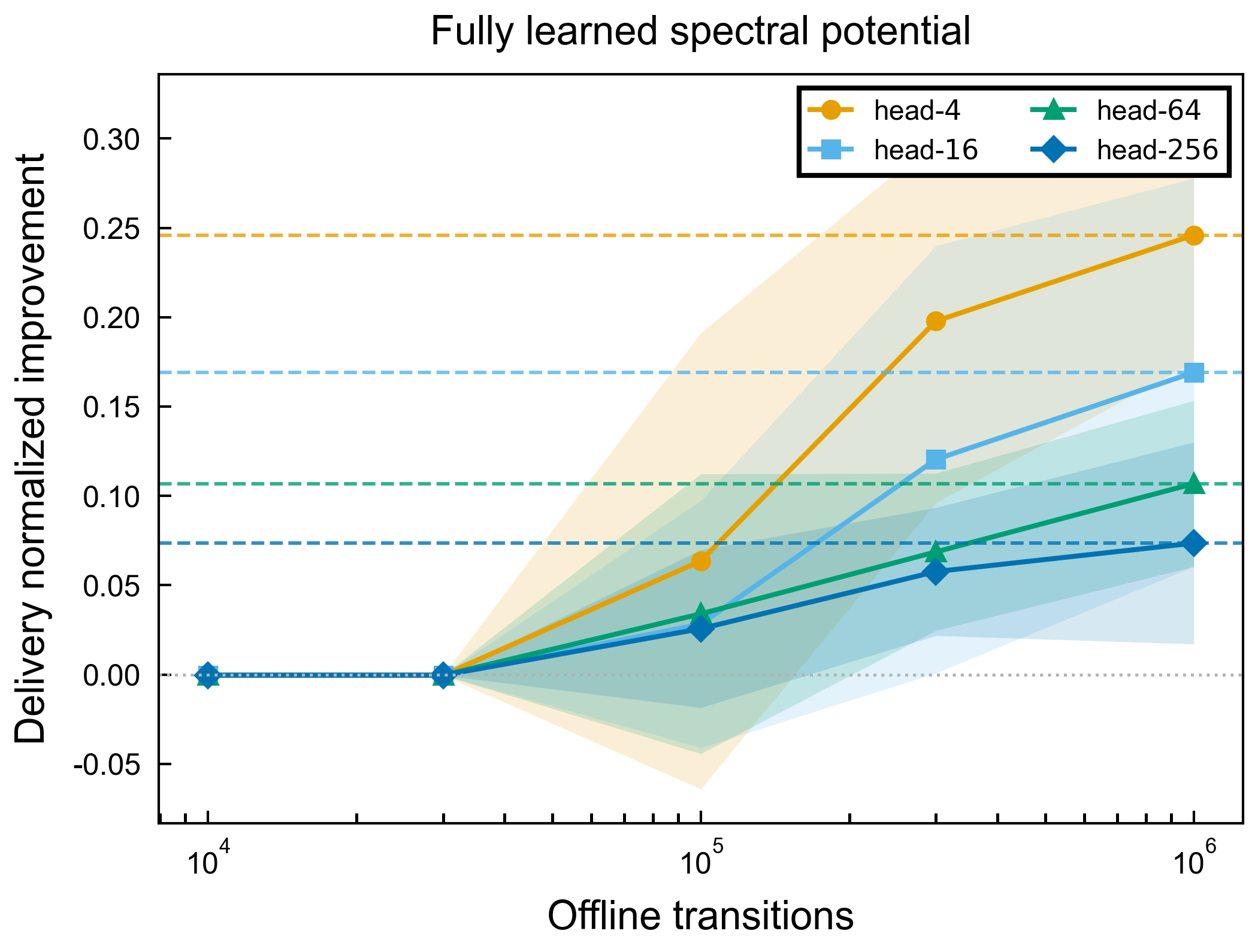}
    \caption{\textbf{Estimation of $P_\pi$ with dataset.} Solid curves show means, and shaded regions indicate std across 5 independent runs.}
    \label{fig:p_pi_estimation}
\end{figure}

As is shown in Figure \ref{fig:p_pi_estimation}, we evaluate replay sizes
$N\in\{10^4,3\times10^4,10^5,3\times10^5,10^6\}$ and save the replay used for every seed.  With $10^4$ or $3\times10^4$ transitions, incomplete state-action coverage causes all variants to remain near zero.  At $10^5$ transitions, learned head-$4$ reaches
$0.064\pm0.128$ normalized improvement, increasing to
$0.198\pm0.102$ at $3\times10^5$.  With $10^6$ transitions it reaches
$0.246\pm0.071$, numerically matching the learned-mode potential computed with
the exact transition matrix.  The same convergence occurs for head-$16$,
head-$64$, and head-$256$, whose $10^6$-transition results are respectively
$0.169\pm0.109$, $0.107\pm0.046$, and $0.074\pm0.056$.  At this sample size the
dataset covers on average $2398.25$ of the $2400$ reachable state--action
pairs, and the weighted potential error for head-$4$ decreases to
$2.1\times10^{-4}$.  Thus the benefit of learned spectral shaping survives
after both the singular functions and the potential are learned from data.

\subsection{Policy Update within Trust Region}

\begin{figure}
    \centering
    \includegraphics[width=\linewidth]{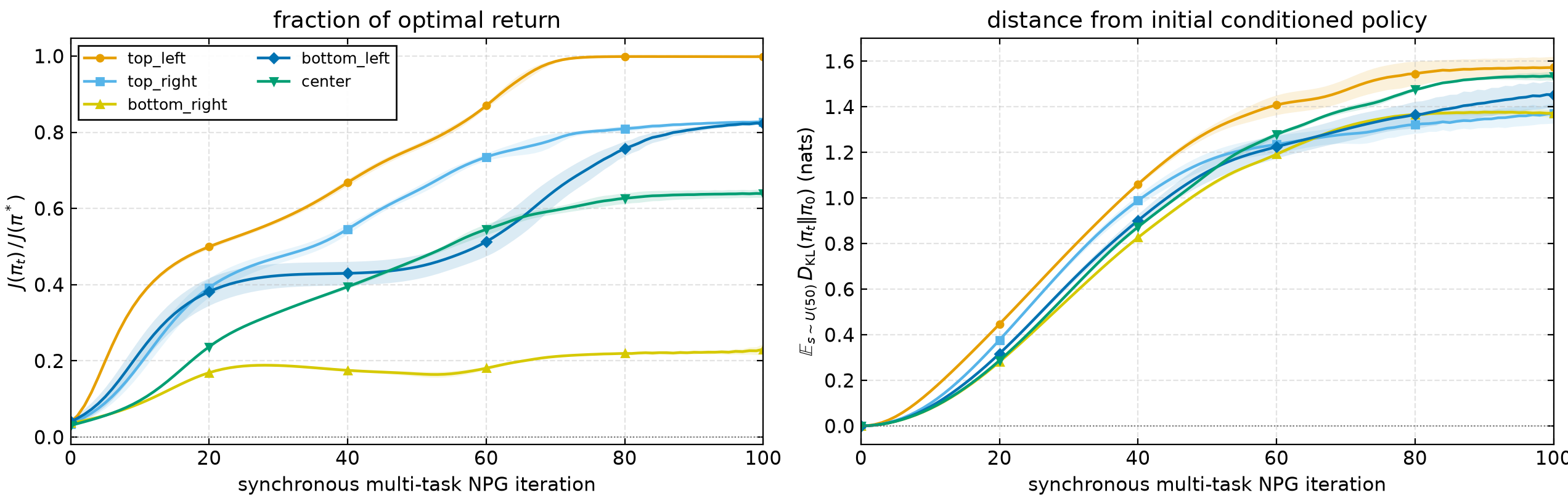}
    \caption{\textbf{Iterative multi-task policy improvement with SUAF in Grid World.} A shared reward-conditioned actor is optimized using synchronous multi-task natural policy gradient (NPG), with SUAF critics reused between refreshes triggered by the trust-region criterion. Solid curves show means, and shaded regions indicate std across 3 independent runs. \textbf{Left:} Return normalized by the task-specific optimal return, $J_r(\pi_t)/J_r(\pi_r^*)$. \textbf{Right:} KL divergence from the initial task-conditioned policy, averaged uniformly over the 50 states. Colors indicate reward tasks.}
    \label{fig:update_grid_world}
\end{figure}

\begin{figure}
    \centering
    \includegraphics[width=\linewidth]{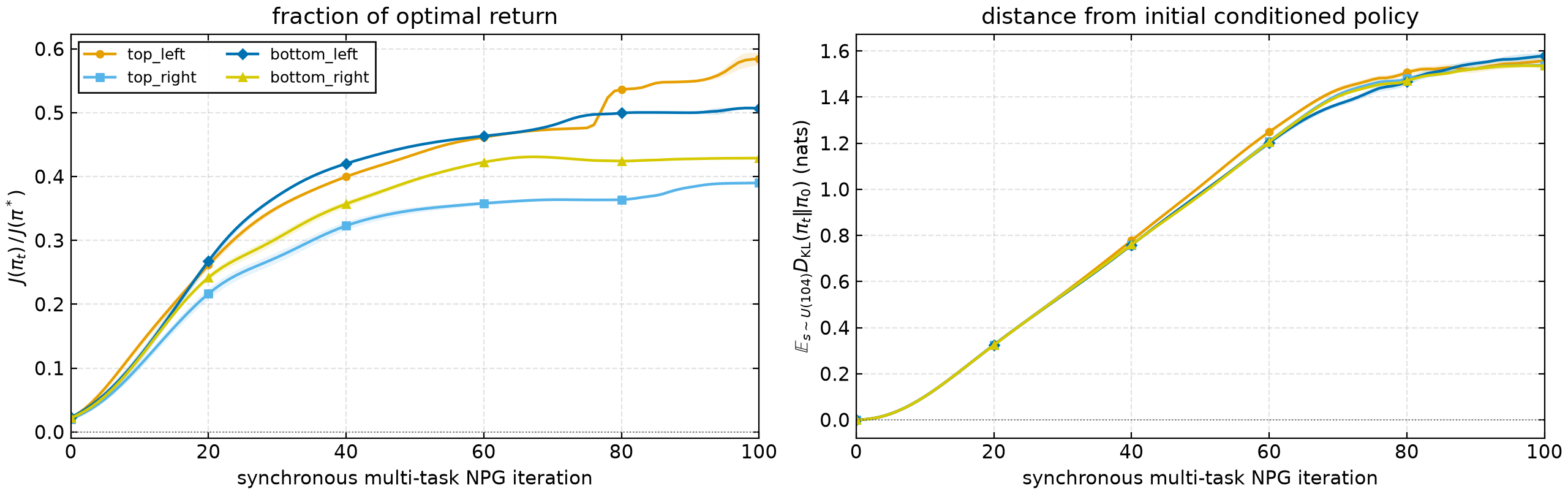}
    \caption{\textbf{Iterative multi-task policy improvement with SUAF in Four-rooms.} A shared reward-conditioned actor is optimized using synchronous multi-task natural policy gradient (NPG), with SUAF critics reused between refreshes triggered by the trust-region criterion. Solid curves show means, and shaded regions indicate std across 3 independent runs.\textbf{Left:} Return normalized by the task-specific optimal return, $J_r(\pi_t)/J_r(\pi_r^*)$. \textbf{Right:} KL divergence from the initial task-conditioned policy, averaged uniformly over the 104 states. Colors indicate reward tasks.}

    \label{fig:update_four_room}
\end{figure}

\begin{figure}
    \centering
    \includegraphics[width=\linewidth]{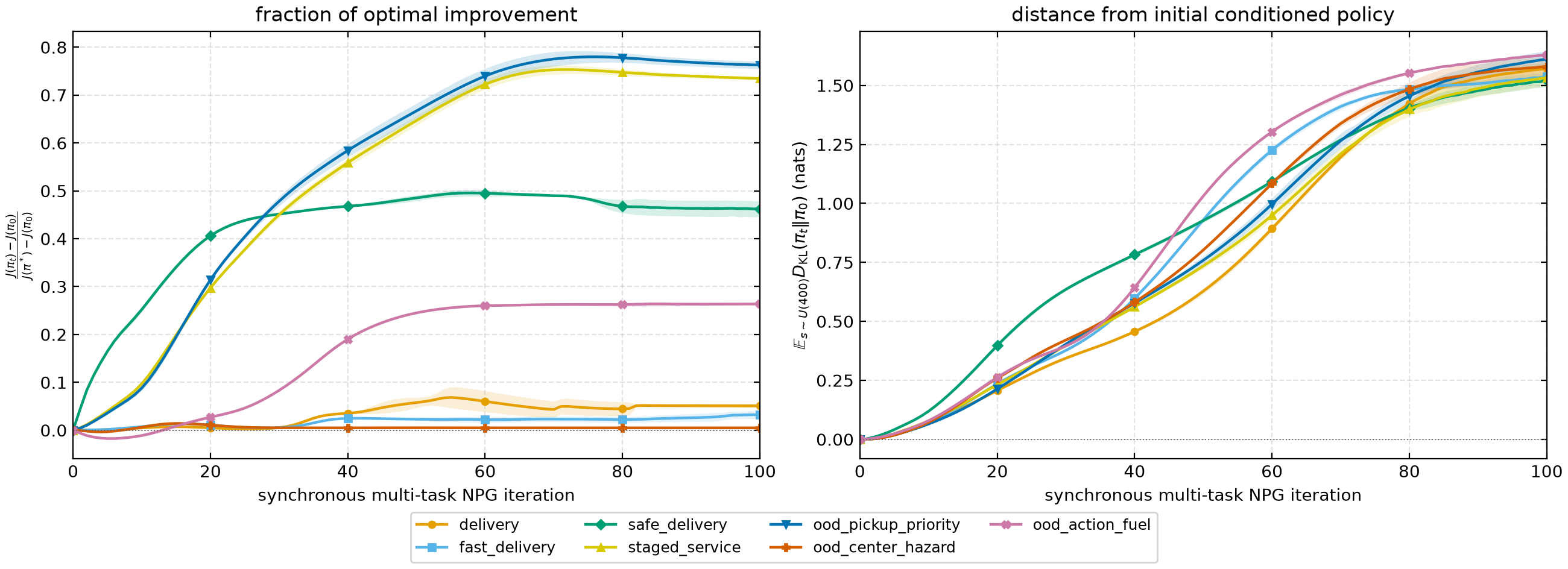}
    \caption{\textbf{Iterative multi-task policy improvement with SUAF in Taxi.} A shared reward-conditioned actor is optimized using synchronous multi-task natural policy gradient (NPG), with SUAF critics reused between refreshes triggered by the trust-region criterion. Solid curves show means, and shaded regions indicate std across 3 independent runs.\textbf{Left:} Return normalized by the task-specific optimal return, $J_r(\pi_t)/J_r(\pi_r^*)$. \textbf{Right:} KL divergence from the initial task-conditioned policy, averaged uniformly over the 104 states. Colors indicate reward tasks.}
    \label{fig:update_taxi}
\end{figure}

After pretraining the SUAF features, we use the right features as a reward encoder. For a reward $r$, its task representation is
\[
    \beta(r)
    =
    \big(
    \langle g_1,r\rangle_{\nu^{+,\pi_{\rm ref}}},
    \ldots,
    \langle g_k,r\rangle_{\nu^{+,\pi_{\rm ref}}}
    \big).
\]
A single reward-conditioned actor is shared across all tasks:
\[
    \pi_\theta(a\mid s,r)
    =
    \pi_\theta(a\mid s,\beta(r)).
\]
Thus, the actor receives both the current state and the SUAF reward
representation, while all rewards are optimized jointly using the same
parameter vector $\theta$. This allows policy knowledge to be shared across
tasks rather than training a separate actor for each reward.

The SUAF critic is associated with a reference policy
$\pi_{\rm ref}$. During policy optimization, its features are reused while
the current policy remains inside a trust region. We refresh the critic only
when $ D_{\rm KL}\!\left(\pi_\theta\,\|\,\pi_{\rm ref}\right)>\delta$.
At a refresh, new on-policy transitions are collected, the critic features
$f$ and $g$ are warm-started from their previous parameters and recalibrated,
and the current policy becomes the new reference policy. The reward
coordinate supplied to the actor may remain defined in the original
pretrained coordinate system, providing a stable task representation while
the policy-specific critic is updated.

We evaluate whether SUAF supports repeated policy improvement with a shared reward-conditioned actor and periodic critic refreshes. In Grid World, all five tasks achieve higher returns after optimization, although their learning trajectories differ. The top-left task reaches approximately the optimal return by iteration 80, while top-right and bottom-left reach approximately $83\%$ of their respective optima by iteration 100. The center task reaches approximately $64\%$, whereas bottom-right improves more modestly to approximately $23\%$ and exhibits a temporary decline during training. These results demonstrate that SUAF reward representations can support joint optimization of multiple tasks through a single actor, while also revealing substantial variation in the effectiveness of the updates across rewards.

In Four Rooms, all four tasks improve from their initial policies, reaching approximately $59\%$, $51\%$, $43\%$, and $39\%$ of optimal return for top-left, bottom-left, bottom-right, and top-right, respectively. Most tasks improve rapidly during the early iterations and subsequently progress more slowly, with top-left showing additional gains late in training. The remaining optimality gaps indicate that repeated updates with a shared actor and refreshed SUAF critics do not uniformly recover optimal behavior within the evaluated budget.

The right panels show that the policies move progressively away from their initial task-conditioned policies, with the final mean KL divergence ranging from approximately $1.4$ to $1.6$ nats in Grid World and clustering around $1.5$--$1.6$ nats in Four Rooms. Similar cumulative policy displacements can produce substantially different returns, emphasizing that update magnitude alone does not determine improvement. These plots measure divergence from the initial policy under a uniform state distribution. Local reuse is instead governed by the refresh criterion relative to the current reference policy. Together, the results demonstrate the feasibility of combining SUAF-based task conditioning with iterative policy optimization and critic refreshes, while exposing task-dependent limitations in the resulting performance.

\section{Ablation Studies}

\subsection{Does Separating the Immediate Contrast Matter for Control?}
\label{sec:full_advantage_control_ablation}

\begin{figure}
    \centering
    \includegraphics[width=\linewidth]{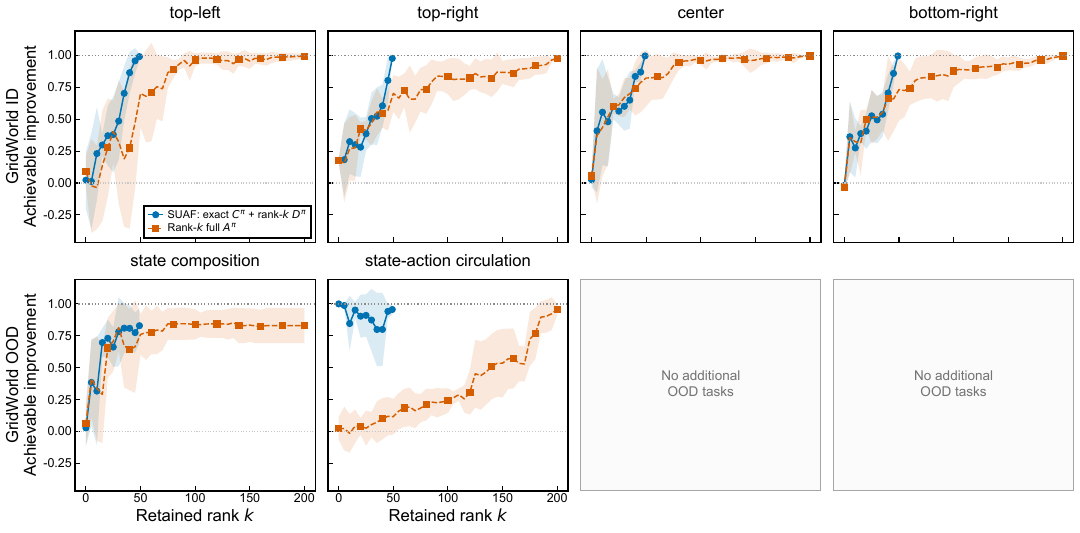}
    \caption{\textbf{Comprasion between SUAF and SVD in one-step improvement in GridWorld.} The results are reported as mean with std of 5 independent runs.}
    \label{fig:gridworld_greedy}
\end{figure}

\begin{figure}
    \centering
    \includegraphics[width=\linewidth]{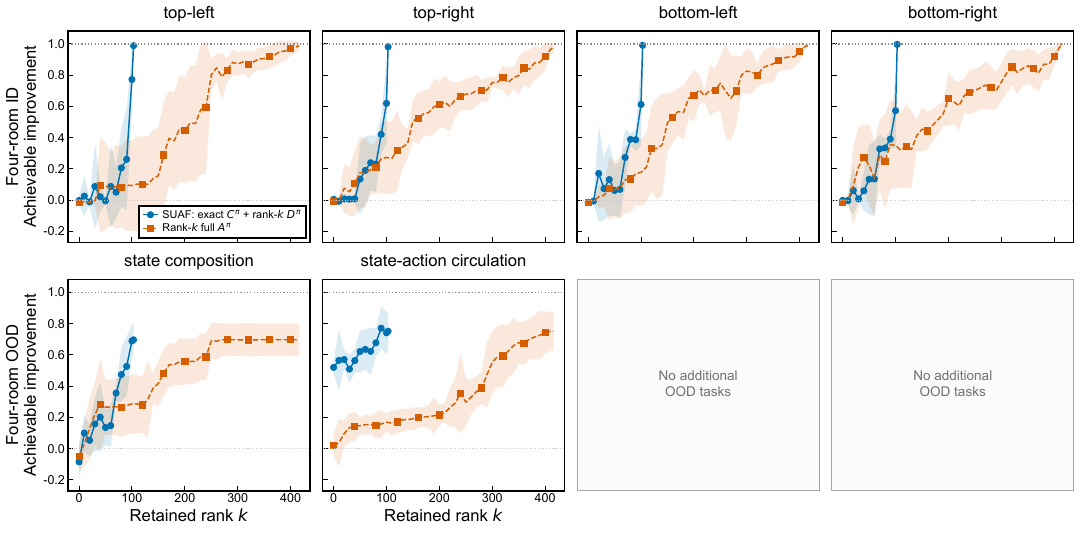}
    \caption{\textbf{Comprasion between SUAF and SVD in one-step improvement in Four-room.} The results are reported as mean with std of 5 independent runs.}
    \label{fig:four_rooms_greedy}
\end{figure}

\begin{figure}
    \centering
    \includegraphics[width=\linewidth]{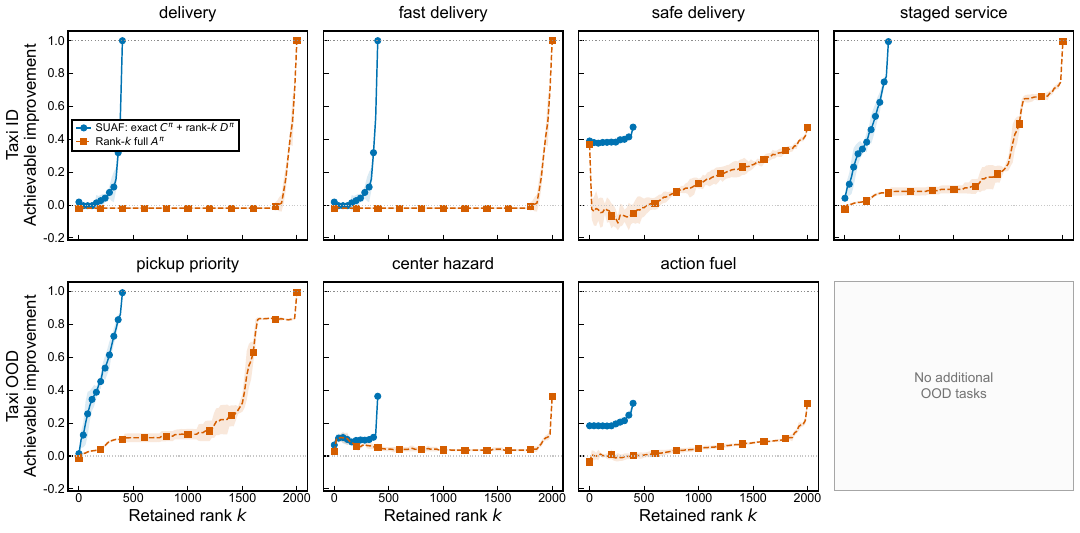}
    \caption{\textbf{Comprasion between SUAF and SVD in one-step improvement in Taxi.} The results are reported as mean with std of 5 independent runs.}
    \label{fig:taxi_greedy}
\end{figure}

Theorem~\ref{thm:removing_spectral_floor} shows that directly
compressing the full advantage operator must spend representation
capacity on the immediate action contrast, whereas SUAF retains this
known term exactly and uses its spectral modes only for delayed credit.
We next examine whether this distinction matters for downstream control,
rather than only for operator reconstruction error.

\paragraph{Compared operators.}
For a fixed reference policy $\pi_0$, write the advantage operator as
\begin{equation}
    \mathcal A^{\pi_0}
    =
    C_{\pi_0}+\mathcal D^{\pi_0},
    \qquad
    \mathcal D^{\pi_0}
    =
    \frac{\gamma}{1-\gamma}\mathcal T^{+,\pi_0}.
\end{equation}
We compare two population, reward-independent rank-$k$ constructions.
The SUAF approximation retains the immediate contrast exactly and
truncates only the delayed operator:
\begin{equation}
    \widehat A^{\mathrm{SUAF}}_{r,k}
    =
    C_{\pi_0}r+
    \mathcal D^{\pi_0}_k r,
\end{equation}
where $\mathcal D^{\pi_0}_k$ is the leading rank-$k$ weighted SVD
approximation of $\mathcal D^{\pi_0}$.  The ablation instead directly
truncates the full advantage operator:
\begin{equation}
    \widehat A^{\mathrm{Full}}_{r,k}
    =
    \left(\mathcal A^{\pi_0}\right)_k r,
\end{equation}
where $\left(\mathcal A^{\pi_0}\right)_k$ is the leading rank-$k$
approximation of the full operator under the same weighted
$L^2(d^{\pi_0})$--$L^2(\nu^{+,\pi_0})$ geometry.

Here, $k$ denotes the number of retained spectral modes that must be
represented.  The exact term $C_{\pi_0}r$ in SUAF introduces no learned
parameters: it is computed directly from the downstream reward and the
known reference policy.  Thus, the comparison allocates the same
\emph{learned spectral capacity} to both methods.

\paragraph{Control protocol.}
To isolate the structural effect from representation-learning error, we
compute both decompositions from the exact finite MDP.  For every
rank $k$ and downstream reward $r$, we construct a one-step greedy
policy
\begin{equation}
    \pi_k(a\mid s)
    =
    \mathbf 1
    \left\{
        a\in\arg\max_{a'}
        \widehat A_{r,k}^{\pi_0}(s,a')
    \right\},
\end{equation}
with fixed tie breaking.  We measure its achievable improvement by
\begin{equation}
    \operatorname{AI}(\pi_k)
    =
    \frac{
        J_r(\pi_k)-J_r(\pi_0)
    }{
        J_r(\pi_r^\star)-J_r(\pi_0)
    }.
    \label{eq:achievable_improvement}
\end{equation}
An achievable improvement of one therefore indicates that the one-step
policy reaches the optimal return, while zero corresponds to no
improvement over the reference policy. 

We repeat the population experiment over five independently sampled
full-support reference policies and report the mean and Student-$t$
$95\%$ confidence interval.  The uncertainty therefore measures
sensitivity to the reference policy, rather than stochastic error in
the exact SVD.

\paragraph{Results.}
Figure~\ref{fig:gridworld_greedy}, \ref{fig:four_rooms_greedy}, \ref{fig:taxi_greedy} show that the operator-level
spectral floor has a direct consequence for control.  The delayed
operators have ranks $49$, $103$, and $399$ in GridWorld, Four-room,
and Taxi, respectively.  At these same rank budgets, SUAF has already
recovered the exact advantage of the reference policy, whereas the
rank-$k$ decomposition of the full advantage operator still omits
many immediate action-contrast directions.



Overall, these results complement
Theorem~\ref{thm:removing_spectral_floor}: the immediate-term spectral
floor is not merely an operator-norm artifact, but can substantially
reduce downstream policy improvement under a fixed representation
budget.

\subsection{Does other methods benefit from delayed advantage?}

\begin{table}[t]
  \centering
  \small
  \caption{Full-rank greedy one-step control with frozen FB representations. The exact-$C_\pi$ variants (labeled delayed) use $C_\pi r+\widehat D^\pi r$ and original ones use $\widehat A^\pi r$. Values are normalized improvement from the reference policy. Result is reported with 5 independent runs.}
  \label{tab:exact-c-fb-ablation}
  \resizebox{\textwidth}{!}{
  \begin{tabular}{ll|cccccc}
    \toprule
    Environment & task & FB & FB + delayed & One-Step FB & One-Step FB + delayed & Ours w/o delayed & Ours \\
    \midrule
    \multirow{3}{*}{GridWorld}
    & top-left 
    & 0.979$_{\pm0.038}$ & 0.978$_{\pm0.032}$ 
    & 0.767$_{\pm0.126}$ & 0.759$_{\pm0.115}$ 
    & 0.746$_{\pm 0.079}$ & 0.893$_{\pm0.057}$ \\

     & state composition 
     & 0.890$_{\pm0.106}$ & 0.801$_{\pm0.104}$ 
     & 0.777$_{\pm0.044}$ & 0.772$_{\pm0.021}$ 
     & 0.640$_{\pm0.040}$ & 0.771$_{\pm0.017}$ \\
     
     & state-action circulation 
     & 0.058$_{\pm0.068}$ & 0.554$_{\pm0.045}$ (${\uparrow}$) 
     & 0.072$_{\pm0.041}$ & 0.685$_{\pm0.054}$ ($\uparrow$)
     & 0.860$_{\pm0.082}$ & 0.975$_{\pm0.021}$ \\
    \midrule
    \multirow{3}{*}{Four-room} 
    & top-left 
    & 0.508$_{\pm0.006}$ & 0.510$_{\pm0.008}$
    & 0.022$_{\pm0.003}$ & 0.020$_{\pm0.004}$
    & 0.561$_{\pm0.103}$ & 0.671$_{\pm0.031}$ \\

     & state composition 
     & 0.660$_{\pm0.117}$ & 0.595$_{\pm0.094}$
     & 0.321$_{\pm0.026}$ & 0.322$_{\pm0.021}$ 
     & 0.601$_{\pm0.036}$ & 0.640$_{\pm0.017}$ \\
     
     & state-action circulation 
     & 0.111$_{\pm0.068}$ & 0.398$_{\pm0.033}$ ($\uparrow$) 
     & 0.038$_{\pm0.017}$ & 0.483$_{\pm0.024}$ ($\uparrow$)  
     & 0.582$_{\pm0.026}$ & 0.632$_{\pm0.007}$\\

     \midrule
    \multirow{2}{*}{walker}
    & flip 
    & $277_{\pm100}$ & $262_{\pm90}$
    & $388_{\pm30}$  & $360_{\pm20}$ 
    & $147_{\pm120}$ & $357_{\pm42}$ \\
    & run 
    & $194_{\pm20}$ & $262_{\pm90}$
    & $198_{\pm30}$ & $200_{\pm18}$ 
    & $187_{\pm102}$ & $210_{\pm20}$ \\
    \bottomrule
  \end{tabular}}
\end{table}

To test the source of performance improvement, we adapt FB and One-Step FB to the same delayed-operator target as SUAF, rather than training a full successor representation and subtracting an estimated identity component afterwards.  Let $X_t=(S_t,A_t)$ under a fixed reference policy $\pi_0$, and sample a strictly positive geometric lag,
\begin{equation}
  L\sim \operatorname{Geom}(1-\gamma), \qquad L\in\{1,2,\ldots\},
  \qquad Y^+=X_{t+L}.
  \label{eq:fb-positive-lag-sampling}
\end{equation}
Thus the positive pairs used by the FB-style representation losses are drawn from
\begin{equation}
  M^{+,\pi_0}
  =(1-\gamma)P_{\pi_0}(I-\gamma P_{\pi_0})^{-1},
  \label{eq:fb-positive-lag-kernel}
\end{equation}
instead of from the usual full geometric successor kernel, which includes the
lag-zero atom.  We otherwise retain the original FB or One-Step FB network,
optimizer, batch size, and orthonormalization/LSIF objective.  In particular,
the diagonal positive-pair term is evaluated on $(X,Y^+)$, while the factual
one-step transition remains available for the bootstrap term.  Consequently,
the resulting reward-conditioned score is trained to represent the delayed
action-dependent component, denoted by $\widehat D^{\pi_0}_{\rm base}r$.
At evaluation, we do not estimate the immediate term.  We compute it exactly
from the known reference policy,
\begin{equation}
  (C_{\pi_0}r)(s,a)
  =r(s,a)-\sum_{\tilde a}\pi_0(\tilde a\mid s)r(s,\tilde a),
  \label{eq:exact-immediate-term}
\end{equation}
and use the critic
\begin{equation}
  \widehat A^{\pi_0}_{\rm base}r
  =C_{\pi_0}r+\widehat D^{\pi_0}_{\rm base}r.
  \label{eq:fb-exact-c-critic}
\end{equation}
The \eqref{eq:exact-immediate-term} requires neither a transition model
nor reward-specific representation training.  It is exact for every test
reward once $\pi_0$ is fixed.  This intervention is non-vacuous only for an
action-dependent reward: if $r(s,a)=r(s)$, then $C_{\pi_0}r=0$ and the delayed
critic and \eqref{eq:fb-exact-c-critic} coincide identically.

\paragraph{Why the ablation is restricted to FB and One-Step FB.}
This experiment requires a baseline whose representation-learning objective can be modified in a controlled way by replacing its successor positive examples with the pairs in \eqref{eq:fb-positive-lag-sampling}. FB and One-Step FB have explicit forward--backward successor objectives with such a positive-pair/bootstrapping decomposition, so the change preserves their architecture and learning rule while changing only the operator being learned.

The remaining baselines do not admit an analogous one-line intervention.
Laplacian and BYOL-$\gamma$ learn state geometry rather than a specified
successor operator; ICVF and HILP use task- or intention-conditioned value
objectives; and TD-JEPA uses a representation-specific prediction objective.
None exposes an explicit successor positive-pair term whose lag-zero mass can
be removed while retaining the original objective.  Adding a new
reward-conditioned delayed TD or successor-feature head to each method would
introduce an additional model, loss, and optimization budget.  Such a
comparison would measure that added head rather than the original baseline.
We therefore apply the exact-$C_{\pi_0}$ delayed-operator ablation only to FB
and One-Step FB.

\paragraph{Delayed ablation.}
Table~\ref{tab:exact-c-fb-ablation} isolates the effect of separating the immediate reward contrast \(C_\pi r\) from the delayed component. For state-only rewards, \(C_\pi r=0\), so adding an exact \(C_\pi\) term leaves performance unchanged, as expected. For state-action rewards, however, explicitly computing \(C_\pi r\) yields large gains for the baselines, confirming that direct low-rank compression of the full advantage operator suffers from the immediate-term spectral floor. Crucially, even when the baselines are given the same exact \(C_\pi r\), SUAF still shows great advantage over them, showing that its learned delayed spectral representation is itself stronger. Overall, the ablation supports the core design: exact immediate contrast plus a low-rank spectral model of the delayed operator.

\paragraph{How does the dataset size affect performance?}

\begin{figure}
    \centering
    \includegraphics[width=\linewidth]{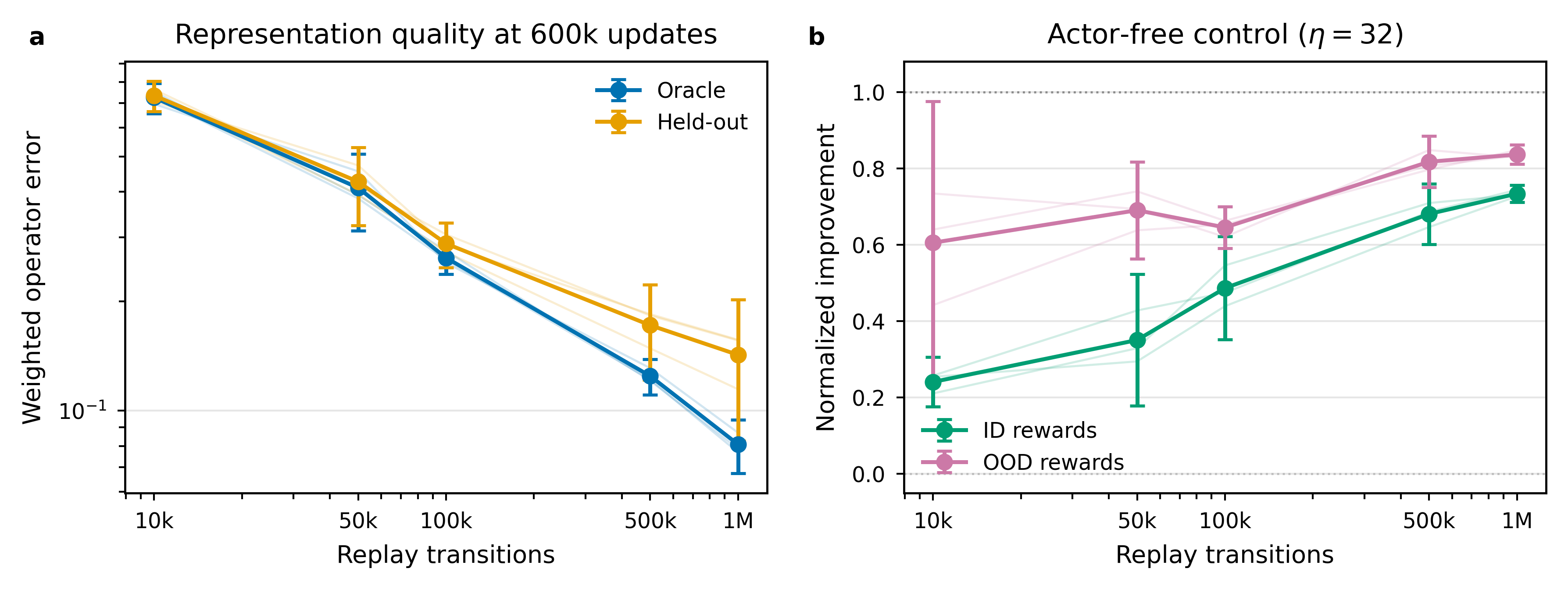}
    \caption{\textbf{Effects of training dataset on feature learning in GridWorld.} Results are reported with mean $\pm$ 95\% CI with 3 independent runs.}
    \label{fig:gridworld_learning}
\end{figure}

\begin{figure}
    \centering
    \includegraphics[width=\linewidth]{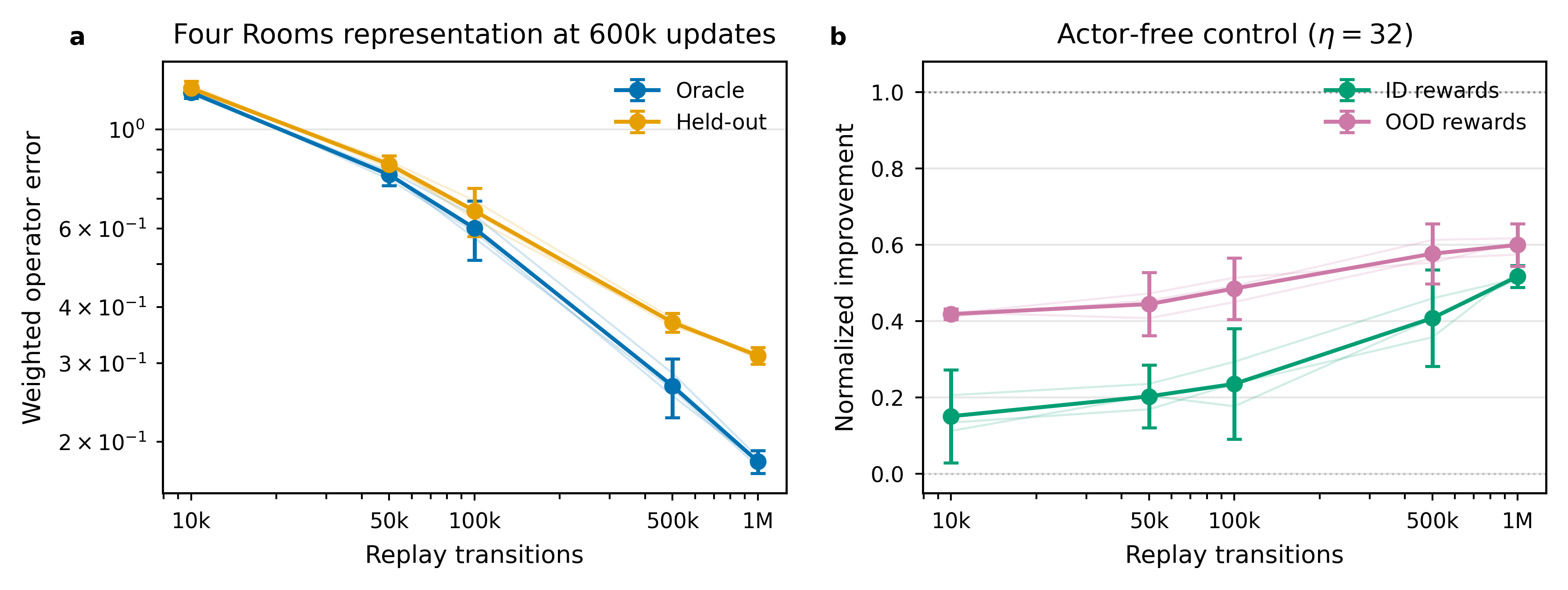}
    \caption{\textbf{Effects of training dataset on feature learning in Four-room.} Results are reported with mean $\pm$ 95\% CI with 3 independent runs.}
    \label{fig:fourroom_learning}
\end{figure}

\begin{figure}
    \centering
    \includegraphics[width=\linewidth]{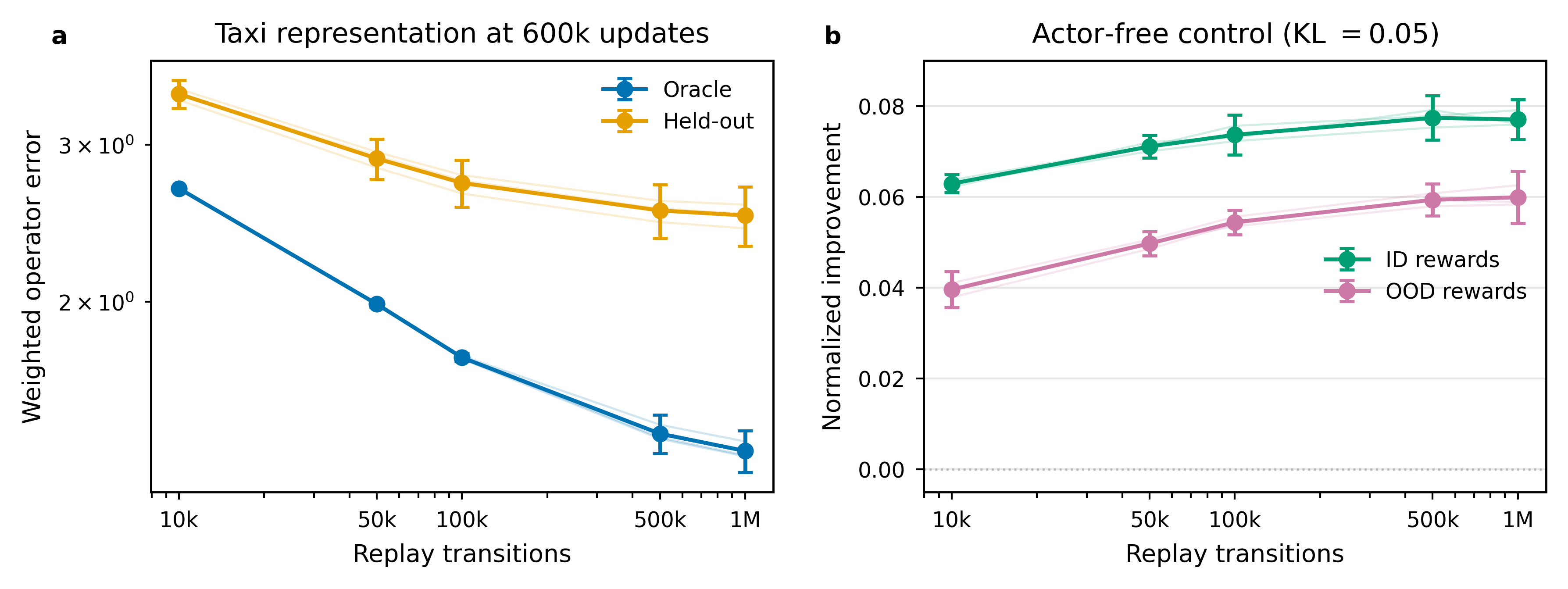}
    \caption{\textbf{Effects of training dataset on feature learning in Taxi.} Results are reported with mean $\pm$ 95\% CI with 3 independent runs.}
    \label{fig:taxi_learning}
\end{figure}

\subsection{How does estimation of behavior policy affect the final results?}

\begin{figure}
    \centering
    \includegraphics[width=0.5\linewidth]{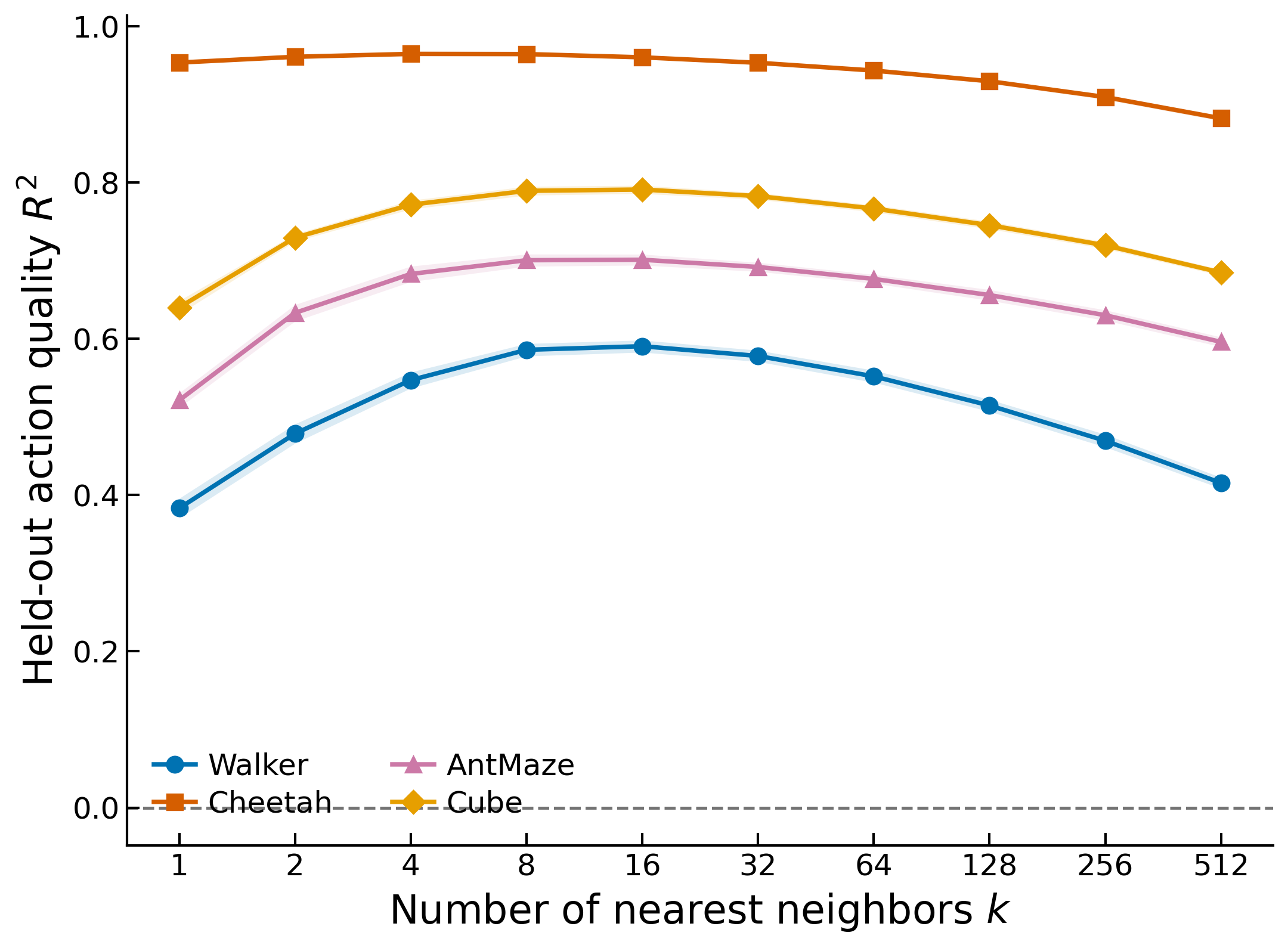}
    \caption{\textbf{Quality of kNN estimation measured by held-out action \(R^2\)}. The results are reported by 95\% CI of 5 independent episode split.}
    \label{fig:kNN_num}
\end{figure}

\begin{table*}[htbp]
\centering
\caption{\textbf{Ablations on kNN action numbers $k$.} The full zero-shot evaluation results on 8 ExORL tasks and 10 OGBench tasks. The results are reported as \textit{mean} over 3 seeds with \textit{std} while $k=16$ using the original results.}
\label{tab:ablation_kNN}
\resizebox{\textwidth}{!}{
\begin{tabular}{l l|c c c c c}
\toprule
Domain & Task & $k=1$ & $k=4$ & $k=16$ & $k=64$ & $k=256$ \\
\midrule
\rowcolor{gray!15} \multicolumn{7}{l}{\textbf{ExORL}: Locomotion}\\
\multirow{5}{*}{walker} & overall
& 117 \(\pm\) 12 
& 269 \(\pm\) 24
& \textbf{445 \(\pm\) 39} 
& 266 \(\pm\) 24 
& {58} \(\pm\) 11
\\

& flip 
& 43 \(\pm\) 6
& 252 \(\pm\) 12 
& \textbf{357 \(\pm\) 42}
& 281 \(\pm\) 2 
& {38 \(\pm\) 7}  
 \\

& run 
& 73 \(\pm\) 11 
& 184 \(\pm\) 13 
& \textbf{210 \(\pm\) 20}
& 108 \(\pm\) 33 
& {85 \(\pm\) 23} 
 \\

& stand 
& 172 \(\pm\) 20 
& 256 \(\pm\) 46
& \textbf{670 \(\pm\) 100}
& 286 \(\pm\) 12 
& {63 \(\pm\) 23} 
 \\

& walk 
& 181 \(\pm\) 7 
& 383 \(\pm\) 3
& \textbf{542 \(\pm\) 41}
& 391 \(\pm\) 56 
& {46 \(\pm\) 20} 
\\

\midrule
\multirow{5}{*}{cheetah} 
& overall 
& 395 \(\pm\) 21 
& 376 \(\pm\) 28
& \textbf{375 \(\pm\) 21}
& 278 \(\pm\) 21 
& 152 \(\pm\) 13 
\\

& run 
& 107 \(\pm\) 21 
& 94 \(\pm\) 33 
& \textbf{100} \(\pm\) 12
& 76 \(\pm\) 25 
& {28 \(\pm\) 17} 
 \\

& run backward 
& 149 \(\pm\) 16 
& 161 \(\pm\) 12 
& \textbf{159 \(\pm\) 70}
& 96 \(\pm\) 8 
& 57 \(\pm\) 11 
\\

& walk 
& 497 \(\pm\) 53 
& 447 \(\pm\) 44 
& \textbf{420 \(\pm\) 32}
& 199 \(\pm\) 30 
& 58 \(\pm\) 14 
\\

& walk backward 
& 826 \(\pm\) 75
& 804 \(\pm\) 80 
& \textbf{820 \(\pm\) 232}
& 742 \(\pm\) 26 
& 256 \(\pm\) 37 
\\
\midrule
\rowcolor{gray!15} \multicolumn{7}{l}{\textbf{OGBench}: Navigation and manipulation}\\
\multirow{6}{*}{antmaze large} 
& overall 
& 2 \(\pm\) 1 
& 17 \(\pm\) 5 
& \textbf{35 \(\pm\) 5}
& 17 \(\pm\) 6 
& 5 \(\pm\) 3 
\\

& task 1 
& 0 \(\pm\) 0 
& 10 \(\pm\) 3 
& \textbf{36 \(\pm\) 8}
& 20 \(\pm\) 8 
& 0 \(\pm\) 0 
\\

& task 2 
& 0 \(\pm\) 0 
& 10 \(\pm\) 4 
& \textbf{24 \(\pm\) 1}
& 11 \(\pm\) 3 
& 6 \(\pm\) 4 
\\

& task 3 
& 4 \(\pm\) 1
& 27 \(\pm\) 6 
& \textbf{57 \(\pm\) 5}
& 25 \(\pm\) 14 
& {6 \(\pm\) 2 }
\\

& task 4 
& 0 \(\pm\) 0 
& 5 \(\pm\) 2 
& \textbf{12 \(\pm\) 6}
& 3 \(\pm\) 1
& 0 \(\pm\) 0 
 \\

& task 5 
& 7 \(\pm\) 2 
& 33 \(\pm\) 19 
& \textbf{44 \(\pm\) 9}
& 28 \(\pm\) 10 
& 11 \(\pm\) 8
 \\

\midrule
\multirow{6}{*}{cube single play} 
& overall 
& 2 \(\pm\) 1 
& 14 \(\pm\) 6 
& \textbf{21 \(\pm\) 10}
& 13 \(\pm\) 5
& 3 \(\pm\) 2 
\\

& task 1 
& 2 \(\pm\) 1 
& 14 \(\pm\) 9 
& \textbf{20 \(\pm\) 7}
& 16 \(\pm\) 8 
& 4 \(\pm\) 3 
 \\

& task 2 
& 2 \(\pm\) 1 
& 12 \(\pm\) 4 
& \textbf{17 \(\pm\) 12}
& 10 \(\pm\) 5 
& 4 \(\pm\) 2 
\\

& task 3 
& 3 \(\pm\) 2 
& 13 \(\pm\) 8 
& \textbf{18 \(\pm\) 9} 
& 9 \(\pm\) 5 
& 1 \(\pm\) 1 
\\

& task 4 
& 0 \(\pm\) 0 
& 15 \(\pm\) 4 
& \textbf{21 \(\pm\) 12}
& 12 \(\pm\) 10 
& 1 \(\pm\) 0 
\\

& task 5 
& 4 \(\pm\) 2 
& 17 \(\pm\) 9
& \textbf{28 \(\pm\) 20}
& 18 \(\pm\) 9 
& 3 \(\pm\) 1 
 \\
\bottomrule
\end{tabular}
}
\end{table*}

We use the cross-episode held-out action \(R^2\) to measure the quality of the kNN estimate of the behavior policy \(\pi_b(a\mid s)\), where higher values indicate that the neighboring actions more accurately predict the behavior action at the current state. The \(R^2\) is computed on held-out data from different episodes as
\[
R^2
=
1-
\frac{
\sum_{i=1}^{N}
\left\|
a_i-\widehat a_k(s_i)
\right\|_2^2
}{
\sum_{i=1}^{N}
\left\|
a_i-\bar a_{\mathrm{ref}}
\right\|_2^2
},
\]
where
\[
\widehat a_k(s_i)
=
\frac{1}{k}\sum_{j\in\mathcal N_k(s_i)}a_j
\]
is the average of the \(k\) nearest-neighbor actions for state \(s_i\), and \(\bar a_{\mathrm{ref}}\) is the average action over the entire reference set. Thus:
\begin{itemize}
    \item \(R^2=1\): perfect prediction of the held-out action;
    \item \(R^2=0\): equivalent to always predicting the global mean action;
    \item \(R^2<0\): worse than the global mean action.
\end{itemize}
The reference and held-out queries come from different episodes, which avoids leakage from adjacent trajectory samples and prevents inflated results.

All four environments exhibit a clear bias--variance trade-off: too few neighbors lead to high estimation variance, whereas too many neighbors mix in samples whose states differ substantially. Walker, AntMaze, and Cube perform best around \(k=16\), while Cheetah achieves its best performance at \(k=4\sim8\). Overall, \(k=16\) is a reasonably robust unified choice across environments. Furthermore, we train with different number of kNN actions and found the result agree with the outcome.

\section{Theory Proofs}
\label{app:proofs}

Table~\ref{tab:theorem_proof_map} pairs each main-text theorem with its proof in this appendix. The assumptions used throughout are collected in Appendix~\ref{app:assumptions}.

\begin{table}[htbp]
\centering
\caption{Main-text theorems and their corresponding proofs.}
\label{tab:theorem_proof_map}
\small
\renewcommand{\arraystretch}{1.15}
\begin{tabularx}{\linewidth}{@{}l>{\raggedright\arraybackslash}Xl@{}}
\toprule
Theorem & Result & Proof \\
\midrule
Theorem~\ref{thm:advantage-representation}
& Universal advantage function representation
& Appendix~\ref{sec:proof_theorem_3.1} \\
Theorem~\ref{thm:average_case_optimality}
& Average-case optimality of the rank-$k$ representations
& Appendix~\ref{app:proof_average_case_optimality} \\
Theorem~\ref{thm:removing_spectral_floor}
& Removing the immediate-term spectral floor
& Appendix~\ref{sec:immediate_spectral_floor} \\
Theorem~\ref{thm:suaf_policy_improvement_one_step}
& Boltzmann policy improvement with truncated SUAF
& Appendix~\ref{app:proof_one_step_improvement} \\
Theorem~\ref{thm:suaf_policy_improvement}
& Policy improvement using transferred SUAF
& Appendix~\ref{app:proof_transferred_improvement} \\
Theorem~\ref{cor:truncated_suaf_pg}
& Policy-gradient error with truncated SUAF
& Appendix~\ref{app:proof_truncated_suaf_pg} \\
\bottomrule
\end{tabularx}
\end{table}


\subsection{Assumptions}

\label{app:assumptions}

In this work, unless stated clearly, we consider a fixed discounted MDP with a known discount factor \(0<\gamma<1\) and a reference policy \(\pi\). Our construction uses the following assumptions:

\begin{enumerate}
\item \textbf{Access to the reference policy.} We can evaluate \(\pi(a\mid s)\) and sample actions from \(\pi(\cdot\mid s)\). This allows us to center functions across actions at a given state and to define the state--action transitions under \(\pi\).

\item \textbf{Access to transition data.} We have trajectories under \(\pi\), or appropriately weighted transition data, from which the discounted occupancy \(d^\pi\), one-step pairs, and positive-lag pairs can be estimated. The transition probabilities need not be known explicitly.

\item \textbf{Access to downstream rewards.} Representation learning does not use task rewards. When adapting to a task, its bounded reward function \(r(s,a)\) can be evaluated at the state-action pairs needed to compute the immediate reward contrast and the reward-feature coefficients.

\item \textbf{Well-defined feature distributions.} The weighted feature construction is restricted to the supports of \(d^\pi\) and \(\nu^{+,\pi}\), where the required expectations and second moments are finite.

\end{enumerate}

The exact finite-MDP policy-recovery result additionally assumes the full-support and exact-representation conditions stated in Theorem \ref{thm:suaf_oracle_control}.

\subsection{Reason of exclude the immediate term and Proof of Theorem \ref{thm:removing_spectral_floor}}
\label{sec:immediate_spectral_floor}

The immediate reward contrast has a known high-rank structure that need not be learned from future transitions. The following result makes this precise in a common input-output geometry.

\begin{lemma}[Spectral floor induced by the immediate term]
\label{lem:immediate_spectral_floor}
Throughout this lemma, both rewards and advantages belong to $\mathcal H=L^2(d^\pi)$, and all adjoints, singular values, and operator norms refer to this space. Consider a finite MDP with $0<\gamma<1$ and $d^\pi(s,a)>0$ for all state-action pairs. Let $P_\pi$ be the state-action transition operator and define the unnormalized-return advantage operator and its delayed component by
\[
\mathcal A_\pi=C_\pi(I-\gamma P_\pi)^{-1},
\qquad
\mathcal D_\pi=\mathcal A_\pi-C_\pi
=\frac{\gamma}{1-\gamma}\mathcal T^{+,\pi}.
\]
Set $q=\operatorname{rank}(C_\pi)=|\mathcal S|(|\mathcal A|-1)$ and let $\tau_1\ge\cdots\ge\tau_q\ge0$ be the singular values of $\mathcal D_\pi$, padded with zeros. Then $\mathcal A_\pi$ has exactly $q$ nonzero singular values, given by
\begin{equation}
\label{eq:immediate_spectral_floor}
\sigma_i(\mathcal A_\pi)=\sqrt{1+\tau_i^2}, \qquad i=1,\ldots,q.
\end{equation}
Consequently, for $0\le k<q$,
\begin{align}
\inf_{\operatorname{rank}(K)\le k}
\|\mathcal A_\pi-K\|_{\mathrm{op}}^2
&=1+\tau_{k+1}^2\geq 1,\label{eq:full_advantage_rank_error}\\
\inf_{\operatorname{rank}(K)\le k}
\|\mathcal A_\pi-(C_\pi+K)\|_{\mathrm{op}}^2
&=\tau_{k+1}^2.\label{eq:delayed_advantage_rank_error}
\end{align}
\end{lemma}

\begin{proof}
Since $d^\pi(s,a)=d_S^\pi(s)\pi(a\mid s)$, the action-centering operator $C_\pi$ is the orthogonal projection onto $\mathcal H_C=\{u:\mathbb E_{a\sim\pi(\cdot\mid s)}u(s,a)=0\text{ for all }s\}$. The next action is sampled from $\pi$, so $P_\pi C_\pi=0$. Hence $(I-\gamma P_\pi)^{-1}C_\pi=C_\pi$ and
\[
\mathcal D_\pi C_\pi=0,
\qquad C_\pi\mathcal D_\pi=\mathcal D_\pi.
\]
Using $C_\pi^*=C_\pi=C_\pi^2$ therefore gives
\[
\mathcal A_\pi\mathcal A_\pi^*
=(C_\pi+\mathcal D_\pi)(C_\pi+\mathcal D_\pi)^*
=C_\pi+\mathcal D_\pi\mathcal D_\pi^*.
\]
On $\mathcal H_C$, this is $I+\mathcal D_\pi\mathcal D_\pi^*$. On $\mathcal H_C^\perp$, it is zero. This proves \eqref{eq:immediate_spectral_floor}. The two approximation identities follow from best rank-$k$ approximation in operator norm, applying it to $\mathcal A_\pi$ and $\mathcal D_\pi$, respectively.
\end{proof}

Therefore, even when the delayed component is low rank, the full advantage operator remains a spectral floor of one on the $q$ action-centered directions. Retaining $C_\pi r$ exactly avoids spending learned representation capacity on this known immediate structure. The comparison allocates rank $k$ to the learned component. $C_\pi+K$ itself need not have rank $k$. The unit floor concerns unnormalized returns in the common $L^2(d^\pi)$ geometry: the immediate term in the normalized successor measure instead has coefficient $1-\gamma$. Accordingly, the result motivates separating the immediate term without asserting a universal spectral-decay ordering between matrices normalized using different marginals. Note that \citet{ollivier2025featuresbestsuccessorfeatures} uses the spectrum analysis of $\mathcal{A}_\pi$, which introduce a non-zero error in the approximation of $\mathcal{A}_\pi$.

Based on this, we here prove the Theorem \ref{thm:removing_spectral_floor}.

\begin{theorem}[Removing the immediate-term spectral floor]
Consider a finite MDP with $0<\gamma<1$ and strictly positive
$d^\pi$ and $\nu^{+,\pi}$.
Let
\[
\alpha_\pi
=
\min_s\frac{d_S^\pi(s)}{\nu_S^{+,\pi}(s)},
\qquad
q=|\mathcal S|(|\mathcal A|-1).
\]
For every $0\le k<q$,
\begin{align}
&\inf_{\operatorname{rank}(K)\le k}
\sup_{\|r\|_{\nu^{+,\pi}}\le1}
\|A_r^\pi-Kr\|_{d^\pi}^2
\ge
\alpha_\pi+
\left(\frac{\gamma}{1-\gamma}\right)^2
\left(\sigma_{k+1}^{+,\pi}\right)^2,
\end{align}
where singular values beyond the rank of
$\gT^{+,\pi}$ are defined to be zero.
In contrast, retaining $C_\pi r$ exactly and truncating
only the future term incurs worst-case error without $\alpha_\pi$ as shown in Theorem \ref{thm:average_case_optimality}.
If $\nu^{+,\pi}=d^\pi$, then $\alpha_\pi=1$
and \eqref{eq:full_advantage_floor} holds with equality.
\end{theorem}

\begin{proof}
All adjoints and singular values below are taken between
$L^2(\nu^{+,\pi})$ and $L^2(d^\pi)$.
Write
\[
\mathcal A_\pi=C_\pi+\mathcal D_\pi,
\qquad
\mathcal D_\pi
=\frac{\gamma}{1-\gamma}\gT^{+,\pi}.
\]
Both $d^\pi$ and $\nu^{+,\pi}$ have conditional action
distribution $\pi$. Consequently, the adjoint of
$C_\pi:L^2(\nu^{+,\pi})\to L^2(d^\pi)$ is
\[
C_\pi^*u(s,a)
=
\frac{d_S^\pi(s)}{\nu_S^{+,\pi}(s)}
(C_\pi u)(s,a).
\]
Its range consists of functions with zero
policy-conditional action mean.
Since $P_\pi$ annihilates every such function,
\[
\mathcal D_\pi C_\pi^*=0,
\qquad
C_\pi\mathcal D_\pi^*=0.
\]
Hence
\[
\mathcal A_\pi\mathcal A_\pi^*
=
C_\pi C_\pi^*
+
\mathcal D_\pi\mathcal D_\pi^*.
\]

Let $\mathcal H_C=\operatorname{range}(C_\pi)$,
viewed as a subspace of $L^2(d^\pi)$.
This space has dimension $q$, and on $\mathcal H_C$,
\[
C_\pi C_\pi^*
=
W_\pi,
\qquad
(W_\pi u)(s,a)
=
\frac{d_S^\pi(s)}{\nu_S^{+,\pi}(s)}u(s,a).
\]
In particular, $W_\pi\succeq\alpha_\pi I$.
Both summands vanish on $\mathcal H_C^\perp$.
Eigenvalue monotonicity therefore gives, for $i\le q$,
\[
\sigma_i(\mathcal A_\pi)^2
\ge
\alpha_\pi+\sigma_i(\mathcal D_\pi)^2
=
\alpha_\pi+
\left(\frac{\gamma}{1-\gamma}\right)^2
\left(\sigma_i^{+,\pi}\right)^2.
\]
The best rank-$k$ operator-norm approximation error is
$\sigma_{k+1}(\mathcal A_\pi)^2$, proving the lower bound.
The SUAF error follows from truncating the singular
expansion of $\mathcal D_\pi$.

When $\nu^{+,\pi}=d^\pi$, $W_\pi=I$ on
$\mathcal H_C$, so all the above eigenvalue inequalities
become equalities.
\end{proof}

\subsection{Proof of Theorem \ref{thm:average_case_optimality}}
\label{app:proof_average_case_optimality}
\begin{theorem}[Average-case optimality of the rank-$k$ representations]
Let reward function $r$ be sampled from uniform distribution over the unit sphere of $L^2(\nu^{+,\pi})$, and let $m=|\operatorname{supp}(\nu^{+,\pi})|$. Among all reward-independent rank-$k$ approximations of the positive-lag future operator $\gT^{+,\pi}$
in $A_r^\pi$, the top-$k$ truncated representation
\begin{equation*}
{A}_{r,k}^\pi
=
C_\pi r
+
\frac{\gamma}{1-\gamma}
\sum_{i=1}^k
\sigma_i^{+,\pi}f_i
\langle g_i,r\rangle_{\nu^{+,\pi}}
\end{equation*}
minimizes the expected advantage approximation error, i.e.,
$$
\mathbb E_r
\left[
\left\|A_r^\pi-{A}_{r,k}^\pi\right\|_{d^\pi}^2
\right]
= \inf_{\text{rank}(\widehat{\gT}^{+,\pi})\leq k}
\mathbb E_r
\left[
\left\|A_r^\pi-\hat{A}_{r}^\pi\right\|_{d^\pi}^2
\right]
= \frac{1}{m}
\left(\frac{\gamma}{1-\gamma}\right)^2
\sum_{i>k}\left(\sigma_i^{+,\pi}\right)^2,
$$
and top-k truncation is also optimal under minimax case, i.e.,
\begin{equation*}
    \sup_{\| r \|_{\nu^{+,\pi}}\leq 1}\left\|A_r^\pi-{A}_{r,k}^\pi\right\|_{d^\pi}^2 
    = \inf_{\text{rank}(\widehat{\gT}^{+,\pi})\leq k}\sup_{\| r \|_{\nu^{+,\pi}}\leq 1} \left\|A_r^\pi-\hat{A}_{r}^\pi\right\|_{d^\pi}^2
    = \left(\frac{\gamma}{1-\gamma}\right)^2 \left(\sigma^{+,\pi}_{k+1}\right)^2.
\end{equation*}
\end{theorem}
\begin{proof}
For brevity, let
$$
\mathcal H_\nu=L^2(\nu^{+,\pi}),
\qquad
\mathcal H_d=L^2(d^\pi),
$$
where both spaces are restricted to the supports of their corresponding distributions. Define the coordinate maps
$$
[\mathcal U_\nu r](s^+,a^+)
=
\sqrt{\nu^{+,\pi}(s^+,a^+)}\,r(s^+,a^+),\quad
[\mathcal U_d h](s,a)
=
\sqrt{d^\pi(s,a)}\,h(s,a).
$$
These maps are isometries because
$$
\|\mathcal U_\nu r\|_2=\|r\|_{\nu^{+,\pi}},
\qquad
\|\mathcal U_d h\|_2=\|h\|_{d^\pi}.
$$
By the definitions of $\gT^{+,\pi}$ and the positive-lag DCDM, for every
$r\in\mathcal H_\nu$,
\begin{align*}
[\mathcal U_d\gT^{+,\pi}r](s,a)
=
\sum_{s^+,a^+}
\widetilde{\boldsymbol B}^{+,\pi}
\big((s,a),(s^+,a^+)\big)
[\mathcal U_\nu r](s^+,a^+).
\end{align*}
Therefore, $\gT^{+,\pi}$ and $\widetilde{\boldsymbol B}^{+,\pi}$ have the
same rank and singular values
$$
\mathcal U_d\gT^{+,\pi}\mathcal U_\nu^{-1}
=
\widetilde{\boldsymbol B}^{+,\pi}.
$$

In particular, the SVD of the DCDM implies
$$
\gT^{+,\pi}r
=
\sum_{i=1}^{R}
\sigma_i^{+,\pi}
f_i \langle g_i, r\rangle_{\nu^{+,\pi}}.
$$
Its rank-$k$ truncation is
$$
\gT_k^{+,\pi}r
=
\sum_{i=1}^{k}
\sigma_i^{+,\pi}
f_i\langle g_i, r\rangle_{\nu^{+,\pi}}.
$$
Now let $\widehat{\gT}^{+,\pi}$ be any reward-independent operator with
$\operatorname{rank}(\widehat{\gT}^{+,\pi})\leq k$, and define
$$
\widehat{\boldsymbol B}
=
\mathcal U_d
\widehat{\gT}^{+,\pi}
\mathcal U_\nu^{-1}.
$$
Since $\mathcal U_d$ and $\mathcal U_\nu$ are invertible isometries,
$$
\operatorname{rank}(\widehat{\boldsymbol B})
=
\operatorname{rank}(\widehat{\gT}^{+,\pi})
\leq k.
$$
Moreover, because the immediate term $C_\pi r$ is retained exactly,
$$
A_r^\pi-\widehat A_r^\pi
=
\frac{\gamma}{1-\gamma}
\left(
\gT^{+,\pi}-\widehat{\gT}^{+,\pi}
\right)r.
$$
For the average-case result, let
$$
\boldsymbol z_r=\mathcal U_\nu r.
$$
Since $\mathcal U_\nu$ is an isometry, if $r$ is uniformly distributed over the unit sphere of $\mathcal H_\nu$, then $\boldsymbol z_r$ is uniformly distributed over the Euclidean unit sphere in $\mathbb R^m$. Hence,
$$
\mathbb E_r
\left[
\boldsymbol z_r\boldsymbol z_r^{\mathrm T}
\right]
=
\frac{1}{m}\boldsymbol I_m.
$$
It follows that
\begin{align*}
\mathbb E_r
\left[
\left\|
A_r^\pi-\widehat A_r^\pi
\right\|_{d^\pi}^2
\right]
&=
\left(\frac{\gamma}{1-\gamma}\right)^2
\mathbb E_r
\left[
\left|
\left(
\widetilde{\boldsymbol B}^{+,\pi}
-
\widehat{\boldsymbol B}
\right)
\boldsymbol z_r
\right|_2^2
\right]\\
&=
\left(\frac{\gamma}{1-\gamma}\right)^2
\operatorname{Tr}
\left[
\left(
\widetilde{\boldsymbol B}^{+,\pi}
-
\widehat{\boldsymbol B}
\right)^{\mathrm T}
\left(
\widetilde{\boldsymbol B}^{+,\pi}
-
\widehat{\boldsymbol B}
\right)
\mathbb E_r
\left[
\boldsymbol z_r\boldsymbol z_r^{\mathrm T}
\right]
\right]
\\
&=
\frac{1}{m}
\left(\frac{\gamma}{1-\gamma}\right)^2
\left\|
\widetilde{\boldsymbol B}^{+,\pi}
-
\widehat{\boldsymbol B}
\right\|_{\mathrm F}^2.
\end{align*}
By the Eckart-Young-Mirsky theorem \cite{eckart1936approximation},
$$
\inf_{\operatorname{rank}(\widehat{\boldsymbol B})\leq k}
\left\|
\widetilde{\boldsymbol B}^{+,\pi}
-
\widehat{\boldsymbol B}
\right\|_{\mathrm F}^2
=
\sum_{i>k}
\left(\sigma_i^{+,\pi}\right)^2,
$$
and the infimum is attained by the truncated SVD
$$
\widetilde{\boldsymbol B}^{+,\pi}_k
=
\sum_{i=1}^k
\sigma_i^{+,\pi}
\boldsymbol\psi_{i,+,\pi}^{S,A}
\left(
\boldsymbol\psi_{i,+,\pi}^{S^+,A^+}
\right)^{\mathrm T}.
$$
This matrix corresponds exactly to $\gT_k^{+,\pi}$ and therefore to
$A_{r,k}^\pi$. Consequently,
$$
\inf_{\operatorname{rank}(\widehat{{\gT}^{+,\pi}})\leq k}
\mathbb E_r
\left[
\left\|
A_r^\pi-\widehat A_r^\pi
\right\|_{d^\pi}^2
\right]
=
\frac{1}{m}
\left(\frac{\gamma}{1-\gamma}\right)^2
\sum_{i>k}
\left(\sigma_i^{+,\pi}\right)^2.
$$
For the minimax result, for any
$\widehat{\gT}^{+,\pi}$ with rank at most $k$,
\begin{align*}
\sup_{\|r\|*{\nu^{+,\pi}}\leq 1}
\left\|
A_r^\pi-\widehat A_r^\pi
\right\|_{d^\pi}^2
&=
\left(\frac{\gamma}{1-\gamma}\right)^2
\sup_{\|r\|_{\nu^{+,\pi}}\leq 1}
\left\|
\left(
\gT^{+,\pi}-\widehat{\gT}^{+,\pi}
\right)r
\right\|_{d^\pi}^2\\
&=
\left(\frac{\gamma}{1-\gamma}\right)^2
\left\|
\gT^{+,\pi}-\widehat{\gT}^{+,\pi}
\right\|_{\mathrm{op}}^2
\\
&=
\left(\frac{\gamma}{1-\gamma}\right)^2
\left\|
\widetilde{\boldsymbol B}^{+,\pi}
-
\widehat{\boldsymbol B}
\right\|_{2}^2,
\end{align*}
where $\|\cdot\|_{\mathrm{op}}$ and $\|\cdot\|_2$ denote the induced
operator norm and the matrix spectral norm, respectively. Applying the
Eckart--Young--Mirsky theorem in the spectral norm gives
$$
\inf_{\operatorname{rank}(\widehat{\boldsymbol B})\leq k}
\left\|
\widetilde{\boldsymbol B}^{+,\pi}
-
\widehat{\boldsymbol B}
\right\|_2
=
\sigma_{k+1}^{+,\pi}.
$$
The truncated matrix
$\widetilde{\boldsymbol B}^{+,\pi}_k$ attains this infimum. Hence,
$$
\sup_{\|r\|_{\nu^{+,\pi}}\leq 1}
\left\|
A_r^\pi-A_{r,k}^\pi
\right\|_{d^\pi}^2
=
\inf_{\operatorname{rank}(\widehat{\gT}^{+,\pi})\leq k}
\sup_{\|r\|_{\nu^{+,\pi}}\leq 1}
\left\|
A_r^\pi-\widehat A_r^\pi
\right\|_{d^\pi}^2
=
\left(\frac{\gamma}{1-\gamma}\right)^2
\left(\sigma_{k+1}^{+,\pi}\right)^2.
$$
When $k<R$, the worst-case error of the truncated representation is
attained, for example, by $r=g_{k+1}$. If $k\geq R$, both the
average-case and minimax errors are zero, with the convention
$\sigma_{k+1}^{+,\pi}=0$.
\end{proof}

\subsection{Proof of Theorem \ref{thm:advantage-representation}}
\label{sec:proof_theorem_3.1}

\begin{theorem}[Universal advantage function representation]
For any reward function \(r:\mathcal S\times\mathcal A\to\mathbb R\), and for every \((s,a)\) in the support of \(d^\pi\), the advantage function admits the exact representation
\begin{equation*}
A_r^\pi(s,a)
=
C_\pi r(s,a)
+
\frac{\gamma}{1-\gamma}
\sum_{i=1}^{R} \sigma_i^{+,\pi} f_i(s,a) \left\langle g_i,r\right\rangle_{\nu^{+,\pi}},
\end{equation*}
where $C_\pi r(s,a):=r(s,a)-\mathbb E_{a\sim\pi(\cdot\mid s)}\left[r(s,a)\right]$ is the immediate action-centered reward and $\left\langle g_i,r\right\rangle_{\nu^{+,\pi}}\!=\!\sum_{s,a}\!\nu^{+,\pi}(s,a)g_i(s,a)r(s,a)$ is the inner product on Hilbert space $L^2(\nu^{+,\pi})$.
Moreover, for any \(k\leq R\), define the rank-\(k\) approximation
\begin{equation*}
A_{r,k}^\pi(s,a)
:=
C_\pi r(s,a)
+
\frac{\gamma}{1-\gamma}
\sum_{i=1}^{k}
\sigma_i^{+,\pi}
f_i(s,a)
\beta_i^{+,\pi}(r).
\end{equation*}
Its approximation error is given exactly by
\begin{equation*}
\left\|
A_r^\pi-A_{r,k}^\pi
\right\|_{L^2(d^\pi)}^2
=
\left(\frac{\gamma}{1-\gamma}\right)^2
\sum_{i=k+1}^{R}
\left(\sigma_i^{+,\pi}\right)^2
\left(\beta_i^{+,\pi}(r)\right)^2.
\end{equation*}
\end{theorem}

\begin{proof}
The SVD of \(\widetilde{\boldsymbol B}^{+,\pi}\), together with the definitions of \(f_i\) and \(g_i\), gives
\begin{align}
&M^{+,\pi}(s^+,a^+\mid s,a)
-
\mathbb E_{a'\sim\pi(\cdot\mid s)}
\left[
M^{+,\pi}(s^+,a^+\mid s,a')
\right]
\nonumber\\
&\qquad=
\nu^{+,\pi}(s^+,a^+)
\sum_{i=1}^{R}
\sigma_i^{+,\pi}
f_i(s,a)
g_i(s^+,a^+).
\label{eq:successor-difference-spectral}
\end{align}
Multiplying both sides by \(r(s^+,a^+)\) and summing over \((s^+,a^+)\) yields
\begin{align}
&\sum_{s^+,a^+}
\Bigg(\!\!
M^{+,\pi}(s^+\!,\!a^+| s,a)
\!-\!
\mathbb E_{a'\sim\pi(\cdot| s)}
\!\!\left[
M^{+,\pi}\!(s^+\!,\!a^+| s,a')
\right]
\!\!\!\Bigg)
\!r(s^+\!,\!a^+)\notag\\
&\qquad =
\sum_{i=1}^{R}
\!\sigma_i^{+,\pi}
f_i(s,a)
\beta_i^{+,\pi}(r).
\label{eq:future-reward-readout}
\end{align}
By the definition of the normalized positive-lag successor measure,
\begin{equation}
\frac{\gamma}{1-\gamma}
\sum_{s^+,a^+}
M^{+,\pi}(s^+,a^+\mid s,a)
r(s^+,a^+)
=
\sum_{t=1}^{\infty}
\gamma^t
\mathbb E_\pi
\left[
r(S_t,A_t)
\mid S_0=s,A_0=a
\right].
\end{equation}
Therefore,
\begin{equation}
Q_r^\pi(s,a)
=
r(s,a)
+
\frac{\gamma}{1-\gamma}
\sum_{s^+,a^+}
M^{+,\pi}(s^+,a^+\mid s,a)
r(s^+,a^+).
\end{equation}
Averaging this identity over \(a'\sim\pi(\cdot\mid s)\) gives \(V_r^\pi(s)\). Subtracting \(V_r^\pi(s)\) from \(Q_r^\pi(s,a)\) and applying \eqref{eq:future-reward-readout} proves \eqref{eq:advantage-spectral-representation}.

Finally, the functions \(\{f_i\}_{i=1}^{R}\) are orthonormal in \(L^2(d^\pi)\):
\begin{equation}
\mathbb E_{(S,A)\sim d^\pi}
\left[
f_i(S,A)f_j(S,A)
\right]
=
\delta_{ij}.
\end{equation}
Hence the omitted spectral components are mutually orthogonal, and Parseval's identity gives
\begin{equation}
\left\|
A_r^\pi-A_{r,k}^\pi
\right\|_{L^2(d^\pi)}^2
=
\left(\frac{\gamma}{1-\gamma}\right)^2
\sum_{i=k+1}^{R}
\left(\sigma_i^{+,\pi}\right)^2
\left(\beta_i^{+,\pi}(r)\right)^2.
\end{equation}
\end{proof}

\subsection{SUAF will Find Optimal Policy in Zero-shot Manner}
\label{app:suaf_oracle_control}

\paragraph{Conventions and assumptions.}
Let $n=|\mathcal S||\mathcal A|$, and represent rewards and normalized
discounted occupancies as column vectors in $\mathbb R^n$.
In particular,
\[
 d^\pi(s,a)=(1-\gamma)\sum_{t\ge0}\gamma^t
 \Pr_{\mu,\pi}(s_t=s,a_t=a),
 \quad J_r(\pi)=\frac{r^\top d^\pi}{1-\gamma}.
\]
Optimality throughout is with respect to the fixed initial distribution
$\mu$. At states with zero occupancy, the recovered policy can be
chosen arbitrarily. Full support of $d_0$ and $\nu$ makes all weighted
feature coordinates well-defined. The statements concern exact SUAF
features, exact reference distributions, and exact optimization.

Define the state lifting matrix $H\in\mathbb R^{n\times|\mathcal S|}$,
the reference-policy averaging matrix
$\Pi_0\in\mathbb R^{|\mathcal S|\times n}$, and the environment transition
matrix $T\in\mathbb R^{n\times|\mathcal S|}$ by
\begin{align*}
 (Hw)(s,a)&=w(s),\\
 (\Pi_0x)(s)&=\sum_a\pi_0(a\mid s)x(s,a),\\
 (Tw)(s,a)&=\sum_{s'}T(s'\mid s,a)w(s').
\end{align*}
Thus $\Pi_0H=I$, $P_0=T\Pi_0$, and $C_0=I-H\Pi_0$.
The full advantage operator and its delayed component are
\[
 \mathcal A_0=C_0(I-\gamma P_0)^{-1},\qquad
 D_0=\mathcal A_0-C_0=cF\Sigma G^\top W_\nu.
\]
Here the positive-lag outcome distribution is
$\nu^\top=d_0^\top M^{+,\pi_0}$, with
$M^{+,\pi_0}=(1-\gamma)P_0(I-\gamma P_0)^{-1}$.
The feature normalizations are
\[
 F^\top W_dF=I_R,\qquad G^\top W_\nu G=I_R,
 \qquad C_0F=F.
\]
All transposes are ordinary coordinate transposes.

\paragraph{Flow constraints and policy recovery.}
Set $B=H^\top-\gamma T^\top$. The feasible occupancy set is
\begin{equation}
\label{eq:app_flow_polytope}
 \mathcal D_\mu=\{d\ge0:Bd=(1-\gamma)\mu\}.
\end{equation}
Necessity follows from the discounted occupancy identity.
For sufficiency, given such a $d$, put $d_S(s)=\sum_a d(s,a)$ and
define $\pi_d(a\mid s)=d(s,a)/d_S(s)$ whenever $d_S(s)>0$.
At a zero-mass state choose any action distribution. Then
$d(s,a)=d_S(s)\pi_d(a\mid s)$ everywhere, and the flow constraint becomes
\[
 d_S=(1-\gamma)\mu+\gamma(P_S^{\pi_d})^\top d_S.
\]
Its unique solution is the normalized discounted state occupancy of
$\pi_d$, since $I-\gamma(P_S^{\pi_d})^\top$ is invertible. Thus
$d=d^{\pi_d}$. Summing the flow constraints also gives
$\mathbf 1^\top d=1$, so $\mathcal D_\mu$ is compact.

A simple occupancy-flow identity makes SUAF sufficient for control. In a finite MDP with $0<\gamma<1$, fix $\mu$ and $\pi_0$ with full-support $d_0=d^{\pi_0}$ and $\nu=\nu^{+,\pi_0}$.
Let $F,G,\Sigma$ collect all $R$ SUAF modes, and write
$W_d=\operatorname{diag}(d_0)$, $W_\nu=\operatorname{diag}(\nu)$,
and $c=\gamma/(1-\gamma)$.
\begin{theorem}[Policy recovery]
\label{thm:suaf_oracle_control}
Let $Z$ be a given basis matrix for the subspace with
$\sum_a z(s,a)=0$ for every $s$, with $q=|\mathcal S|(|\mathcal A|-1)$ columns. Define 
\begin{equation}
\label{eq:suaf_occupancy_basis}
\Phi=Z+cW_\nu G\Sigma F^\top Z,
\quad
\mathcal D_\mu
=\{d_0+\Phi w:\ w\in\mathbb R^q,\ d_0+\Phi w\ge0\},
\end{equation}
then $\mathcal D_\mu$ is the set of all stationary-policy
occupancies. For any reward $r$, maximizing
$r^\top(d_0+\Phi w)$ over this set and normalizing the optimizer
statewise recovers a $\mu$-optimal policy.
\end{theorem}
The matrix $Z$ is prescribed, with correction having rank at most $R\le|\mathcal S|-1$. SUAF requires no further environment interaction. With fewer modes, we restrict search to feasible directions.

\begin{proof}[Proof of Theorem~\ref{thm:suaf_oracle_control}]
The elementary identities needed for the construction are
\begin{equation}
\label{eq:app_advantage_flow_identities}
 B\mathcal A_0^\top=0,
 \qquad \mathcal A_0C_0=C_0.
\end{equation}
Indeed, $P_0H=T$, and hence
$B^\top=H-\gamma T=(I-\gamma P_0)H$.
Consequently, $\mathcal A_0B^\top=C_0H=0$.
Also, $P_0C_0=T\Pi_0C_0=0$, so
$(I-\gamma P_0)^{-1}C_0=C_0$ and
$\mathcal A_0C_0=C_0^2=C_0$.

Because $H$ has full column rank and $I-\gamma P_0$ is invertible,
$B$ has rank $|\mathcal S|$. Therefore $\dim\ker B=q$.
The columns of $Z$ span $\ker H^\top$, so
$C_0^\top Z=(I-\Pi_0^\top H^\top)Z=Z$.
It follows that
\[
 \mathcal A_0^\top Z
 =C_0^\top Z+D_0^\top Z
 =Z+cW_\nu G\Sigma F^\top Z=\Phi.
\]
\eqref{eq:app_advantage_flow_identities} yields $B\Phi=0$.
Moreover,
\[
 C_0^\top\Phi
 =C_0^\top\mathcal A_0^\top Z
 =(\mathcal A_0C_0)^\top Z
 =Z.
\]
Thus $\operatorname{rank}(\Phi)=q$, since $Z$ has rank $q$.
We conclude that $\operatorname{col}(\Phi)=\ker B$.
As $Bd_0=(1-\gamma)\mu$, every solution of the flow equalities is
uniquely expressible as $d_0+\Phi w$.
Intersecting with the nonnegative orthant proves
\eqref{eq:suaf_occupancy_basis}.

The reward objective satisfies $r^\top d=(1-\gamma)J_r(\pi_d)$.
Optimization over the displayed representation is therefore exactly
optimization over all stationary-policy occupancies. The maximum is
attained by compactness, and statewise normalization recovers a
$\mu$-optimal stationary policy. In a finite discounted MDP,
stationary policies suffice for optimality.
\end{proof}

\paragraph{Why the learned correction has rank at most
$|\mathcal S|-1$.}
Using the resolvent identity,
\[
 D_0=\gamma C_0T(I-\gamma\Pi_0T)^{-1}\Pi_0.
\]
Since $T\mathbf 1_{|\mathcal S|}=\mathbf 1_n$ and
$C_0\mathbf 1_n=0$, the matrix $C_0T$ has rank at most
$|\mathcal S|-1$. Hence
\[
 R=\operatorname{rank}(D_0)\le |\mathcal S|-1,
 \qquad \operatorname{rank}(\Phi-Z)\le R.
\]
For example, $Z$ can be constructed from columns
$e_{(s,a)}-e_{(s,a_0)}$, one for each state and each nonreference
action. Thus $Z$ requires no transition-model estimation. This rank
bound concerns the learned correction, not the $q$ free coordinates
of the full occupancy polytope.

\begin{corollary}[Nested feasible policy search]
\label{cor:suaf_nested_control}
Under the same assumptions, define $u_i=W_df_i+c\sigma_iW_\nu g_i$ and $U_k=[u_1,\ldots,u_k]$.
Every nonnegative $d_0+U_k\alpha$ is a valid occupancy.
Let $\pi_{r,k}$ be recovered from an exact solution of
$\max_{\alpha:\,d_0+U_k\alpha\ge0}
r^\top(d_0+U_k\alpha)$.
Then, for $0\le k<R$,
\begin{equation}
\label{eq:suaf_nested_returns}
J_r(\pi_0)
\le J_r(\pi_{r,k})
\le J_r(\pi_{r,k+1})
\le \max_\pi J_r(\pi).
\end{equation}
\end{corollary}

However, the improvement is non-strict and retaining all $R$ directions alone need not span $\mathcal D_\mu$.

\begin{proof}[Proof of Corollary~\ref{cor:suaf_nested_control}]
Because $d_0(s,a)=d_{0,S}(s)\pi_0(a\mid s)$, action centering is
self-adjoint in $L^2(d_0)$:
$C_0^\top W_d=W_dC_0$.
For each exact SUAF mode, $C_0f_i=f_i$ and
$F^\top W_df_i=e_i$. Therefore
\begin{align*}
 \mathcal A_0^\top W_df_i
 &=C_0^\top W_df_i+cW_\nu G\Sigma F^\top W_df_i\\
 &=W_df_i+c\sigma_iW_\nu g_i=u_i.
\end{align*}
\eqref{eq:app_advantage_flow_identities} gives $Bu_i=0$.
Consequently, $d_0+U_k\alpha$ always satisfies the exact flow
equalities and is an occupancy if and only if it is nonnegative.

Define
\[
 \mathcal D_k=(d_0+\operatorname{col}(U_k))\cap\mathbb R_+^n,
 \qquad \mathcal D_0=\{d_0\}.
\]
These are nonempty compact subsets of $\mathcal D_\mu$, and
\[
 \{d_0\}=\mathcal D_0\subseteq\mathcal D_1
 \subseteq\cdots\subseteq\mathcal D_R\subseteq\mathcal D_\mu.
\]
Maximizing the same linear reward objective over these nested sets,
and then dividing by $1-\gamma$, proves
\eqref{eq:suaf_nested_returns}. No claim of strict improvement or of optimal mode selection for control is required.
\end{proof}

\paragraph{Completion to a globally sufficient search space.}
The $R$ nonzero future modes need not supply all $q$ occupancy
directions. Define
\[
 \mathcal N=\ker H^\top\cap\ker D_0^\top.
\]
For $z\in\mathcal N$, $C_0^\top z=z$ and
$\mathcal A_0^\top z=z$, so $z\in\ker B$.
Since $D_0^\top=cW_\nu G\Sigma F^\top$ has rank $R$ and
$F^\top W_dF=I_R$, its restriction to $\ker H^\top$ also has rank
$R$. Indeed, each $W_df_i$ belongs to $\ker H^\top$.
Consequently $\dim\mathcal N=q-R$ and
\[
 \ker H^\top=\operatorname{span}\{W_df_i\}_{i=1}^R
 \oplus\mathcal N.
\]
The map $\mathcal A_0^\top$ is a bijection from $\ker H^\top$ to
$\ker B$, as follows from the proof of the theorem. Applying it
to the preceding decomposition gives
\[
 \ker B=\operatorname{span}\{u_i\}_{i=1}^R\oplus\mathcal N.
\]
Thus adding a basis of $\mathcal N$ makes the search globally
sufficient at $k=R$. Identifying this completion generally requires
the full exact representation. It is not implied by knowing only
the first $k<R$ pairs. Alternatively, the full construction with $Z$
in Theorem~\ref{thm:suaf_oracle_control} supplies all directions
directly.

\paragraph{Scope.}
These are exact SUAFs. With learned features, errors can break the identity $Bu_i=0$, and nonnegativity alone no longer certifies feasibility in the true MDP. Furthermore, monotonicity in $k$ follows from nesting, not from a theorem that leading singular modes minimize control regret. The construction does not assert that $R$ is the dimension of the entire policy space or that it is directly comparable to the number of basis matrices in a successor-measure representation.

\subsection{One step update with SUAF}
\label{app:proof_one_step_improvement}

\begin{lemma}[Boltzmann update satisfies the improvement condition]
\label{lem:boltzmann_improvement}
Let \(\pi\) be a behavior policy, and let \(A_{r,k}^{\pi}\) be an advantage score satisfying the centering condition
\begin{equation}
\mathbb E_{a\sim \pi(\cdot\mid s)}
\big[A_{r,k}^{\pi}(s,a)\big]
=
0
\qquad
\forall s.
\label{eq:centered_advantage}
\end{equation}
For any inverse temperature \(\tau>0\), define the Boltzmann update
\begin{equation}
\pi_r^+(a\mid s)
=
\frac{
\pi(a\mid s)\exp\big(\tau A_{r,k}^{\pi}(s,a)\big)
}{
Z^\pi(s)
},
Z^\pi(s)
=
\int
\pi(a'\mid s)
\exp\big(\tau A_{r,k}^{\pi}(s,a')\big)
\,da'.
\label{eq:boltzmann_update}
\end{equation}
Then
\begin{equation}
\mathbb E_{a\sim \pi_r^+(\cdot\mid s)}
\big[A_{r,k}^{\pi}(s,a)\big]
\ge
0
\qquad
\forall s.
\label{eq:boltzmann_improvement}
\end{equation}
\end{lemma}

\begin{proof}
Fix \(s\). Write \(\mathbb E_\pi[\cdot]
=
\mathbb E_{a\sim \pi(\cdot\mid s)}[\cdot]\)
and \(A=A_{r,k}^{\pi}(s,a)\).
From \eqref{eq:boltzmann_update},
\[
\pi_r^+(a\mid s)
=
\frac{\pi(a\mid s)e^{\tau A}}{Z^\pi(s)},
\qquad
Z^\pi(s)=\mathbb E_\pi[e^{\tau A}].
\]
Thus
\[
\mathbb E_{\pi_r^+}[A]
=
\frac{\mathbb E_\pi[A e^{\tau A}]}{\mathbb E_\pi[e^{\tau A}]}.
\]
Equivalently, using the KL identity,
\[
\tau\,\mathbb E_{\pi_r^+}[A]
=
\mathrm{KL}\big(\pi_r^+(\cdot\mid s)\,\|\,\pi(\cdot\mid s)\big)
+
\log \mathbb E_\pi[e^{\tau A}].
\]
Since \(\log\) is concave, Jensen's inequality gives
\[
\log \mathbb E_\pi[e^{\tau A}]
\ge
\tau\,\mathbb E_\pi[A]
=
0,
\]
where the last equality follows from \eqref{eq:centered_advantage}. Hence
\[
\tau\,\mathbb E_{\pi_r^+}[A]
\ge
\mathrm{KL}\big(\pi_r^+(\cdot\mid s)\,\|\,\pi(\cdot\mid s)\big)
\ge
0.
\]
Dividing by \(\tau>0\) yields \eqref{eq:boltzmann_improvement}.
\end{proof}

\begin{remark}[Automatic centering in the SUAF construction]
\label{rem:centering_suaf}
In our construction,
\[
A_{r,k}^{\pi}
=
C_\pi r
+
\frac{\gamma}{1-\gamma}
\sum_{i=1}^k
\sigma_i^{+,\pi}f_i
\langle g_i,r\rangle_{\nu^{+,\pi}}.
\]
By definition of \(C_\pi\), the term \(C_\pi r\) is centered under \(\pi\). Moreover, the property \(f_i=C_\pi f_i\) ensures that the remaining term is also centered under \(\pi\). Therefore
\[
\mathbb E_{a\sim\pi(\cdot\mid s)}
\big[A_{r,k}^{\pi}(s,a)\big]
=
0
\qquad
\forall s.
\]
Hence the centering condition \eqref{eq:centered_advantage} in Lemma~\ref{lem:boltzmann_improvement} is automatically satisfied, and the Boltzmann update \eqref{eq:boltzmann_update} satisfies the improvement condition \eqref{eq:boltzmann_improvement}.
\end{remark}

\begin{theorem}[Policy improvement with truncated SUAF]
Consider a policy $\pi_r^+$ using Boltzmann update and greedy update.
Assume $d^{\pi_r^+}\ll d^\pi$ and define $\chi_k^2 := \mathbb E_{d^\pi} [(\frac{d^{\pi_r^+}}{d^\pi}-1)^2]<\infty $ and $\mathcal I_k(r):= \mathbb E_{d^{\pi_r^+}}[A_{r,k}^{\pi}]/(1-\gamma)$.
Then $\mathcal I_k(r)\ge0$ and
\[
    J_r(\pi_r^+)-J_r(\pi)
    \ge
    \mathcal I_k(r)
    -
    \frac{\gamma\sqrt{\chi_k^2}}{(1-\gamma)^2}
    \left[
        \sum_{i>k}
        (\sigma_i^{+,\pi})^2\beta_i(r)^2
    \right]^{1/2}.
\]
Consequently, policy improvement is guaranteed whenever the
predicted gain exceeds the spectral-tail penalty.
For $k=R$, the penalty vanishes.
\end{theorem}

\begin{proof}
By Lemma \ref{lem:boltzmann_improvement}, we have that $\E_{\pi_r^+}[A_{r,k}^\pi(s,a)] \geq 0$ for every $s$. 
Write $e_{r,k}=A_r^\pi-A_{r,k}^\pi$.
The performance difference identity gives
\[
    J_r(\pi_r^+)-J_r(\pi)
    =
    \mathcal I_k(r)
    +
    \frac{1}{1-\gamma}
    \mathbb E_{d^{\pi_r^+}}[e_{r,k}].
\]
Both $A_r^\pi$ and $A_{r,k}^\pi$ are action-centered
under $\pi$, so $\mathbb E_{d^\pi}[e_{r,k}]=0$.
Therefore, Cauchy--Schwarz yields
\[
    \left|\mathbb E_{d^{\pi_r^+}}[e_{r,k}]\right|
    =
    \left|
    \mathbb E_{d^\pi}
    \left[
        \left(\frac{d^{\pi_r^+}}{d^\pi}-1\right)e_{r,k}
    \right]
    \right|
    \le
    \sqrt{\chi_k^2}\,\|e_{r,k}\|_{d^\pi}.
\]
By orthonormality of the left singular functions,
\[
    \|e_{r,k}\|_{d^\pi}^2
    =
    \left(\frac{\gamma}{1-\gamma}\right)^2
    \sum_{i>k}(\sigma_i^{+,\pi})^2\beta_i(r)^2.
\]
Combining these identities proves the claim.
\end{proof}

Actually, we have a stronger version of this theorem as Theorem \ref{thm:suaf_policy_improvement_one_step_formal}. We write it formal as following:

\begin{theorem}[Policy improvement with truncated SUAF Formal]
\label{thm:suaf_policy_improvement_one_step_formal}
Let $h=A_{r,k}^{\pi}$ and $e=A_r^\pi-h$. Let $\pi_r^+$ be
obtained either by a greedy update with respect to $h$, or by the
Boltzmann update
\[
    \pi_\tau(a\mid s)
    =\frac{\pi(a\mid s)\exp(\tau h(s,a))}
           {\mathbb E_{a'\sim\pi(\cdot\mid s)}
            [\exp(\tau h(s,a'))]},
    \qquad \tau>0.
\]
Assume $d^{\pi_r^+}\ll d^\pi$ and
$w:=d^{\pi_r^+}/d^\pi\in L^2(d^\pi)$. Define
\[
    \bar w(s):=\mathbb E_{a\sim\pi(\cdot\mid s)}[w(s,a)],
    \qquad
    \chi_{k,\mathrm{act}}^2
    :=\mathbb E_{d^\pi}\bigl[(w(s,a)-\bar w(s))^2\bigr],
\]
and
$\mathcal I_k(r):=\mathbb E_{d^{\pi_r^+}}[h]/(1-\gamma)$.
Then $\mathcal I_k(r)\geq 0$ and
\[
    J_r(\pi_r^+)-J_r(\pi)
    \geq
    \mathcal I_k(r)
    -
    \frac{\gamma\sqrt{\chi_{k,\mathrm{act}}^2}}{(1-\gamma)^2}
    \left[
        \sum_{i>k}
        (\sigma_i^{+,\pi})^2\beta_i(r)^2
    \right]^{1/2}.
\]
Moreover, $\chi_{k,\mathrm{act}}^2\leq
\chi_k^2:=\mathbb E_{d^\pi}[(w-1)^2]$. Thus the bound is at
least as tight as the bound using $\chi_k^2$, and its penalty
vanishes when $k=R$.

For the Boltzmann update, write $\pi_r^+=\pi_\tau$ and suppose
\[
    B:=\sup_{s,a,b}|h(s,a)-h(s,b)|<\infty,
    \qquad
    D:=\sup_{s,a,b}|A_r^\pi(s,a)-A_r^\pi(s,b)|<\infty.
\]
With $m:=\langle h,A_r^\pi\rangle_{d^\pi}$, we additionally have
\[
    J_r(\pi_\tau)-J_r(\pi)
    \geq
    \frac{\tau}{1-\gamma}
    \left(
        m-\frac{DB^2}{8(1-\gamma)}\tau
    \right).
\]
Consequently, if $m>0$, every
\[
    0<\tau<\frac{8(1-\gamma)m}{DB^2}
\]
guarantees strict policy improvement. In particular,
\[
    m
    =\|h\|_{d^\pi}^2+\langle C^\pi r,e\rangle_{d^\pi}
    \geq
    \|h\|_{d^\pi}^2-\|C^\pi r\|_{d^\pi}\|e\|_{d^\pi}.
\]
For state-only rewards, $C^\pi r=0$ and hence
$m=\|h\|_{d^\pi}^2>0$ whenever $h\neq 0$.
\end{theorem}

\begin{proof}
Both $h$ and $A_r^\pi$ are action-centered under $\pi$, so
$\mathbb E_{a\sim\pi(\cdot\mid s)}h(s,a)
=\mathbb E_{a\sim\pi(\cdot\mid s)}e(s,a)=0$.
A greedy update satisfies
$\mathbb E_{\pi_r^+}[h(s,a)\mid s]=\max_a h(s,a)\geq 0$.
For the Boltzmann update, let
$Z_\tau(s)=\mathbb E_\pi[\exp(\tau h(s,a))\mid s]$.
Then
\[
    \tau\mathbb E_{\pi_\tau}[h(s,a)\mid s]
    =
    D_{\mathrm{KL}}\bigl(\pi_\tau(\cdot\mid s)
                         \Vert\pi(\cdot\mid s)\bigr)
    +\log Z_\tau(s)\geq 0,
\]
where $\log Z_\tau(s)\geq
\tau\mathbb E_\pi[h(s,a)\mid s]=0$ by Jensen's inequality.
Thus $\mathcal I_k(r)\geq 0$ in either case.

The performance-difference identity and the SUAF expansion give
\[
\begin{aligned}
    J_r(\pi_r^+)-J_r(\pi)
    & =\mathcal I_k(r)
      +\frac{1}{1-\gamma}\mathbb E_{d^{\pi_r^+}}[e],\\
    e& =\frac{\gamma}{1-\gamma}
      \sum_{i>k}\sigma_i^{+,\pi}\beta_i(r)f_i,\\
    \|e\|_{d^\pi}^2
    &=\left(\frac{\gamma}{1-\gamma}\right)^2
      \sum_{i>k}(\sigma_i^{+,\pi})^2\beta_i(r)^2.
\end{aligned}
\]
Since $e$ is conditionally action-centered,
\[
    \mathbb E_{d^{\pi_r^+}}[e]
    =\mathbb E_{d^\pi}[we]
    =\mathbb E_{d^\pi}[(w-\bar w)e].
\]
Cauchy--Schwarz proves the first bound. Conditional orthogonality
also gives
\[
    \chi_k^2
    =\mathbb E_{d^\pi}[(w-\bar w)^2]
     +\mathbb E_{s\sim d_S^\pi}[(\bar w(s)-1)^2]
    \geq\chi_{k,\mathrm{act}}^2.
\]

It remains to prove the temperature bound. For each state, define
$F_s(t)=\mathbb E_{a\sim\pi_t(\cdot\mid s)}
[A_r^\pi(s,a)]$, so that $F_s(0)=0$. Differentiating the
Boltzmann distribution yields
\[
    F_s'(t)
    =\operatorname{Cov}_{\pi_t}
       (A_r^\pi(s,\cdot),h(s,\cdot)).
\]
The covariance bound for bounded random variables implies
$|F_s'(t)|\leq DB/4$. Differentiating once more gives
\[
    F_s''(t)
    =\operatorname{Cov}_{\pi_t}
       \left(
           A_r^\pi(s,\cdot),
           \bigl(h(s,\cdot)-\mathbb E_{\pi_t}h(s,\cdot)\bigr)^2
       \right),
    \qquad
    |F_s''(t)|\leq \frac{DB^2}{4}.
\]
Consequently,
\[
    F_s(\tau)\geq
    \tau F_s'(0)-\frac{DB^2}{8}\tau^2,
    \qquad
    |F_s(\tau)|\leq\frac{DB}{4}\tau.
\]
The same covariance bound applied to action-set indicators gives
$\|\pi_\tau(\cdot\mid s)-\pi(\cdot\mid s)\|_1
\leq \tau B/2$. The discounted occupancy equations therefore imply
\[
    \|d_S^{\pi_\tau}-d_S^\pi\|_1
    \leq\frac{\gamma\tau B}{2(1-\gamma)}.
\]
Since
$m=\mathbb E_{s\sim d_S^\pi}[F_s'(0)]$, combining the preceding
bounds yields
\[
\begin{aligned}
    \mathbb E_{s\sim d_S^{\pi_\tau}}[F_s(\tau)]
    &\geq
    \mathbb E_{s\sim d_S^\pi}[F_s(\tau)]
    -\|d_S^{\pi_\tau}-d_S^\pi\|_1
       \sup_s|F_s(\tau)|\\
    &\geq
    \tau m-\frac{DB^2}{8(1-\gamma)}\tau^2.
\end{aligned}
\]
The performance-difference identity now proves the stated
temperature bound.

Finally, the retained SUAF modes are orthogonal in $L^2(d^\pi)$
to the discarded modes. Since
$h=C^\pi r$ plus the retained modes and $e$ consists of discarded
modes, $\langle h,e\rangle_{d^\pi}
=\langle C^\pi r,e\rangle_{d^\pi}$. This proves the expression
and lower bound for $m$.
\end{proof}

Here, we are ready to prove Theorem \ref{thm:suaf_policy_improvement_one_step}.

\begin{theorem}[Boltzmann policy improvement with truncated SUAF]
Let $A_{r,k}^{\pi}$ be the exact rank-$k$ SUAF advantage
approximation, and let $\pi_\tau(a| s) \propto \pi(a| s)e^{\tau h(s,a)}$ with $\tau>0$. Define $B_k:=\sup_{s,a,b}|A_{r,k}^{\pi}(s,a)-A_{r,k}^{\pi}(s,b)|$, $\varepsilon_k(r)=\frac{\gamma}{1-\gamma} \left[\sum_{i>k} (\sigma_i^{+,\pi})^2\beta_i(r)^2\right]^{1/2}$ and $\Delta_k(r):=\|A_{r,k}^{\pi}\|_{d^\pi}^2-\|C^\pi r\|_{d^\pi}\varepsilon_k(r)$
If $\Delta_k(r)>0$, then it guarantees $J_r(\pi_\tau)>J_r(\pi)$ if temperature satisfies
$
0<\tau<
\frac{4(1-\gamma)^2\Delta_k(r)}
     {R_{\max}B_k^2}.
$
For state-only rewards, $C^\pi r=0$ and
$
\Delta_k(r)
=\left(\frac{\gamma}{1-\gamma}\right)^2
  \sum_{i\le k}(\sigma_i^{+,\pi})^2\beta_i(r)^2.
$
In this case, if $\Delta_k(r)>0$, the larger sufficient range
\[
    0<\tau<
    \frac{4\gamma}{R_{\max}B_k^2}
    \sum_{i\le k}(\sigma_i^{+,\pi})^2\beta_i(r)^2
\]
also guarantees strict improvement.
\end{theorem}

\begin{proof}
Write $h=A_{r,k}^{\pi}$, $e=A_r^\pi-h$, and
$m=\langle h,A_r^\pi\rangle_{d^\pi}$. By orthogonality of the
retained and discarded SUAF modes,
\[
    m=\|h\|_{d^\pi}^2+\langle C^\pi r,e\rangle_{d^\pi}
    \geq \|h\|_{d^\pi}^2
         -\|C^\pi r\|_{d^\pi}\|e\|_{d^\pi}
    =\Delta_k(r),
\]
where $\|e\|_{d^\pi}=\varepsilon_k(r)$.

Let
$D=\sup_{s,a,b}|A_r^\pi(s,a)-A_r^\pi(s,b)|$.
Since $|r(s,a)|\leq R_{\max}$,
$|Q_r^\pi(s,a)|\leq R_{\max}/(1-\gamma)$.
Actions at the same state share the same value baseline; hence
\[
    D
    =\sup_{s,a,b}|Q_r^\pi(s,a)-Q_r^\pi(s,b)|
    \leq\frac{2R_{\max}}{1-\gamma}.
\]
Applying the previously established temperature bound gives
\[
\begin{aligned}
    J_r(\pi_\tau)-J_r(\pi)
    &\geq
    \frac{\tau}{1-\gamma}
    \left(m-\frac{DB_k^2}{8(1-\gamma)}\tau\right)\\
    &\geq
    \frac{\tau}{1-\gamma}
    \left(
        \Delta_k(r)
        -\frac{R_{\max}B_k^2}{4(1-\gamma)^2}\tau
    \right).
\end{aligned}
\]
The right-hand side is strictly positive under the first
stated condition on $\tau$.

If $r=r(s)$, then $C^\pi r=0$. Orthonormality of the SUAF
left features therefore gives
\[
    \Delta_k(r)=\|h\|_{d^\pi}^2
    =\left(\frac{\gamma}{1-\gamma}\right)^2
      \sum_{i\leq k}
      (\sigma_i^{+,\pi})^2\beta_i(r)^2.
\]
Furthermore,
$Q_r^\pi(s,a)
=r(s)+\gamma\mathbb E_{s'\sim P(\cdot\mid s,a)}
[V_r^\pi(s')]$.
The immediate reward cancels when comparing actions at the
same state, so
\[
    D\leq\frac{2\gamma R_{\max}}{1-\gamma}.
\]
The temperature bound now yields
\[
    J_r(\pi_\tau)-J_r(\pi)
    \geq
    \frac{\tau}{1-\gamma}
    \left(
        \Delta_k(r)
        -\frac{\gamma R_{\max}B_k^2}
               {4(1-\gamma)^2}\tau
    \right).
\]
Substituting the expression for $\Delta_k(r)$ shows that this
quantity is strictly positive under the second stated
condition on $\tau$.
\end{proof}

\subsection{Estimation Error of Advantage Function Under Trust Region}

For the trust region algorithms, our estimation of advantage function can be bounded. Furthermore, we claim that trust-region-controlled SUAF reuse yields reward-uniform policy-gradient error bounds, with reliable ascent whenever the gradient signal exceeds the reuse error.

\begin{lemma}[Reward-independent transfer bound]
\label{thm:suaf_policy_transfer}
Denote $\Delta A_r^{\pi^\prime} = A_r^{\pi^\prime}-\widehat A_r$. Under the above setting, the approximation in
\eqref{eq:transferred_advantage} satisfies
\begin{equation*}
\begin{aligned}
    \sup_{\|r\|_{\nu^{+,\pi}}\leq 1}
    \left\|\Delta A_r^{\pi^\prime}\right\|_{d^{\pi^\prime}}^2
    &\leq
    \frac{\gamma L_{\max}^2}{2(1-\gamma)}
    \mathcal D_{\mathrm{KL}}(\pi'\|\pi),\\
    \sup_{\|r\|_{\nu^{+,\pi}}\leq 1}
    \left\|\Delta A_r^{\pi^\prime}\right\|_{d^{\pi}}^2
    &\leq
    \frac{(4-3\gamma) L_{\max}^2}{2(1-\gamma)}
    \mathcal D_{\mathrm{KL}}(\pi\|\pi^\prime)
\end{aligned}
\end{equation*}
\end{lemma}

\begin{proof}
Fix a reward $r$ with $\|r\|_{\nu^{+,\pi}}\leq 1$.
Write $d_S=d_S^{\pi'}$, and define
\[
    \Delta V_r(s):=V_r^{\pi'}(s)-V_r^\pi(s),
    \qquad
    q_r(s):=
    \mathbb E_{a\sim\pi'(\cdot\mid s)}[A_r^\pi(s,a)].
\]
Throughout the proof, $\|u\|_{d_S}^2=\sum_s d_S(s)u^2(s)$.
Let $P_S^{\pi'}$ denote the state transition operator:
\[
    (P_S^{\pi'}u)(s)
    =
    \sum_{a,s'}\pi'(a\mid s)T(s'\mid s,a)u(s').
\]

\paragraph{Step 1: Expressing the transfer error.}
The Bellman equations imply
\begin{equation}
\label{eq:transfer_value_difference}
    \Delta V_r=q_r+\gamma P_S^{\pi'}\Delta V_r,
\end{equation}
in that
$q_r(s)=\mathbb E_{a\sim\pi'}[Q_r^\pi(s,a)]-V_r^\pi(s)$.
Moreover, since
$\widehat A_r=C_{\pi'}Q_r^\pi$ and
$A_r^{\pi'}=C_{\pi'}Q_r^{\pi'}$, we have
\[
    A_r^{\pi'}-\widehat A_r
    =
    \gamma C_{\pi'}h_r,
    \qquad
    h_r(s,a):=
    \sum_{s'}T(s'\mid s,a)\Delta V_r(s').
\]
Therefore, using the conditional variance identity and Jensen's inequality,
\begin{align}
    \|A_r^{\pi'}-\widehat A_r\|_{d^{\pi'}}^2
    &=
    \gamma^2\left(
        \mathbb E_{(s,a)\sim d^{\pi'}}[h_r(s,a)^2]
        -
        \|P_S^{\pi'}\Delta V_r\|_{d_S}^2
    \right)
    \nonumber\\
    &\leq
    \gamma^2\left(
        \mathbb E_{s\sim d_S}
        [P_S^{\pi'}(\Delta V_r^2)(s)]
        -
        \|P_S^{\pi'}\Delta V_r\|_{d_S}^2
    \right).
\label{eq:transfer_variance_bound}
\end{align}

\paragraph{Step 2: Using the discounted occupancy identity.}
The state occupancy satisfies
\[
    d_S=(1-\gamma)\mu+\gamma d_S P_S^{\pi'}.
\]
Consequently,
\[
    \gamma\,
    \mathbb E_{s\sim d_S}
    [P_S^{\pi'}(\Delta V_r^2)(s)]
    =
    \|\Delta V_r\|_{d_S}^2
    -(1-\gamma)\|\Delta V_r\|_\mu^2
    \leq
    \|\Delta V_r\|_{d_S}^2.
\]
Combining this inequality with
\eqref{eq:transfer_variance_bound} and
$\gamma P_S^{\pi'}\Delta V_r=\Delta V_r-q_r$ yields
\begin{align}
    \|A_r^{\pi'}-\widehat A_r\|_{d^{\pi'}}^2
    &\leq
    \gamma\|\Delta V_r\|_{d_S}^2
    -
    \|\Delta V_r-q_r\|_{d_S}^2
    \nonumber\\
    &=
    -(1-\gamma)
    \left\|\Delta V_r-\frac{q_r}{1-\gamma}\right\|_{d_S}^2
    +
    \frac{\gamma}{1-\gamma}\|q_r\|_{d_S}^2
    \nonumber\\
    &\leq
    \frac{\gamma}{1-\gamma}\|q_r\|_{d_S}^2.
\label{eq:transfer_energy_bound}
\end{align}

\paragraph{Step 3: Bounding the policy perturbation by KL.}
Since $\mathbb E_{a\sim\pi(\cdot\mid s)}[A_r^\pi(s,a)]=0$,
\[
    q_r(s)
    =
    \sum_a
    \bigl(\pi'(a\mid s)-\pi(a\mid s)\bigr)A_r^\pi(s,a).
\]
For any two probability distributions, their difference in the
expectation of a function is bounded by the function's range times
their total variation distance. Hence, by the definition of $L_{\max}$,
\[
    |q_r(s)|
    \leq
    L_{\max}\,
    \operatorname{TV}\bigl(
        \pi'(\cdot\mid s),\pi(\cdot\mid s)
    \bigr),
\]
where $\operatorname{TV}(p,q)=\frac12\sum_a|p(a)-q(a)|$.
Pinsker's inequality \cite{pinsker1964information}, with natural logarithms, therefore gives
\[
    \|q_r\|_{d_S}^2
    \leq
    \frac{L_{\max}^2}{2}
    \mathbb E_{s\sim d_S^{\pi'}}
    \left[
        D_{\mathrm{KL}}\bigl(
            \pi'(\cdot\mid s)\|\pi(\cdot\mid s)
        \bigr)
    \right].
\]
Substituting into \eqref{eq:transfer_energy_bound} and taking the
supremum over $\|r\|_{\nu^{+,\pi}}\leq 1$ proves
\[
    \sup_{\|r\|_{\nu^{+,\pi}}\leq 1}
    \|A_r^{\pi'}-\widehat A_r\|_{d^{\pi'}}^2
    \leq
    \frac{\gamma L_{\max}^2}{2(1-\gamma)}
    \mathbb E_{s\sim d_S^{\pi'}}
    \left[
        D_{\mathrm{KL}}\bigl(
            \pi'(\cdot\mid s)\|\pi(\cdot\mid s)
        \bigr)
    \right].
\]
\paragraph{Step 4: Controlling the reference-centered error.}
We next establish the bound under $d^\pi$.
Define
\[
    \Delta Q_r:=Q_r^{\pi'}-Q_r^\pi=\gamma h_r,
    \qquad
    \widetilde q_r(s)
    :=
    \mathbb E_{a\sim\pi(\cdot\mid s)}
    [A_r^{\pi'}(s,a)].
\]
The Bellman equations also give
\[
    \Delta V_r
    =
    -\widetilde q_r+\gamma P_S^\pi\Delta V_r,
\]
where $P_S^\pi$ is the state transition operator under $\pi$.
Indeed,
\[
    \widetilde q_r
    =
    \mathbb E_{a\sim\pi}[Q_r^{\pi'}]
    -V_r^{\pi'}
    =
    \gamma P_S^\pi\Delta V_r-\Delta V_r.
\]

Applying the conditional variance identity and Jensen's inequality
under the reference policy yields
\begin{align*}
    \|C_\pi\Delta Q_r\|_{d^\pi}^2
    &=
    \gamma^2\left(
        \mathbb E_{d^\pi}[h_r^2]
        -
        \|P_S^\pi\Delta V_r\|_{d_S^\pi}^2
    \right)\leq
    \gamma^2\left(
        \mathbb E_{d_S^\pi}
        [P_S^\pi(\Delta V_r^2)]
        -
        \|P_S^\pi\Delta V_r\|_{d_S^\pi}^2
    \right).
\end{align*}
Using
$d_S^\pi=(1-\gamma)\mu+\gamma d_S^\pi P_S^\pi$
and
$\gamma P_S^\pi\Delta V_r=\Delta V_r+\widetilde q_r$,
we obtain
\begin{align}
    \|C_\pi\Delta Q_r\|_{d^\pi}^2
    &\leq
    \gamma\|\Delta V_r\|_{d_S^\pi}^2
    -
    \|\Delta V_r+\widetilde q_r\|_{d_S^\pi}^2
    \nonumber\\
    &=
    -(1-\gamma)
    \left\|
        \Delta V_r+\frac{\widetilde q_r}{1-\gamma}
    \right\|_{d_S^\pi}^2
    +
    \frac{\gamma}{1-\gamma}
    \|\widetilde q_r\|_{d_S^\pi}^2
    \nonumber\\
    &\leq
    \frac{\gamma}{1-\gamma}
    \|\widetilde q_r\|_{d_S^\pi}^2.
    \label{eq:reference_transfer_energy_bound}
\end{align}

Since $\mathbb E_{a\sim\pi'}[A_r^{\pi'}(s,a)]=0$,
\[
    \widetilde q_r(s)
    =
    \sum_a
    \bigl(\pi(a\mid s)-\pi'(a\mid s)\bigr)
    A_r^{\pi'}(s,a).
\]
The definition of $L_{\max}$ and Pinsker's inequality imply
\[
    \|\widetilde q_r\|_{d_S^\pi}^2
    \leq
    \frac{L_{\max}^2}{2}
    \mathbb E_{s\sim d_S^\pi}
    \left[
        D_{\mathrm{KL}}\bigl(
            \pi(\cdot\mid s)\|\pi'(\cdot\mid s)
        \bigr)
    \right].
\]
Consequently,
\begin{equation}
    \|C_\pi\Delta Q_r\|_{d^\pi}^2
    \leq
    \frac{\gamma L_{\max}^2}{2(1-\gamma)}
    \mathcal D_{\mathrm{KL}}(\pi\|\pi').
    \label{eq:reference_centered_transfer_bound}
\end{equation}

\paragraph{Step 5: Accounting for the change of centering.}
The actual transfer error is $C_{\pi'}\Delta Q_r$.
Define the state-dependent discrepancy
\[
    m_r(s)
    :=
    \mathbb E_{a\sim\pi(\cdot\mid s)}[\Delta Q_r(s,a)]
    -
    \mathbb E_{a\sim\pi'(\cdot\mid s)}[\Delta Q_r(s,a)].
\]
Then
\[
    A_r^{\pi'}-\widehat A_r
    =
    C_\pi\Delta Q_r+m_r.
\]
Since $C_\pi\Delta Q_r$ has zero conditional action mean
under $\pi$, it is orthogonal to $m_r$ in $L^2(d^\pi)$.
Therefore,
\begin{equation}
    \|A_r^{\pi'}-\widehat A_r\|_{d^\pi}^2
    =
    \|C_\pi\Delta Q_r\|_{d^\pi}^2
    +
    \|m_r\|_{d_S^\pi}^2.
    \label{eq:reference_transfer_orthogonal_decomposition}
\end{equation}

Subtracting a state-dependent baseline does not change
the action span. Hence, for every $s,a,b$,
\begin{align*}
    |\Delta Q_r(s,a)-\Delta Q_r(s,b)|
    &\leq
    |A_r^{\pi'}(s,a)-A_r^{\pi'}(s,b)|+
    |A_r^\pi(s,a)-A_r^\pi(s,b)|\\
    &\leq 2L_{\max}.
\end{align*}
It follows that
\[
    |m_r(s)|
    \leq
    2L_{\max}\,
    \operatorname{TV}\bigl(
        \pi(\cdot\mid s),\pi'(\cdot\mid s)
    \bigr).
\]
Applying Pinsker's inequality gives
\[
    \|m_r\|_{d_S^\pi}^2
    \leq
    2L_{\max}^2\,
    \mathcal D_{\mathrm{KL}}(\pi\|\pi').
\]
Combining this inequality with
\eqref{eq:reference_centered_transfer_bound} and
\eqref{eq:reference_transfer_orthogonal_decomposition}
yields
\begin{align*}
    \|A_r^{\pi'}-\widehat A_r\|_{d^\pi}^2
    &\leq
    \left(
        \frac{\gamma}{2(1-\gamma)}+2
    \right)
    L_{\max}^2\,
    \mathcal D_{\mathrm{KL}}(\pi\|\pi')\\
    &=
    \frac{(4-3\gamma)L_{\max}^2}{2(1-\gamma)}
    \mathcal D_{\mathrm{KL}}(\pi\|\pi').
\end{align*}
Finally, taking the supremum over the same reward class gives
\[
    \sup_{\|r\|_{\nu^{+,\pi}}\leq1}
    \|A_r^{\pi'}-\widehat A_r\|_{d^\pi}^2
    \leq
    \frac{(4-3\gamma)L_{\max}^2}{2(1-\gamma)}
    \mathcal D_{\mathrm{KL}}(\pi\|\pi').
\]
\end{proof}

\subsection{Reward-independent policy gradient approximation error bound}

In this section, we show that SUAF can be reused within the specified trust region, with reward-uniform bounds on advantage and policy-gradient approximation errors.

\begin{lemma}[Reward-independent policy gradient approximation error bound]
\label{thm:reward_independent_pg}
Given the reference policy $\pi_{\theta}$ in the trust region of $\pi_{\text{ref}}$, i.e.,
$
\gD(\pi_\theta \| \pi_{\text{ref}}) < \delta,
$
the approximation error of using SUAF of $\pi_{\text{ref}}$ for policy gradient satisfies that
\begin{equation*}
    \sup_{\| r \|_{\nu^{+,{\pi_\text{ref}}}}\leq 1} \left\| \nabla_\theta J_r(\pi_\theta) - \widehat G_r \right\|_{F_\theta^{\dagger}}^2 < \frac{\gamma L_{\max}^2}{2(1-\gamma)} \delta,
\end{equation*}
where $\widehat{G}_r = \E_{(s,a)\sim d^{\pi_\theta}}[\nabla_\theta \log \pi_\theta \widehat A_r]$ with $\widehat A_r = C_{\pi_\theta}A_r^{\pi_{\text{ref}}}$ is the approximated policy gradient and $F_\theta = \E_{d^{\pi_\theta}}[\nabla_\theta\log\pi_\theta(\nabla_\theta\log\pi_\theta)^\text{T}]$ denotes the Fisher matrix according to policy $\pi_\theta$ with $\dagger$ denoting pseudo-inverse. 
\end{lemma}
\begin{proof}
Fix a reward $r$ with $\|r\|_\nu\le1$, and set
$$
    e_r
    :=
    A_r^{\pi_\theta}-\widehat A_r,
    G_r(\theta) = \nabla_\theta J_r(\pi_\theta),
    b_r
    :=
    G_r(\theta)
    -
    \widehat G_r(\theta)
    =
    \mathbb E_{d^{\pi_\theta}}[\nabla_\theta\log \pi_\theta e_r].
$$
Write
\begin{equation}
    \varepsilon_\delta
    :=
    \left(
        \frac{\gamma L_{\max}^2}{2(1-\gamma)}
        \delta
    \right)^{1/2}.
\end{equation}
By Theorem \ref{thm:suaf_policy_transfer}, the advantage transfer bound yields
\begin{equation}
    \|e_r\|_{d^{\pi_\theta}}^2
    \le
    \frac{\gamma L_{\max}^2}{2(1-\gamma)}
    \gD_{\mathrm{KL}}(\pi_\theta\|\pi_{\mathrm{ref}})
    <
    \varepsilon_\delta^2.
\end{equation}
For any vector $u$, Cauchy--Schwarz gives
\begin{equation}
    |u^\top b_r|
    \le
    \left(
        \mathbb E_{d^{\pi_\theta}}
        [(u^\top\nabla_\theta\log \pi_\theta)^2]
    \right)^{1/2}
    \|e_r\|_{d^{\pi_\theta}}
    =
    \sqrt{u^\top F_\theta u}\,
    \|e_r\|_{d^{\pi_\theta}}.
\end{equation}
Taking $u=F_\theta^\dagger b_r$ and using
$F_\theta^\dagger F_\theta F_\theta^\dagger
=F_\theta^\dagger$, we obtain $\|b_r\|_{F_\theta^\dagger}^2 \le \|b_r\|_{F_\theta^\dagger}\|e_r\|_{d^{\pi_\theta}}$, which further gives
\[
    \|b_r\|_{F_\theta^\dagger}^2
    \le
    \|e_r\|_{d^{\pi_\theta}}^2 < \varepsilon_\delta^2.
\]
Note that the above inequality holds for all reward function. Take the supremum and we prove the theorem.
\end{proof}

\subsection{Proof of Corollary \ref{thm:reward_independent_pg}}
\label{app:proof_transferred_improvement}
\begin{corollary}[Policy improvement using transferred SUAF]
Under the conditions of Theorem~\ref{thm:reward_independent_pg}, fix a reward $r$ with $\|r\|_\nu\le1$ and define
$
    u_r
    :=
    F_\theta^\dagger\widehat G_r(\theta),
    s_r
    :=
    \|\widehat G_r(\theta)\|_{F_\theta^\dagger},
    \varepsilon_\delta
    :=
    \left(
        \frac{\gamma L_{\max}^2}{2(1-\gamma)}
        \delta
    \right)^{1/2}
$
Then
$
    \nabla_\theta J_r(\pi_\theta)^\top u_r
    \geq
    s_r(s_r-\varepsilon_\delta).
$
In particular, if $s_r>\varepsilon_\delta$, then $u_r$ is a strict ascent direction, and there exists $\eta_0>0$ such that
\begin{equation}
    J_r(\pi_{\theta+\eta u_r})
    >
    J_r(\pi_\theta),
    \qquad
    0<\eta<\eta_0.
\end{equation}
Moreover, if $\theta\mapsto J_r(\pi_\theta)$ has an $L$-Lipschitz gradient on a convex neighborhood containing the update segment, strict improvement holds whenever
\begin{equation}
    0<\eta<
    \frac{2s_r(s_r-\varepsilon_\delta)}
         {L\|u_r\|_2^2}.
\end{equation}
\end{corollary}
\begin{proof}
Fix a reward $r$ with $\|r\|_\nu\le1$, by Theorem \ref{thm:reward_independent_pg}, we have that 
$$
\|G_r-\widehat G_r\|_{F_\theta^\dagger} \leq \varepsilon_\delta.
$$
Therefore, Cauchy--Schwarz for the positive semi-definite
form induced by $F_\theta^\dagger$ gives
\begin{align}
    G_r^\top u_r
    &=
    \widehat G_r^\top F_\theta^\dagger\widehat G_r
    +
    (G_r-\widehat G_r)^\top
    F_\theta^\dagger\widehat G_r
    \nonumber\\
    &\ge
    s_r^2
    -
    \|G_r-\widehat G_r\|_{F_\theta^\dagger}s_r
    \nonumber\\
    &\ge
    s_r(s_r-\varepsilon_\delta).
\end{align}
When $s_r>\varepsilon_\delta$, this directional
derivative is strictly positive. Differentiability implies
\[
    J_r(\pi_{\theta+\eta u_r})
    -
    J_r(\pi_\theta)
    =
    \eta G_r^\top u_r+o(\eta)\geq0
\]
for all sufficiently small $\eta>0$.
Under the additional smoothness assumption,
\begin{align}
    J_r(\pi_{\theta+\eta u_r})
    -
    J_r(\pi_\theta)
    &\ge
    \eta G_r^\top u_r
    -
    \frac{L\eta^2}{2}\|u_r\|_2^2
    \nonumber\\
    &\ge
    \eta s_r(s_r-\varepsilon_\delta)
    -
    \frac{L\eta^2}{2}\|u_r\|_2^2.
\end{align}
The stated step-size condition makes this lower bound strictly positive.
\end{proof}

\subsection{Proof of Corollary \ref{cor:truncated_suaf_pg}}
\label{app:proof_truncated_suaf_pg}

\begin{corollary}[Policy-gradient error with truncated SUAF]
Let $\pi=\pi_{\mathrm{ref}}$ and $\pi'=\pi_\theta$.
Suppose $\mathcal D_{\mathrm{KL}}(\pi'\|\pi)<\delta.$
Define the transferred rank-$k$ approximation by
\[
\widehat A_r^k
=
C_{\pi'}r+
\frac{\gamma}{1-\gamma}
\sum_{i=1}^k
\sigma_i^{+,\pi}(C_{\pi'}f_i^\pi)
\langle g_i^\pi,r\rangle_{\nu^{+,\pi}},
\]
and let
$
\widehat G_\theta^k(r)
=
\mathbb E_{d^{\pi'}}
[
\nabla_\theta\log\pi_\theta(a\mid s)\,
\widehat A_r^k(s,a)
].
$
For
$\eta_\delta=\min\{1,\sqrt{\delta/[2(1-\gamma)]}\}$, $B(\delta)=\min\{R^2_{\max}, 1+ R^2_{\max}C_\gamma\sqrt{\delta}\}$ with $C_\gamma = \sqrt{(4+\gamma^{-1})/(2(1-\gamma)^2)}$,
\begin{align*}
&\sup_{\|r\|_{\nu^{+,\pi^\prime}}\le1}
\left\|
\nabla_\theta\bar J_r(\pi_\theta)
-\widehat G_\theta^k(r)
\right\|_{F_\theta^\dagger}^2\\
&\quad\quad\le
B(\delta)\left[
\sqrt{\frac{\gamma\delta L_{\max}^2}{2(1-\gamma)}}
+
\sqrt{
\left(\frac{\gamma}{1-\gamma}\right)^2
\left(\sigma_{k+1}^{+,\pi}\right)^2
+
L_{\max}^2\eta_\delta
}
\right]^2.
\end{align*}
\end{corollary}

\begin{proof}
Write
\[
c=\frac{\gamma}{1-\gamma},
\qquad
\nu=\nu^{+,\pi},
\qquad
\nu'=\nu^{+,\pi'},
\]
and let
\[
\mathcal D
=
\mathbb E_{s\sim d_S^{\pi'}}
D_{\mathrm{KL}}\bigl(
\pi'(\cdot\mid s)\|\pi(\cdot\mid s)
\bigr)
<\delta.
\]

\paragraph{Step 1: Controlling the changes in sampling distributions.}
Let $T_0$ be independent of the trajectory with
$\Pr(T_0=t)=(1-\gamma)\gamma^t$ for $t\ge0$.
The distribution of $(S_{T_0},A_{T_0})$ under a policy
$\rho$ is $d^\rho$.
By the chain rule, the KL divergence between the laws
of the trajectories stopped at $T_0$, under $\pi'$ and
$\pi$, equals
\[
\sum_{t=0}^{\infty}\gamma^t
\mathbb E_{\pi'}
\left[
D_{\mathrm{KL}}\bigl(
\pi'(\cdot\mid S_t)\|\pi(\cdot\mid S_t)
\bigr)
\right]
=
\frac{\mathcal D}{1-\gamma}.
\]
Pinsker's inequality therefore yield
\begin{equation}
\operatorname{TV}(d^{\pi'},d^\pi)
\le
\min\left\{
1,\sqrt{\frac{\mathcal D}{2(1-\gamma)}}
\right\}
\le\eta_\delta.
\label{eq:truncated_pg_occupancy_tv}
\end{equation}

To control $\nu'$ and $\nu$, independently draw
$L\ge1$ with
$\Pr(L=\ell)=(1-\gamma)\gamma^{\ell-1}$,
and set $N=T_0+L$.
The distribution of $(S_N,A_N)$ under $\rho$ is
$\nu^{+,\rho}=d^\rho M^{+,\rho}$.
Define
\[
\epsilon_t
=
\mathbb E_{\pi'}
\left[
\operatorname{TV}\bigl(
\pi'(\cdot\mid S_t),\pi(\cdot\mid S_t)
\bigr)
\right].
\]
Jensen's and Pinsker's inequalities imply
\[
\sum_{t=0}^{\infty}(1-\gamma)\gamma^t\epsilon_t^2
\le\frac{\mathcal D}{2}.
\]

For each fixed horizon, replacing the action kernels
one at a time bounds the total variation of the terminal
distribution by the sum of the expected per-step action
total variations under $\pi'$.
Averaging over $N$ consequently gives
\[
\operatorname{TV}(\nu',\nu)
\le
\sum_{t=0}^{\infty}a_t\epsilon_t,
\qquad
a_t:=\Pr(N\ge t).
\]
Here
\[
a_0=1,
\qquad
a_t=\gamma^t+t(1-\gamma)\gamma^{t-1},
\quad t\ge1.
\]
Weighted Cauchy--Schwarz yields
\begin{align*}
\operatorname{TV}(\nu',\nu)^2
&\le
\left(
\sum_{t=0}^{\infty}
\frac{a_t^2}{(1-\gamma)\gamma^t}
\right)
\left(
\sum_{t=0}^{\infty}
(1-\gamma)\gamma^t\epsilon_t^2
\right)\le
\frac{4+\gamma^{-1}}{(1-\gamma)^2}
\frac{\mathcal D}{2}
\le C_\gamma^2\delta.
\end{align*}
Consequently, for every reward satisfying
$\|r\|_{\nu'}\le1$ and $\|r\|_\infty\le R_{\max}$,
\[
\|r\|_\nu^2
\le
\|r\|_{\nu'}^2+
R_{\max}^2\operatorname{TV}(\nu,\nu')
\le
1+R_{\max}^2C_\gamma\sqrt{\delta}.
\]
Since also $\|r\|_\nu^2\le R_{\max}^2$, we obtain
\begin{equation}
\|r\|_\nu^2\le B(\delta).
\label{eq:truncated_pg_reward_norm}
\end{equation}

\paragraph{Step 2: Bounding the spectral residual on the reference reward ball.}
Temporarily fix a reward $u$ with $\|u\|_\nu\le1$.
Define its omitted reward component by
\[
u_{>k}
=
\sum_{i=k+1}^{R}
\langle g_i^\pi,u\rangle_\nu g_i^\pi.
\]
Orthonormality gives $\|u_{>k}\|_\nu\le1$.

The state-action transition operator satisfies
$P_\pi C_\pi=0$, and hence
$\gT^{+,\pi}C_\pi=0$.
Moreover,
$\nu(s,a)=\nu_S(s)\pi(a\mid s)$,
so $C_\pi$ is an orthogonal projection in $L^2(\nu)$.
Every right singular function corresponding to a
nonzero singular value therefore satisfies
\[
C_\pi g_i^\pi=0.
\]
Indeed, for every $v\in L^2(\nu)$,
\[
\langle g_i^\pi,C_\pi v\rangle_\nu
=
\frac{1}{\sigma_i^{+,\pi}}
\langle f_i^\pi,\gT^{+,\pi}C_\pi v\rangle_{d^\pi}
=0.
\]

It follows that $C_\pi u_{>k}=0$ and
\begin{equation}
e_{k,u}
:=
A_u^\pi-A_{u,k}^\pi
=
c\sum_{i=k+1}^{R}
\sigma_i^{+,\pi}f_i^\pi
\langle g_i^\pi,u\rangle_\nu
=
A_{u_{>k}}^\pi.
\label{eq:truncated_pg_residual_advantage}
\end{equation}
Because $\|u_{>k}\|_\nu\le1$, the definition of
$L_{\max}$ bounds the action span of $e_{k,u}$ by
$L_{\max}$.
Its conditional action mean under $\pi$ is zero.
Therefore
\[
|e_{k,u}(s,a)|\le L_{\max}.
\]
The spectral approximation bound also gives
\[
\|e_{k,u}\|_{d^\pi}^2
\le
c^2\bigl(\sigma_{k+1}^{+,\pi}\bigr)^2.
\]
Using \eqref{eq:truncated_pg_occupancy_tv}, we conclude
\begin{align}
\|e_{k,u}\|_{d^{\pi'}}^2
&\le
\|e_{k,u}\|_{d^\pi}^2
+
L_{\max}^2\operatorname{TV}(d^{\pi'},d^\pi)
\nonumber\\
&\le
c^2\bigl(\sigma_{k+1}^{+,\pi}\bigr)^2
+
L_{\max}^2\eta_\delta.
\label{eq:truncated_pg_residual_bound}
\end{align}

\paragraph{Step 3: Combining transfer and truncation errors.}
Let $\widehat A_u^\infty=C_{\pi'}Q_u^\pi$ denote
the transferred full-rank representation.
Since $C_{\pi'}$ annihilates state-dependent functions,
the transferred truncated representation satisfies
$\widehat A_u^k=C_{\pi'}A_{u,k}^\pi$.
Thus
\[
\widehat A_u^\infty-\widehat A_u^k
=
C_{\pi'}e_{k,u}.
\]
The operator $C_{\pi'}$ is an orthogonal projection in
$L^2(d^{\pi'})$, so
\[
\|\widehat A_u^\infty-\widehat A_u^k\|_{d^{\pi'}}
\le
\|e_{k,u}\|_{d^{\pi'}}.
\]
Combining \eqref{eq:truncated_pg_residual_bound}
with the full-rank transfer theorem and the triangle
inequality gives
\begin{align*}
\|A_u^{\pi'}-\widehat A_u^k\|_{d^{\pi'}}
&\le
\|A_u^{\pi'}-\widehat A_u^\infty\|_{d^{\pi'}}
+
\|\widehat A_u^\infty-\widehat A_u^k\|_{d^{\pi'}}\\
&\le
L_{\max}\sqrt{\frac{\gamma\delta}{2(1-\gamma)}}
+
\sqrt{
c^2\bigl(\sigma_{k+1}^{+,\pi}\bigr)^2
+
L_{\max}^2\eta_\delta
}.
\end{align*}

For $\bar J_u=(1-\gamma)J_u$, the policy-gradient
identity and Cauchy--Schwarz imply
\[
\left\|
\nabla_\theta\bar J_u(\pi_\theta)
-\widehat G_\theta^k(u)
\right\|_{F_\theta^\dagger}
\le
\|A_u^{\pi'}-\widehat A_u^k\|_{d^{\pi'}}.
\]
Indeed, writing
$s_\theta=\nabla_\theta\log\pi_\theta(a\mid s)$,
the gradient error is
$\mathbb E_{d^{\pi'}}[s_\theta
(A_u^{\pi'}-\widehat A_u^k)]$,
and $F_\theta=\mathbb E_{d^{\pi'}}[s_\theta s_\theta^\top]$.

\paragraph{Step 4: Returning to the target reward class.}
Now let $\|r\|_{\nu'}\le1$ and
$\|r\|_\infty\le R_{\max}$.
If $t:=\|r\|_\nu>0$, set $u=r/t$.
Then $\|u\|_\nu=1$.
Both the exact policy gradient and its SUAF estimate
are linear in the reward, so their error for $r$
equals $t$ times their error for $u$.
Using \eqref{eq:truncated_pg_reward_norm}, we obtain
\begin{align*}
&
\left\|
\nabla_\theta\bar J_r(\pi_\theta)
-\widehat G_\theta^k(r)
\right\|_{F_\theta^\dagger}^2\le
B(\delta)
\left[
L_{\max}\sqrt{\frac{\gamma\delta}{2(1-\gamma)}}
+
\sqrt{
c^2\bigl(\sigma_{k+1}^{+,\pi}\bigr)^2
+
L_{\max}^2\eta_\delta
}
\right]^2.
\end{align*}
The case $t=0$ is immediate under the full-support
assumption.
Taking the supremum over
$\{r:\|r\|_{\nu'}\le1,\ \|r\|_\infty\le R_{\max}\}$
completes the proof.
\end{proof}

\section{Algorithms}

\subsection{Algorithm to Estimate SUAF}
\label{sec:leanring_suaf}

We provide the derivations omitted from the main text.

\paragraph{Reverse conditional and marginal distributions.}
For one-step transitions $(s,a)\sim d^\pi$ and $(s^\prime,a^\prime)\sim P_\pi(\cdot|s,a)=\pi(a^\prime|s^\prime)T(s^\prime|s,a)$, the reverse conditional is
\[
p_\pi(s,a|s^\prime,a^\prime) = \frac{d^\pi(s,a)P_\pi(s^\prime,a^\prime|s,a)}{\nu_1^\pi(s^\prime,a^\prime)},
\]
with marginal
\[
\nu_1^\pi(\cdot,\cdot):=\sum_{s,a} d^\pi(s,a) P_\pi(\cdot,\cdot|s,a).
\]
All density ratios reduce to one when $d^\pi$ is stationary.

\paragraph{Adjoint interpretation.}
Denote by $(\gT^{+,\pi})^*$ the adjoint of $\gT^{+,\pi}$. Then $b=(\gT^{+,\pi})^*f_\theta$ and $h-\E_\pi[h|s]=\gT^{+,\pi}g_\omega$, which yields the two equivalent expressions in \eqref{eq:suaf_bidirectional}.

\paragraph{Backward prediction with marginal correction.}
When the sampling distribution is not stationary, we define
\[
w_1(s^\prime,a^\prime)=\frac{\nu_1^\pi(s^\prime,a^\prime)}{d^\pi(s^\prime,a^\prime)},
\qquad
w_+(s^\prime,a^\prime)=\frac{d^\pi(s^\prime,a^\prime)}{\nu^{+,\pi}(s^\prime,a^\prime)}.
\]
Writing $b=w_+v$, the backward recursion becomes
\[
v(s^\prime, a^\prime)
=
w_1(s^\prime, a^\prime)\,
\mathbb E\!\left[
(1-\gamma)f_\theta(s,a)+\gamma v(s,a)
\mid s^\prime, a^\prime
\right],
\]
and $\widehat b_\theta=w_+v_\eta$.

\paragraph{Surrogate backward loss.}
Below, we treat \(\sg(\cdot)\) as a stop-gradient constant. There are only two key identities:
\[
\E_{\nu}\left[\frac{d^\pi}{\nu}h\right]=\E_{d^\pi}[h],
\qquad
\E_{\nu_B}\left[w_1 h\right]=\E_{P}h,
\]
where \(P\) is the joint distribution after one step starting from \(d^\pi\).

Let
\[
T(s,a)=(1-\gamma)f_\theta(s,a)+\gamma v_{\bar\eta}(s,a).
\]
Taking one step from \(d^\pi\) yields
\[
P(s',a')=\sum_{s,a}d^\pi(s,a)P_\pi(s',a'\mid s,a).
\]
Define
\[
\nu_B(s',a')=\gamma P(s',a')+(1-\gamma)d_0(s',a'),
\qquad
w_1(s',a')=\frac{P(s',a')}{\nu_B(s',a')}.
\]
This is exactly the density ratio. The original backward loss with density ratio can be written in weighted least-squares form
\[
\gL_{\mathrm B}(\eta)
=
\frac12\E_{(s',a')\sim \nu_B}
\left[
\left\|v_\eta(s',a')-w_1(s',a')
\sg\!\left(\E_{P_\pi}[T(s,a)\mid s',a']\right)
\right\|^2
\right]
+\text{const}.
\]
where the constant is independent of \(v_\eta\). Expanding the square:
\[
\gL_{\mathrm B}
=
\frac12\E_{\nu_B}\|v_\eta\|^2
-
\E_{\nu_B}\left[v_\eta^\top w_1
\sg\!\left(\E_{P_\pi}[T\mid s',a']\right)\right]
+\text{const}.
\]
Using \(w_1\nu_B=P\), the second term becomes
\[
\E_{\nu_B}\left[v_\eta^\top w_1
\sg\!\left(\E_{P_\pi}[T\mid s',a']\right)\right]
=
\E_{(s,a)\sim d^\pi,(s',a')\sim P_\pi}
\left[v_\eta(s',a')^\top \sg(T(s,a))\right].
\]
The first term splits as:
\[
\E_{\nu_B}\|v_\eta\|^2
=
\gamma \E_{P}\|v_\eta\|^2
+
(1-\gamma)\E_{d_0}\|v_\eta\|^2.
\]
Therefore
\[
\gL_{\mathrm B}
=
\frac{\gamma}{2}\E_{P}\|v_\eta\|^2
-
\E_{P}\left[v_\eta^\top \sg(T)\right]
+
\frac{1-\gamma}{2}\E_{d_0}\|v_\eta\|^2
+\text{const}.
\]
Now complete the square:
\[
\frac{\gamma}{2}\E_{P}
\left\|v_\eta-\frac{\sg(T)}{\gamma}\right\|^2
=
\frac{\gamma}{2}\E_{P}\|v_\eta\|^2
-
\E_{P}\left[v_\eta^\top \sg(T)\right]
+
\frac{1}{2\gamma}\E_{P}\|\sg(T)\|^2.
\]
The last term is independent of \(v_\eta\), hence
\begin{align*}
\gL_{\mathrm B}
& =
\frac{\gamma}{2}\E_{(s,a)\sim d^\pi,(s',a')\sim P_\pi}
\left\|
v_\eta(s',a')
-
\frac{\sg\bigl((1-\gamma)f_\theta(s,a)+\gamma v_{\bar\eta}(s,a)\bigr)}{\gamma}
\right\|^2\\
&\qquad +
\frac{1-\gamma}{2}\E_{d_0}\|v_\eta\|^2
+\text{const}.
\end{align*}
This is exactly \(\widehat{\gL}_{\mathrm B}(\eta)\), up to a constant independent of \(v_\eta\).

Taking the stationary point with respect to \(v_\eta(s',a')\):
\[
\gamma P(s',a')\left(v_\eta-\frac{\E[T\mid s',a']}{\gamma}\right)
+(1-\gamma)d_0(s',a')v_\eta=0,
\]
i.e.,
\[
\bigl(\gamma P+(1-\gamma)d_0\bigr)v_\eta
=
P\,\E[T\mid s',a'].
\]
Thus
\[
v_\eta(s',a')
=
\frac{P(s',a')}{\gamma P(s',a')+(1-\gamma)d_0(s',a')}
\E[T\mid s',a']
=
w_1(s',a')\E[T\mid s',a'].
\]
This is exactly the same fixed point as the original loss with density ratio. When \(d^\pi\) is stationary and \(d_0=d^\pi\), \(P=d^\pi\), \(\nu_B=d^\pi\), so \(w_1\equiv 1\).

\paragraph{Surrogate \(g\)-loss.}
Let \(\nu=\nu^{+,\pi}\). The original \(g\)-loss with density ratio can be written as
\[
\gL_g(\omega)
=
\frac12\E_{\nu}
\left[
g_\omega^\top \sg[C_f]g_\omega
\right]
-
\E_{\nu}
\left[
\frac{d^\pi}{\nu}
g_\omega^\top \sg(v_\eta)
\right]
+\text{const}.
\]
The second term directly uses the density ratio identity:
\[
\E_{\nu}
\left[
\frac{d^\pi}{\nu}
g_\omega^\top \sg(v_\eta)
\right]
=
\sum_{s,a} \nu(s,a)\frac{d^\pi(s,a)}{\nu(s,a)}
g_\omega^\top \sg(v_\eta(s,a)) 
=
\E_{d^\pi}
\left[
g_\omega^\top \sg(v_\eta)
\right].
\]
Therefore
\[
\gL_g(\omega)
=
\frac12\E_{\nu^{+,\pi}}
\left[
g_\omega^\top \sg[C_f]g_\omega
\right]
-
\E_{d^\pi}
\left[
g_\omega^\top \sg(v_\eta)
\right]
+\text{const}.
\]
This is exactly
\[
\widehat{\gL}_g(\omega)
=
-
\E_{(s,a)\sim d^\pi}
\left[
g_\omega^\top \sg(v_\eta(s,a))
\right]
+
\frac12
\E_{\nu^{+,\pi}}
\left[
g_\omega^\top \sg[C_f]g_\omega
\right].
\]
up to a constant independent of \(g_\omega\).

Equivalently, if the original \(g\)-loss is written in weighted least-squares form
\[
\gL_g(\omega)
=
\frac12\E_{\nu}
\left\|
g_\omega-\frac{d^\pi}{\nu}\sg(C_f^{-1}v_\eta)
\right\|_{\sg[C_f]}^2
+\text{const},
\]
expanding also gives the same result: the quadratic term remains under \(\nu\), the density ratio in the cross term is cancelled by \(\nu\), becoming an expectation under \(d^\pi\).

Differentiating with respect to \(g_\omega\), treating \(\sg[C_f]\) as a constant matrix, yields
\[
\E_{\nu^{+,\pi}}[\sg[C_f]g_\omega]
=
\E_{d^\pi}[\sg(v_\eta)].
\]
This is the same stationary condition as the original loss with density ratio.



\paragraph{Main body of algorithm} For simplicity, we denote $X=(S,A)$ and $Y_1 = (S^\prime,A^\prime)$. Based on the new notations, we describe the algorithm of estimating SUAF with neural networks and bidirectional TD as follows:

\begin{algorithm}[H]
\caption{Neural Bidirectional TD Soft-HGR}
\label{alg:neural-bidirectional-td-soft-hgr}
\small
\begin{algorithmic}[1]
\Require Fixed reference policy $\pi$; rank $k$; $0<\gamma<1$; outer iterations $M$;
         first-step replay $\gD_{d_0}$, on-policy trajectory replay $\Dtr$ and independent calibration/diagnostic replay $\Dval$;
         TD steps $H_F,H_B$, feature steps $H_f,H_g$, learning rates, and target rates $\tau_F,\tau_B$.
\Statex \textbf{Optimized networks:} feature parameters $\theta,\omega$;
         forward predictor $h_\xi:\cX\to\mathbb R^k$ and backward auxiliary predictor
         $v_\eta:\cX\to\mathbb R^k$; Polyak targets $\bar\xi,\bar\eta$.
\Statex \textbf{Derived objects:} centered features $f_\theta,g_\omega$,
         sample moments, and final $k\times k$ whitening/SVD transforms; no feature tables.

\State Prepare one-step pairs $(X,Y_1)$ and positive-lag pairs $(X,Y^+)$ using the sampling rule below.
\State Initialize nonzero feature networks and predictors; set $\bar\xi\gets\xi$, $\bar\eta\gets\eta$.

\For{$m=1,\ldots,M$}
    \State Freeze $\omega$ and its centering statistics.
    \For{$h=1,\ldots,H_F$}
        \State Sample a minibatch $(X,Y_1)$ from the one-step replay.
        \State $Z_F\gets\sg[(1-\gamma)g_\omega(Y_1)+\gamma h_{\bar\xi}(Y_1)]$.
        \State $\xi\gets\xi-\alpha_F\nabla_\xi\Emp\|h_\xi(X)-Z_F\|_2^2$.
        \State $\bar\xi\gets(1-\tau_F)\bar\xi+\tau_F\xi$.
    \EndFor
    \State Estimate $\widehat C_g\gets\Emp_{Y^+}[g_\omega(Y^+)g_\omega(Y^+)^\text{T}]$.
    \For{$j=1,\ldots,H_f$}
        \State Sample $X\sim d^\pi$; evaluate the action-centered feature $f_\theta(X)$.
        \State $\displaystyle \mathcal L_f\gets\Emp_X\left[-f_\theta(X)^\text{T}\sg[h_\xi(X)] +\tfrac12 f_\theta(X)^\text{T}\sg[\widehat C_g]f_\theta(X)\right]$.
        \State $\theta\gets\theta-\alpha_f\nabla_\theta\mathcal L_f$.
               \Comment{Only the left feature network is updated}
    \EndFor

    \State Freeze the updated $\theta$.
    \For{$h=1,\ldots,H_B$}
        \State Sample a minibatch $(X,Y_1)$ from the same one-step replay.
        \State $Z_B \gets \operatorname{sg}\bigl((1\!-\!\gamma)f_\theta(X)
\!+\!\gamma v_{\bar\eta}(X\bigr))$
        \State $\widehat{\gL}_{\mathrm B}(\eta) \gets \frac{\gamma}{2}\mathbb E \left\|v_\eta(Y_1)\!-\!\frac{Z_B}{\gamma}\right\|^2 +\frac{1-\gamma}{2}\mathbb E_{\gD_{d_0}}\|v_\eta\|^2$
        \State $\eta\gets\eta-\alpha_B\nabla_\eta\widehat{\gL}_{\mathrm B}$.
        \State $\bar\eta\gets(1-\tau_B)\bar\eta+\tau_B\eta$.
    \EndFor
    \State Estimate $\widehat C_f\gets\Emp_X[f_\theta(X)f_\theta(X)^\text{T}]$.
    \For{$j=1,\ldots,H_g$}
        \State Sample $Y^+\sim\nu^{+,\pi}$; evaluate the marginal-centered feature $g_\omega(Y^+)$.
        \State $\displaystyle \widehat{\gL}_g\gets-\Emp_X\left[g_\omega^\text{T}\operatorname{sg}(v_\eta(X))\right] + \frac12 \Emp_{Y^+}\left[g_\omega^\text{T}\operatorname{sg}[C_f]g_\omega\right]$.
        \State $\omega\gets\omega-\alpha_g\nabla_\omega\widehat{\gL}_g$.
               \Comment{Only the right feature network is updated}
    \EndFor
\EndFor
\State Freeze all networks; estimate calibration moments using $\Dval$.
\State Obtain $(\widehat f,\widehat g,\widehat\Sigma)$ by the small-matrix canonicalization below.
\State \Return Canonical feature functions $\widehat f,\widehat g$, estimated singular values
       $\widehat\Sigma$, and predictors $h_\xi,v_\eta$.
\end{algorithmic}
\end{algorithm}



\paragraph{Sampling distributions.}
Write $X_t=(S_t,A_t)$ and
$P_\pi((s',a')\mid(s,a))=T(s'\mid s,a)\pi(a'\mid s')$.
All trajectories start from $S_0\sim\mu$ and follow the fixed policy $\pi$.
Assume the conditional expectations and second moments below are finite.
Draw independent times $T_0$ and $L$ with
\[
 \Pr(T_0=t)=(1-\gamma)\gamma^t\quad(t\geq0),\qquad
 \Pr(L=\ell)=(1-\gamma)\gamma^{\ell-1}\quad(\ell\geq1).
\]
Set $X=X_{T_0}$, $Y_1=X_{T_0+1}$, and $Y^+=X_{T_0+L}$. Then
\[
 X\sim d^\pi,\quad Y_1\sim\nu_1^\pi=d^\pi P_\pi,\quad
 Y^+\mid X\sim M^{+,\pi}(\cdot\mid X),\quad
 Y^+\sim\nu^{+,\pi}=d^\pi M^{+,\pi}.
\]
Replay sampling must implement these weights. Uniformly sampling all stored transitions does not generally do so. Finite trajectories require an explicit tail treatment. Rejecting unavailable lags or clipping them changes the target distribution. For episodic tasks, an absorbing-state convention must agree with the MDP definition. TD uses one-step pairs for the predictors. Positive-lag samples are still used for right-marginal moments, density-ratio estimation, and calibration.

\paragraph{Centering and the soft-HGR objective.}
Let $\widetilde f_\theta,\widetilde g_\omega:\cX\to\mathbb R^k$ be raw network outputs.
The feature functions used throughout the algorithm are
\begin{align*}
 f_\theta(s,a)&=\widetilde f_\theta(s,a)
       -\mathbb E_{\widetilde a\sim\pi(\cdot\mid s)}[\widetilde f_\theta(s,\widetilde a)],\\
 g_\omega(y)&=\widetilde g_\omega(y)
       -\mathbb E_{Y^+\sim\nu^{+,\pi}}[\widetilde g_\omega(Y^+)].
\end{align*}
Action integrals are approximated with policy samples and marginal means with reference samples. During a feature update, gradients pass through the centering operation of that feature. Predictor targets and the opposite covariance are detached. The population objective is
\begin{align*}
 \mathcal J(\theta,\omega)
 &=\mathbb E_{X,Y^+}[f_\theta(X)^\text{T} g_\omega(Y^+)]
       -\tfrac12\tr(C_fC_g),\\
 C_f&=\mathbb E_{d^\pi}[f_\theta f_\theta^\text{T}],\\
 C_g&=\mathbb E_{\nu^{+,\pi}}[g_\omega g_\omega^\text{T}].
\end{align*}
Conditional centering restricts the learned dependence to the additional effect of the action given the state. Empirical centering and covariance estimates introduce sampling error. In particular, a noisy action-average estimate also perturbs second moments. Increasing the action-sample budget or using independent replicas for moment estimation can address this issue.

\paragraph{Why the two TD predictors provide the feature gradients.}
For fixed feature networks, the intended predictor solutions are
\[
 h(x)=\int M^{+,\pi}(y\mid x)g_\omega(y)\,dy,\qquad
 v(y)=\frac{\int d^\pi(x)M^{+,\pi}(y\mid x)f_\theta(x)\,dx}{d^\pi(y)}.
\]
Integrals denote sums in the tabular case. The positive-lag resolvent identity gives
\begin{align*}
 h(x)&=\mathbb E[(1-\gamma)g_\omega(Y_1)+\gamma h(Y_1)\mid X=x],\\
 v(y)&=w_1(y)\mathbb E[(1-\gamma)f_\theta(X)+\gamma v(X)\mid Y_1=y].
\end{align*}
Both conditional expectations here use the one-step joint law
$d^\pi(x)P_\pi(y\mid x)$. The output correction, in contrast, yields the
positive-lag backward conditional expectation:
\[
 w_+(y)v(y)=\mathbb E[f_\theta(X)\mid Y^+=y].
\]
Consequently,
\[
 \mathbb E_{X,Y^+}[f_\theta(X)^\text{T} g_\omega(Y^+)]
 =\mathbb E_{d^\pi}[f_\theta^\text{T} h]
 =\mathbb E_{\nu^{+,\pi}}[g_\omega^\text{T} w_+v].
\]
With exact predictors, ratios, centering, and moments, $\mathcal L_f$ and
$\mathcal L_g$ therefore yield the negative block gradients of $\mathcal J$.
While updating $f_\theta$, $h$ depends only on the fixed $g_\omega$. While updating
$g_\omega$, $v$ depends only on the fixed $f_\theta$. Neither update requires
differentiating through a TD solution. Finite TD fitting and statistical errors
produce approximate block gradients. The construction alone does not establish
convergence of nonlinear TD or global optimization of the neural objective.

\paragraph{Discounted-return post-calibration.}
After representation learning, we freeze the feature networks
$f_\theta$ and $g_\omega$ and recalibrate their low-dimensional
operator using an independent set of trajectory snippets.
For each anchor $X_n=Z_{t_n}$, define the truncated discounted
feature return
\begin{equation}
    Y^{g}_{n,H}
    =
    \sum_{\ell=1}^{H}
    (1-\gamma)\gamma^{\ell-1}
    \bar g_\omega(Z_{t_n+\ell}),
\end{equation}
where $\bar g_\omega$ is centered under the empirical positive-lag
marginal. Equivalently, the trajectory snippets define the
Rao--Blackwellized positive-lag joint estimator
\begin{equation}
    \widehat J_H(x,y)
    =
    \frac{1}{N}
    \sum_{n=1}^{N}
    \mathbf 1\{X_n=x\}
    \sum_{\ell=1}^{H}
    (1-\gamma)\gamma^{\ell-1}
    \mathbf 1\{Z_{t_n+\ell}=y\}.
\end{equation}
We normalize $\widehat J_H$ by
$\sum_{\ell=1}^{H}(1-\gamma)\gamma^{\ell-1}=1-\gamma^H$.
The omitted tail mass is therefore $\gamma^H$.

Let $\widehat d$ and $\widehat\nu$ denote the two marginals of
$\widehat J_H$. Using the action-centered left feature
$\bar f_\theta$ and marginal-centered right feature
$\bar g_\omega$, we estimate
\begin{align}
    \widehat C_f
    &=
    \mathbb E_{\widehat d}
    [\bar f_\theta(X)\bar f_\theta(X)^\top],
    \\
    \widehat C_g
    &=
    \mathbb E_{\widehat\nu}
    [\bar g_\omega(Y)\bar g_\omega(Y)^\top],
    \\
    \widehat K_H
    &=
    \mathbb E_{\widehat J_H}
    [\bar f_\theta(X)\bar g_\omega(Y)^\top]
    =
    \frac{1}{N}\sum_{n=1}^{N}
    \bar f_\theta(X_n)(Y^g_{n,H})^\top .
\end{align}
Unlike a single geometric-lag endpoint, the discounted feature return
integrates over all future lags in the snippet and removes the
additional geometric-lag sampling variance.

We use ridge-regularized whitening maps
\begin{equation}
    W_f^{\lambda_f}
    =
    (\widehat C_f+\lambda_f I)^{-1/2},
    \qquad
    W_g^{\lambda_g}
    =
    (\widehat C_g+\lambda_g I)^{-1/2},
\end{equation}
and compute the thin SVD
\begin{equation}
    W_f^{\lambda_f}
    \widehat K_H
    W_g^{\lambda_g}
    =
    U\widehat\Sigma V^\top .
\end{equation}
The corresponding calibrated features are
\begin{equation}
    \widehat f(x)
    =
    U^\top W_f^{\lambda_f}\bar f_\theta(x),
    \qquad
    \widehat g(y)
    =
    V^\top W_g^{\lambda_g}\bar g_\omega(y).
\end{equation}
Because of ridge regularization, these features satisfy
\begin{equation}
    W_f^{\lambda_f}
    (\widehat C_f+\lambda_f I)
    (W_f^{\lambda_f})^\top
    =I,
\end{equation}
and analogously for $g$. They are not exactly whitened with respect
to $\widehat C_f$ and $\widehat C_g$ when the corresponding ridge
parameters are nonzero.

Optionally, we continuously shrink the calibrated singular values:
\begin{equation}
    \widetilde\sigma_i
    =
    a_i\widehat\sigma_i,
    \qquad
    a_i
    =
    \frac{\widehat\sigma_i^2}
         {\widehat\sigma_i^2+\tau^2}.
\end{equation}
The resulting rank-$k$ operator is
\begin{equation}
    \widehat{\mathcal T}^{+,k}r
    =
    \sum_{i=1}^{k}
    \widetilde\sigma_i\,
    \widehat f_i
    \left\langle
        \widehat g_i,r
    \right\rangle_{\widehat\nu}.
\end{equation}

The ridge parameters $(\lambda_f,\lambda_g)$ are selected on an independent calibration-validation split. After fixing them, the shrinkage parameter $\tau$ is selected on the same validation criterion without using any downstream task reward. The reported advantage errors are evaluated only after all calibration hyperparameters have been fixed.


\paragraph{What is learned.}
The returned functions approximate the reward-independent positive-lag,
action-centered operator through
\[
 (\Tp r)(s,a)\approx\widehat f(s,a)^\text{T}\widehat\Sigma
       \mathbb E_{Y^+\sim\nu^{+,\pi}}[\widehat g(Y^+)r(Y^+)].
\]
For a downstream reward, the corresponding advantage estimate is
\begin{align*}begin{align*}
 \widehat A_r^\pi(s,a)&&=C_\pi r(s,a)
 +\frac{\gamma}{1-\gamma}\widehat f(s,a)^\text{T}\widehat\Sigma
       \mathbb E_{\nu^{+,\pi}}[\widehat g\,r],\\\\
 C_\pi r(s,a)&&=r(s,a)-\mathbb E_{a'\sim\pi(\cdot\mid s)}[r(s,a')].
\end{align*}end{align*}
Training uses no downstream reward. The two predictor networks propagate feature information through one-step transitions. Soft-HGR updates the feature networks and canonicalization fixes their final coordinates. Sample-efficiency gains over
direct positive-lag training require empirical validation.


\section{Environment and Evaluation Details}
\label{app:environment_details}

This section specifies the finite-state environments used for the spectral
diagnostics and one-step reward-transfer experiments.  In every experiment,
the transition kernel, initial distribution, and reference policy are fixed
before the reward is selected.  Consequently, changing a downstream reward
does not change the data-generating process or the learned representation.
Unless stated otherwise, we use $\gamma=0.9$.

\subsection{Discrete Finite Environments}
\subsubsection{Navigation environments}
\label{app:navigation_environments}

Both navigation domains have the five actions
$\{\textsc{up},\textsc{right},\textsc{down},\textsc{left},\textsc{stay}\}$.
Transitions are deterministic.  A movement into a wall or outside the map
leaves the agent in its current cell. The \textsc{stay} action also leaves the
state unchanged.  The initial distribution $\mu$ is uniform over all free
cells.  Rewards do not terminate an episode: we evaluate the corresponding
continuing discounted MDP exactly.

\begin{figure}[t]
    \centering
    \includegraphics[width=\linewidth]{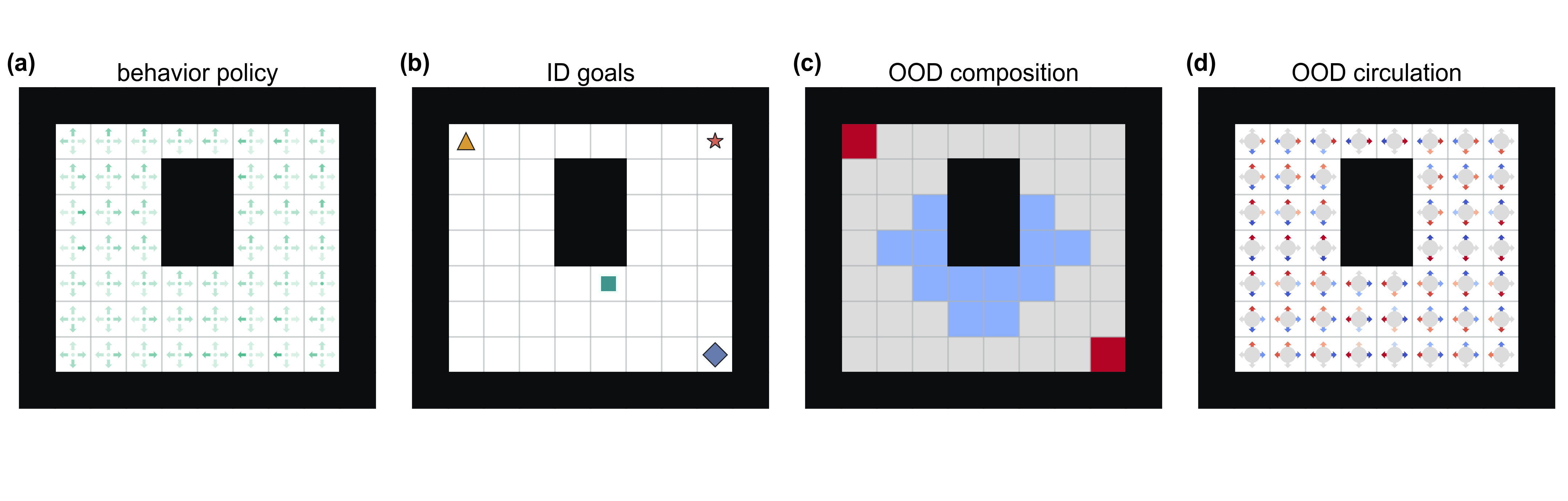}
    \caption{Behavior policy, environment, in-distribution and out-of-distribution tasks for grid world.}
    \label{fig:env_grid_world}
\end{figure}

\begin{figure}[ht]
    \centering
    \includegraphics[width=\linewidth]{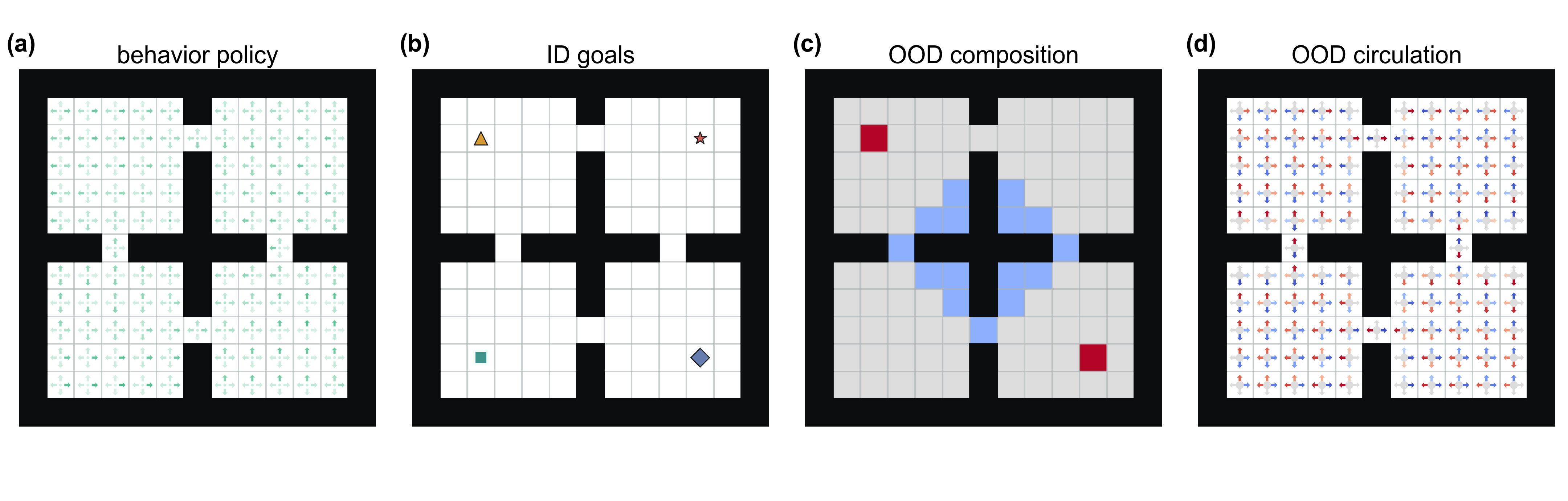}
    \caption{Behavior policy, environment, in-distribution and out-of-distribution tasks for Four-room.}
    \label{fig:env_four_room}
\end{figure}

\paragraph{Grid World.}
As is shown in Figure \ref{fig:env_grid_world}, Grid World uses a $7\times 8$ canvas.  The six cells
\[
 \{(i,j):i\in\{1,2,3\},\ j\in\{3,4\}\}
\]
are blocked, leaving $50$ states and $250$ state--action pairs.  Coordinates
are zero-indexed from the top-left corner.  The four in-distribution (ID) goal
locations are top-left $(0,0)$, top-right $(0,7)$, center $(4,4)$, and
bottom-right $(6,7)$.  The population positive-lag DCDM has nonzero rank $49$
under the reference policy used below.

\paragraph{Four-room.}
As is shown in Figure \ref{fig:env_four_room}, Four-room uses an $11\times11$ canvas.  An interior vertical wall lies at
column $5$ and an interior horizontal wall lies at row $5$, with doorways at
$(1,5)$, $(8,5)$, $(5,2)$, and $(5,8)$.  Removing the wall cells leaves $104$
states and $520$ state--action pairs.  The four ID goals are placed at
$(1,1)$, $(1,9)$, $(9,1)$, and $(9,9)$.  The corresponding population DCDM
rank is $103$.

\begin{table}[t]
  \centering
  \caption{Finite-state environments used in the reward-transfer experiments.
  The DCDM rank is computed for the fixed reference policy at $\gamma=0.9$.}
  \label{tab:environment_summary}
  \small
  \begin{tabular}{lrrrrl}
    \toprule
    Environment & Grid size & $|\mathcal S|$ & $|\mathcal A|$
      & DCDM rank & Reference policy \\
    \midrule
    Grid World & $7\times8$ & 50 & 5 & 49 & 5 balanced \\
    Four-room & $11\times11$ & 104 & 5 & 103 & 5 balanced \\
    Taxi & $5\times5$ & 500 & 6 & 399 & 5 balanced \\
    \bottomrule
  \end{tabular}
\end{table}

\subsubsection{Reference policies and discounted distributions}
\label{app:reference_policy}

The navigation rank-control figures use one reproducible, full-support
\emph{fixed balanced} policy for each environment.  Let
$\pi_{\mathrm{sm}}$ be a smooth random policy obtained by applying a softmax to fixed-seed Fourier features of the cell coordinates.  We use
\[
 \pi_0(a\mid s)
 =0.2\times \frac{1}{5}+0.8\,\pi_{\mathrm{sm}}(a\mid s).
\]
The uniform component guarantees $\pi_0(a\mid s)\geq0.04$.  The same realized
policy is used to collect reward-free data, define the SUAF operator, and
evaluate all rewards within an environment.  It is not resampled per reward or
per retained rank.  Experiments explicitly described as the uniform
reward-free baseline protocol instead use $\pi_0(a\mid s)=1/5$. Those results
are not mixed with the fixed-balanced rank-control curves.

For a state--action pair $X_t=(S_t,A_t)$, the normalized discounted occupancy
and positive-lag marginal are
\begin{align}
 d^{\pi_0}(x)
 &= (1-\gamma)\sum_{t\geq0}\gamma^t
    \Pr_{\mu,\pi_0}(X_t=x),\\
 \nu^{+,\pi_0}(y)
 &= \sum_x d^{\pi_0}(x)M^{+,\pi_0}(y\mid x),
\end{align}
where
$M^{+,\pi_0}(y\mid x)=(1-\gamma)\sum_{\ell\geq1}
\gamma^{\ell-1}\Pr_{\pi_0}(X_\ell=y\mid X_0=x)$.
All task rewards are normalized once using the reference-policy outcome
marginal,
\begin{equation}
 \widetilde r
 =\frac{r}{\|r\|_{\nu^{+,\pi_0}}},\qquad
 \|r\|_{\nu^{+,\pi_0}}^2
 =\sum_x\nu^{+,\pi_0}(x)r(x)^2.
 \label{eq:appendix_reward_normalization}
\end{equation}
The normalization is fixed before policy improvement and is not recomputed
under the improved policy.

\subsubsection{ID and OOD reward families}
\label{app:id_ood_rewards}

\paragraph{Sparse ID goals.}
For a goal cell $g$, the unnormalized reward is state-only,
\begin{equation}
 r_g(s,a)=\mathbf 1\{s=g\},
\end{equation}
for every action.  Grid World uses the four goal cells listed above. Four-room uses one goal inside each room.  These sparse goals form the ID family used
while selecting the representation protocol.

\paragraph{OOD state composition.}
This reward composes two diagonal ID goals and introduces a signed component
that was absent from the ID family:
\begin{equation}
 r_{\mathrm{comp}}(s,a)
 =r_{\mathrm{TL}}(s,a)+r_{\mathrm{BR}}(s,a)
  -\tfrac12\mathbf 1\{s\in\mathcal H\}.
 \label{eq:appendix_composition_reward}
\end{equation}
Here $\mathcal H$ contains all valid cells whose Manhattan distance to the
geometric map center is no larger than the fifth order statistic. Ties may
therefore make $|\mathcal H|>5$.  Thus this task tests a new signed combination of spatial events,
rather than another single-goal readout.

\paragraph{OOD state--action circulation.}
Let $c$ be the geometric map center, $p_s\in\mathbb R^2$ the row--column
coordinate of state $s$, and
\[
 t_s=\frac{(p_{s,2}-c_2,-(p_{s,1}-c_1))}
 {\max\{\|p_s-c\|_2,1\}}
\]
the clockwise tangent direction.  The circulation reward is
\begin{equation}
 r_{\mathrm{circ}}(s,a)
 =t_s^\top\left(\mathbb E[S'\mid s,a]-p_s\right).
 \label{eq:appendix_circulation_reward}
\end{equation}
Unlike all ID goals, this reward depends explicitly on the action and favors
clockwise motion around the map center. The
\eqref{eq:appendix_composition_reward} and
\eqref{eq:appendix_circulation_reward} are normalized using \eqref{eq:appendix_reward_normalization} before evaluation.

\subsubsection{Continuing Taxi}
\label{app:taxi_environment}

\begin{figure}[H]
    \centering
    \includegraphics[width=\linewidth]{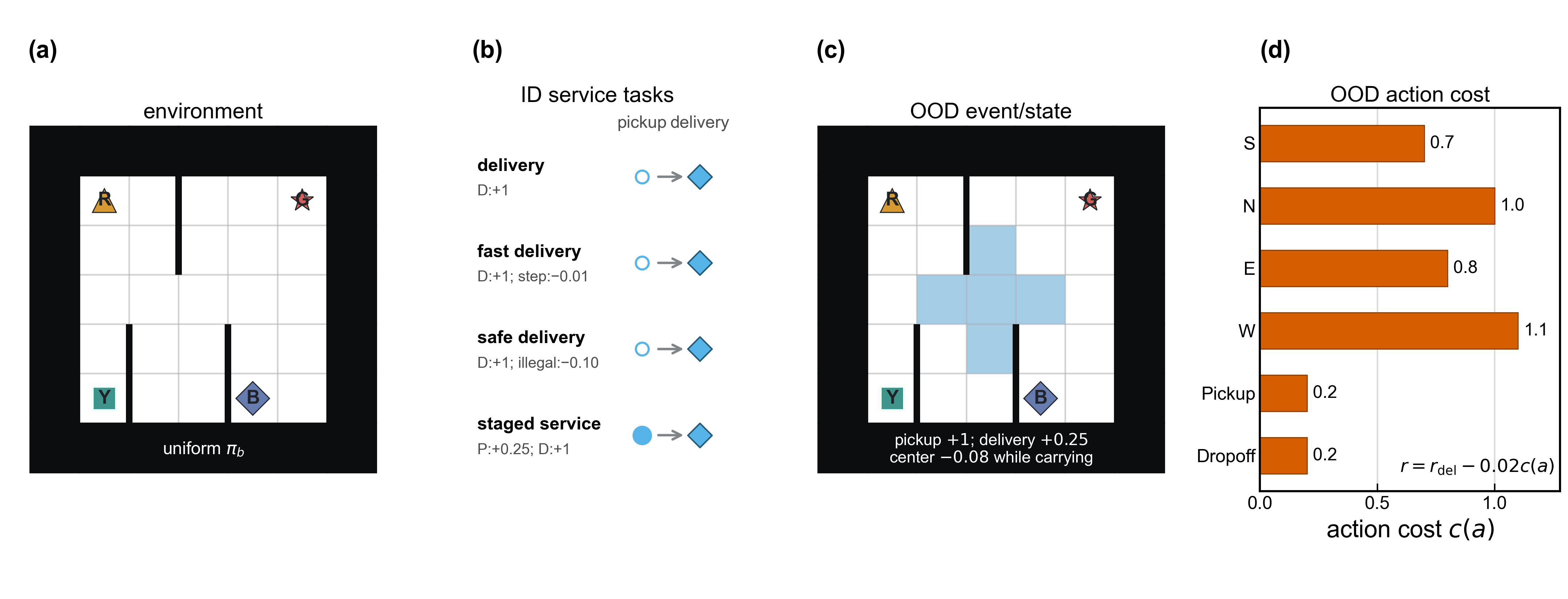}
    \caption{Environment, in-distribution and out-of-distribution tasks for taxi.}
    \label{fig:env_taxi}
\end{figure}

As is shown in Figure \ref{fig:env_taxi}, Taxi uses the Taxi-v3 $5\times5$ road map with $500$ encoded states and six
actions: four movements, pickup, and dropoff.  A state records the taxi
position, one of four passenger landmarks (or that the passenger is inside
the taxi), and one of four destinations.  The initial law is uniform over taxi
positions and over passenger--destination pairs with distinct landmarks,
giving support on $25\times4\times3=300$ encoded states.  Successful delivery
is followed by a draw from this same initial law, producing a continuing MDP
without changing the Taxi-v3 within-task dynamics.  The reference policy is
uniform over the six actions, and the population DCDM rank is $399$.

The four ID rewards are: successful delivery. Delivery with a $0.01$ step
cost. Delivery with a $0.10$ illegal pickup/dropoff penalty, and delivery plus
$0.25$ for successful pickup.  The three OOD rewards change the task structure:
(i) pickup priority, $0.25$ times delivery plus pickup; (ii) center hazard,
delivery minus $0.08$ whenever a passenger-carrying taxi is within Manhattan
distance one of the map center; and (iii) action-dependent fuel cost, delivery
minus $0.02$ times the fixed action-cost vector
$(0.7,1.0,0.8,1.1,0.2,0.2)$.  As in the navigation domains, every Taxi reward
is normalized under $\nu^{+,\pi_0}$, while dynamics, reset law, and reference
policy remain unchanged.

\paragraph{Tasks}

We evaluate reward transfer while holding the MDP dynamics, state and action spaces, behavior policy, and offline replay distribution fixed. Thus, ``OOD'' refers exclusively to a shift in the reward family rather than to a dynamics or observation-distribution shift. In Grid World and Four-room, the ID tasks are four sparse single-goal rewards located in different regions of the environment. We construct two structurally distinct OOD tasks: a compositional reward that combines two distant goals with an unseen negative center hazard, and an action-dependent circulation reward that favors clockwise motion around the center of the map. In continuing Taxi, the ID tasks are four related service objectives: successful delivery, fast delivery with a step cost, safe delivery with an illegal-action penalty, and staged service with an intermediate pickup bonus. The Taxi OOD tasks alter the task structure more substantially by prioritizing pickup over delivery, penalizing carrying the passenger through the central region, or introducing action-dependent fuel costs. All rewards are normalized to unit $\nu^{+,\pi_b}$-weighted norm, and the OOD rewards are excluded from model selection and hyperparameter tuning.

\subsection{Continuous Environments}
\label{app:continuous_environments}

\begin{figure}
    \centering
    \includegraphics[width=\linewidth]{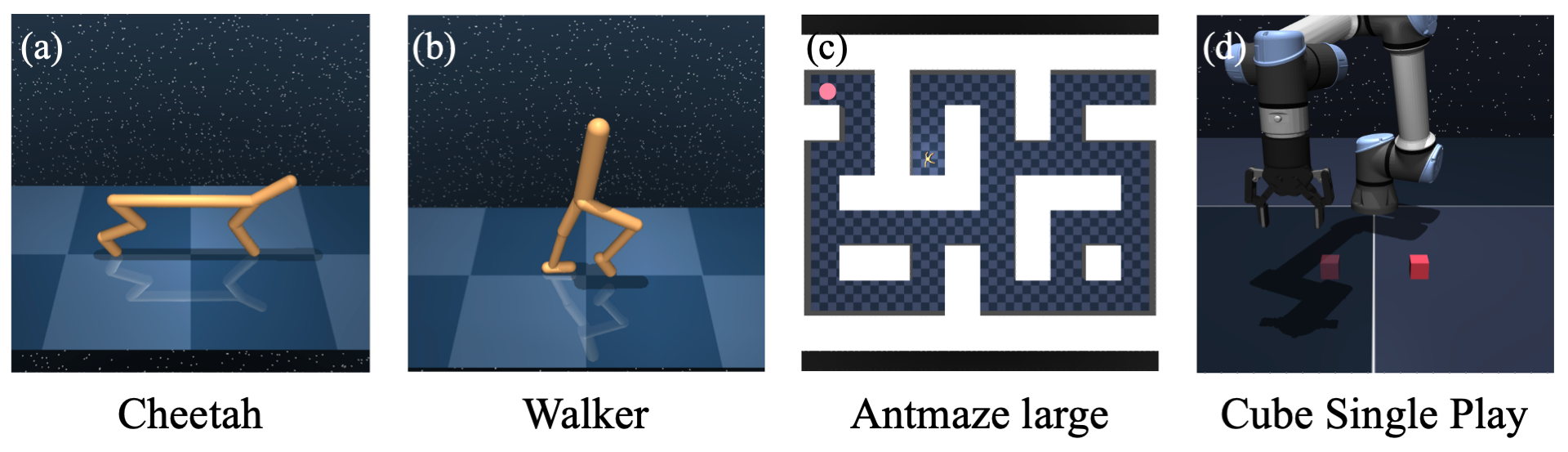}
    \caption{Environments of ExORL and OGBench.}
    \label{fig:continuous_env}
\end{figure}

We evaluate reward-independent representation learning and downstream reward transfer on continuous-control environments from ExORL and OGBench as shown in Figure \ref{fig:continuous_env}. Within each domain, all downstream tasks share the same observation space, action space, transition dynamics, and offline transition dataset. The representation is learned without access to task rewards. At evaluation time, different control problems are created by changing only the downstream reward or goal. This protocol tests whether a single representation of the dynamics can be reused across multiple tasks.

\paragraph{ExORL locomotion environments.}
We use the \textit{walker} and \textit{cheetah} domains from ExORL. These environments evaluate transfer among locomotion objectives that share the same underlying physical system but favor different behaviors, velocities, or directions of motion.

For \textit{walker}, we consider four tasks:
\begin{enumerate}
    \item \textit{flip}, which rewards completing a flipping motion;
    \item \textit{run}, which rewards fast forward locomotion;
    \item \textit{stand}, which rewards maintaining an upright stationary pose;
    \item \textit{walk}, which rewards forward locomotion at a walking speed.
\end{enumerate}
The tasks therefore range from static stabilization to periodic locomotion and
agile whole-body motion.

For \textit{cheetah}, we consider four tasks:
\begin{enumerate}
    \item \textit{run}, which rewards fast forward locomotion;
    \item \textit{run backward}, which rewards fast locomotion in the reverse
    direction;
    \item \textit{walk}, which rewards slower forward locomotion;
    \item \textit{walk backward}, which rewards slower locomotion in the reverse
    direction.
\end{enumerate}
The forward and backward tasks test whether the learned representation can be reused when the preferred direction of motion changes, while the walking and running tasks additionally vary the desired locomotion regime.

\paragraph{OGBench AntMaze Large Navigate.}
This domain requires a continuous-control agent to navigate between distant regions of a large maze. Each task is specified by a starting region and a target region. The maze geometry and transition dynamics remain fixed, but different start-goal pairs may require substantially different long-horizon routes. We consider five tasks:
\begin{enumerate}
    \item bottom left $\rightarrow$ top right;
    \item center $\rightarrow$ top left;
    \item center $\rightarrow$ bottom right;
    \item bottom right $\rightarrow$ center;
    \item bottom left $\rightarrow$ center.
\end{enumerate}
These tasks evaluate whether the learned representation captures the
long-horizon connectivity of the maze well enough to support transfer among
different navigation objectives.

\paragraph{OGBench Cube Single Play.}
This domain is a continuous robotic manipulation problem in which the agent
must pick up a cube and place it at a specified target location. The robot,
object, and contact dynamics are shared by all tasks; only the desired
placement target changes. We evaluate five tasks:
\begin{enumerate}
    \item pick and place to the left;
    \item pick and place to the front;
    \item pick and place to the back;
    \item pick and place along a diagonal direction;
    \item pick and place along an off-diagonal direction.
\end{enumerate}
Unlike the locomotion and navigation domains, successful transfer in this
environment requires the representation to capture contact-rich manipulation
dynamics, including both grasping and task-dependent object transport.

For every domain, \textit{overall} denotes the arithmetic mean of the
evaluation score over all downstream tasks in that domain.

\section{Experimental details}
\label{app:exp_details}
\subsection{Data processing}

\paragraph{Uncertainty estimation.} For each experimental configuration, we repeat the experiment with \(N\) independent random seeds. Let \(x_i\) denote the evaluation metric from the \(i\)-th run at a given rank or training checkpoint. We compute the mean as \(\bar{x}=N^{-1}\sum_{i=1}^{N}x_i\) and the sample standard deviation (std) as \(s=\sqrt{(N-1)^{-1}\sum_{i=1}^{N}(x_i-\bar{x})^2}\). The 95\% confidence interval (CI) for the mean is \(\bar{x}\pm t_{0.975,N-1}s/\sqrt{N}\), where \(t_{0.975,N-1}\) is the 97.5th percentile of Student’s \(t\)-distribution with \(N-1\) degrees of freedom. Standard deviations describe variability across runs, whereas confidence intervals quantify uncertainty in the estimated mean. Intervals are computed separately at each plotted point on the original metric scale, before any axis transformation, and represent pointwise rather than simultaneous confidence intervals.

\subsection{Implementation details and hyperparameters of discrete environments}
\label{app:implementation_hyperparameters}

Tables~\ref{tab:suaf_data_hyperparameters}--\ref{tab:control_hyperparameters}
report the resolved settings used to produce the learned-representation and
one-step-control results.  We distinguish an \emph{optimizer update} from an
outer bidirectional iteration: one outer iteration performs, in order, one update of the forward predictor $h$, the left feature $f$, the reverse predictor $v$, and the right feature $g$, and therefore contains four optimizer updates.  Unless noted otherwise, training tensors use single precision and the small-matrix calibration is performed in double precision.

\begin{table*}[t]
  \centering
  \caption{\textbf{Environment, replay, and representation dimensions.}  A geometric positive lag has law $\Pr(L=\ell)=(1-\gamma)\gamma^{\ell-1}$ for
  $\ell\geq1$. The Taxi model outputs 499 coordinates as an overparameterization diagnostic, while the population DCDM has rank 399. Coordinates beyond rank 399 do not correspond to additional nonzero population SUAF modes.
}
  \label{tab:suaf_data_hyperparameters}
  \small
  \setlength{\tabcolsep}{4.0pt}
  \resizebox{\textwidth}{!}{
  \begin{tabular}{lccc}
    \toprule
    Hyperparameter & GridWorld & Four-room & Taxi \\
    \midrule
    States / actions & $50/5$ & $104/5$ & $500/6$ \\
    State--action indices & $250$ & $520$ & $3000$ \\
    Population DCDM rank & $49$ & $103$ & $399$ \\
    Learned output dimension $k$ & $49$ & $103$ & $499$ \\
    Discount $\gamma$ & $0.9$ & $0.9$ & $0.9$ \\
    Reference policy & fixed balanced & fixed balanced & uniform \\
    Initial law $\mu$ & uniform over free cells & uniform over free cells
      & Taxi-v3 law (300-state support) \\
    Replay transitions & $5\times10^6$ & $10^7$ & $5\times10^6$ \\
    Trajectory horizon & $100$ & $100$ & $100$ \\
    Number of trajectories & $50{,}000$ & $100{,}000$ & $50{,}000$ \\
    Occupancy-anchor sampling & discounted & discounted & discounted \\
    Positive-lag sampling & geometric$(1-\gamma)$ & geometric$(1-\gamma)$
      & geometric$(1-\gamma)$ \\
    Minibatch size & $1024$ & $1024$ & $1024$ \\
    \bottomrule
  \end{tabular}}
\end{table*}

\begin{table*}[t]
  \centering
  \caption{\textbf{Bidirectional-TD SUAF optimization.}  The implementation uses
  trainable tables (embedding matrices), not hidden-layer MLPs.  Thus the table
  dimensions below fully specify the representation architecture.  Target
  copies of $h$ and $v$ have the same shapes but are not optimized directly.}
  \label{tab:suaf_optimization_hyperparameters}
  \small
  \setlength{\tabcolsep}{3.6pt}
  \begin{tabular}{lccc}
    \toprule
    Hyperparameter & GridWorld & Four-room & Taxi \\
    \midrule
    Left feature $f(s,a)$ & $250\times49$ & $520\times103$ & $3000\times499$ \\
    Right feature $g(s)$ & $50\times49$ & $104\times103$ & $500\times499$ \\
    Forward predictor $h(s,a)$ & $250\times49$ & $520\times103$ & $3000\times499$ \\
    Reverse predictor $v(s,a)$ & $250\times49$ & $520\times103$ & $3000\times499$ \\
    Hidden layers / activation & none / none & none / none & none / none \\
    Trainable scalar parameters & $39{,}200$ & $171{,}392$ & $4{,}740{,}500$ \\
    Initialization & \multicolumn{3}{c}{$0.2\,\mathcal N(0,I)$ independently for $f,g,h,v$} \\
    Feature ordering & \multicolumn{3}{c}{sequential nested-SVD mask} \\
    Action centering & \multicolumn{3}{c}{exact expectation under the fixed reference policy} \\
    Right-feature action dependence & \multicolumn{3}{c}{tied across actions: $g(s,a)\equiv g(s)$} \\
    Reverse-TD objective & \multicolumn{3}{c}{boundary-corrected matched squared loss} \\
    Reverse successor coefficient & \multicolumn{3}{c}{$\gamma/2=0.45$} \\
    Reverse boundary coefficient & \multicolumn{3}{c}{$(1-\gamma)/2=0.05$} \\
    Optimizer & \multicolumn{3}{c}{four independent Adam optimizers, one for each of $f,g,h,v$} \\
    Adam $(\beta_1,\beta_2,\epsilon)$ & \multicolumn{3}{c}{$(0.9,0.999,10^{-8})$} \\
    Weight decay / AMSGrad & \multicolumn{3}{c}{$0$ / false} \\
    Initial learning rate & \multicolumn{3}{c}{$3\times10^{-3}$ for all four optimizers} \\
    Learning-rate schedule & \multicolumn{3}{c}{multiply by $0.5$ per $10^5$ optimizer updates; floor $3\times10^{-5}$} \\
    Gradient-norm clipping & \multicolumn{3}{c}{$10$ (separately for each update)} \\
    Polyak coefficient $\tau$ & \multicolumn{3}{c}{$0.2$} \\
    Optimizer updates / outer iteration & \multicolumn{3}{c}{$4$ ($h$, $f$, $v$, then $g$)} \\
    Training budget & $630{,}000$ & $2{,}000{,}000$ & $100{,}000$ \\
    Outer iterations & $157{,}500$ & $500{,}000$ & $25{,}000$ \\
    Diagnostic interval & $10{,}000$ updates & $10{,}000$ updates & $10{,}000$ updates \\
    \bottomrule
  \end{tabular}
\end{table*}

The reverse predictor uses the same paired one-step transition $(X,Y_1)$ in
its quadratic and cross terms.  Its boundary minibatch is sampled independently
from the known $d_0(s,a)=\mu(s)\pi_0(a\mid s)$. No population transition matrix
or population SUAF mode is supplied to training.  Target tables are updated as
$\bar h\leftarrow(1-\tau)\bar h+\tau h$ and analogously for $v$.  The learning
rate at optimizer update $t$ is
$\max\{3\times10^{-5},3\times10^{-3}\,0.5^{t/10^5}\}$.

\begin{table*}[t]
  \centering
  \caption{\textbf{Reward-free checkpoint selection and post-hoc calibration.}  For
  GridWorld and Four-room, $\lambda_f=s_f\operatorname{tr}(C_f)/k$ and
  $\lambda_g=s_g\operatorname{tr}(C_g)/k$.  Shrinkage uses
  $\widetilde\sigma_i=\sigma_i^3/(\sigma_i^2+\tau_\sigma^2)$ with
  $\tau_\sigma=s_\sigma\sigma_1$.}
  \label{tab:suaf_calibration_hyperparameters}
  \small
  \setlength{\tabcolsep}{3.5pt}
  \resizebox{\textwidth}{!}{%
  \begin{tabular}{lccc}
    \toprule
    Hyperparameter & GridWorld & Four-room & Taxi \\
    \midrule
    Calibration construction & discounted feature return & discounted feature return
      & sampled geometric positive lag \\
    Calibration training pairs & $500{,}000$ & $500{,}000$ & $500{,}000$ \\
    Calibration validation pairs & $250{,}000$ & $250{,}000$ & $500{,}000$ \\
    Calibration rollout horizon & $128$ & $128$ & at most $100$ within trajectory \\
    Omitted geometric tail mass & $1.39\times10^{-6}$ & $1.39\times10^{-6}$ & -- \\
    Covariance inverse-sqrt tolerance & $10^{-12}$ & $10^{-12}$ & $10^{-12}$ \\
    Ridge-scale candidates & \multicolumn{2}{c}{$\{0,10^{-7},10^{-6},10^{-5},10^{-4},10^{-3},10^{-2},10^{-1}\}$}
      & fixed $10^{-3}$ \\
    Selected $(s_f,s_g)$ & $(10^{-3},10^{-3})$ & $(10^{-3},10^{-3})$
      & $(10^{-3},10^{-3})$ \\
    Shrinkage-scale candidates & \multicolumn{2}{c}{$\{0,10^{-4},3\!\times\!10^{-4},10^{-3},3\!\times\!10^{-3},10^{-2},3\!\times\!10^{-2},10^{-1}\}$}
      & not swept \\
    Selected $s_\sigma$ & $0$ & $0$ & $0$ (no shrinkage) \\
    Small-matrix factorization & \multicolumn{3}{c}{full SVD of the ridge-whitened $k\times k$ cross moment} \\
    Selection metric & \multicolumn{3}{c}{independent reward-free weighted operator discrepancy} \\
    Selected optimizer updates & $630$k for every run
      & $1860/1080/1490/1730/1310$k & $20/30/20/40$k \\
    \bottomrule
  \end{tabular}
  }
\end{table*}

The calibration split is never used to optimize the raw feature tables.
GridWorld and Four-room share the same independently generated calibration
and validation joint distributions across representation seeds.  Taxi first
draws $10^6$ independent calibration pairs from each seed's replay and uses a
fixed half/half split.  Taxi's ridge scale is fixed rather than selected by the
two-stage navigation-domain validation sweep. This difference is reported
explicitly rather than treating the two procedures as identical.

\begin{table*}[t]
  \centering
  \caption{\textbf{Downstream one-step control and shared settings of the learned
  baseline adapter.}  ``Exact'' refers only to finite-MDP evaluation after
  reward-free representation training. It does not supply oracle quantities to
  SUAF training.}
  \label{tab:control_hyperparameters}
  \small
  \setlength{\tabcolsep}{4pt}
  \resizebox{\textwidth}{!}{
  \begin{tabular}{lccc}
    \toprule
    Hyperparameter & GridWorld & Four-room & Taxi \\
    \midrule
    SUAF rank grid & $0,5,\ldots,45,49$ & $0,10,\ldots,100,103$
      & $0,20,\ldots,380,399,400,420,\ldots,480,499$ \\
    Boltzmann inverse temperatures & \multicolumn{3}{c}{$\{4,32,256\}$} \\
    Additional policy & \multicolumn{3}{c}{greedy $\arg\max_a\widehat A(s,a)$} \\
    Number of policy updates & \multicolumn{3}{c}{$1$ No actor training and no feature refresh} \\
    Reward normalization & \multicolumn{3}{c}{$\|r\|_{\nu^{+,\pi_0}}=1$, fixed under the reference policy} \\
    Return computation & \multicolumn{3}{c}{exact finite Bellman solve from $\mu$} \\
    Optimal-return denominator & \multicolumn{3}{c}{exact value iteration, tolerance $10^{-12}$} \\
    \midrule
    Baseline pretraining updates & $200{,}000$ & $200{,}000$ & $100{,}000$ \\
    Baseline minibatch size & $1024$ & $1024$ & $1024$ \\
    Baseline hidden widths & $(256,256)$ & $(256,256)$ & $(256,256)$ \\
    FB / One-Step FB latent dimension & $49$ & $103$ & $499$ \\
    Other-baseline latent dimension & $50$ & $104$ & $500$ \\
    Baseline reward-readout samples & $120{,}000$ & $120{,}000$ & population readout \\
    Baseline control candidates & \multicolumn{3}{c}{greedy; $\eta\in\{0.25,1,4,16,64,256\}$; matched KL $\in\{0.01,0.05,0.1\}$} \\
    Baseline evaluation interval & $10{,}000$ & $10{,}000$ & $5{,}000$ updates \\
    \bottomrule
  \end{tabular}}
\end{table*}

For Laplacian, BYOL-$\gamma$, TD-JEPA, ICVF, HILP, FB, and One-Step FB, we
start from the upstream configurations in the public One-Step-FB implementation
(revision \texttt{fea347e}), replace every representation, predictor, value,
and actor hidden-width tuple exposed by the configuration with $(256,256)$,
set the latent dimension as in Table~\ref{tab:control_hyperparameters}, set
$\gamma=0.9$, and disable the continuous-action tanh squash.  Our adapter never
optimizes or evaluates the continuous actor: every action maximization is an
exact enumeration of the finite action set.  Actor-only coefficients are
therefore inactive and omitted.  The remaining active method-specific settings
are listed in Table~\ref{tab:baseline_method_hyperparameters}.

\begin{table*}[t]
  \centering
  \caption{\textbf{Active method-specific baseline settings after applying the common
  discrete adapter.}  All methods use a two-hidden-layer MLP with widths
  $(256,256)$ and GELU activations, Optax Adam with learning rate $10^{-4}$,
  target-update coefficient $0.005$, normalized latents, and no image encoder.
  Goal mixtures are ordered as current / geometric same-trajectory / random.
  Here $d$ denotes the environment-specific latent dimension in
  Table~\ref{tab:control_hyperparameters}.}
  \label{tab:baseline_method_hyperparameters}
  \small
  \setlength{\tabcolsep}{3.2pt}
  \resizebox{\textwidth}{!}{%
  \begin{tabular}{lllll}
    \toprule
    Method & Layer normalization & Goal mixture & Aggregation & Active extra setting \\
    \midrule
    Laplacian & none & $0.2/0.5/0.3$ & Q: mean
      & orthonormality coefficient $0$ \\
    BYOL-$\gamma$ & none & $0/1/0$ & Q: mean
      & orthonormality coefficient $0$ \\
    TD-JEPA & representation and predictor & $0.2/0.5/0.3$
      & representation: mean; Q: min & orthonormality $1$; $\phi$ dimension $2d$ \\
    ICVF & none & $0/0.625/0.375$ & Q: mean
      & expectile $0.5$ \\
    HILP & none & $0/0.625/0.375$ & representation: mean; Q: min
      & expectile $0.5$; single-state latent reward \\
    FB & none & $0.2/0.5/0.3$ & representation: mean
      & orthonormality $0$; reward temperature $10$ \\
    One-Step FB & forward and backward representations & $0.2/0.5/0.3$
      & representation: mean; Q: min
      & orthonormality $0$; reward temperature $10$ \\
    \bottomrule
  \end{tabular}
  }
\end{table*}

\subsection{Continuous-Control Environments and Implementation Details}
\label{app:continuous_implementation}

This section specifies the implementation used for the state-based ExORL
domains \textsc{Walker} and \textsc{Cheetah}, and the state-based OGBench
domains \textsc{AntMaze Large Navigate} and \textsc{Cube Single Play}.  We
align the data protocol, network widths, actor parameterization, optimization
budget, zero-shot evaluation, and reporting convention with the public
One-Step FB implementation.  The deliberate SUAF-specific differences are
the positive-lag operator, action-centering of the left feature, nested mode
ordering, bidirectional TD predictors, and post-hoc canonicalization.

\subsubsection{Reproducibility convention.}
The resolved settings below follow the executable commands and defaults in the
public One-Step FB repository rather than silently combining them with prose
defaults.  In particular, the released ExORL command uses a three-layer
backward MLP of width $256$ and the released agent enables LayerNorm in the
forward and backward representations.  Table~3 of the accompanying One-Step
FB paper instead lists a two-layer backward MLP of width $512$ and states that
LayerNorm is disabled.  We use the released command/code configuration and
record this discrepancy here.

\subsubsection{Datasets and evaluation protocol}

\begin{table}[H]
  \centering
  \caption{\textbf{Dataset and evaluation protocol.}  These values reproduce the state-based protocol of One-Step FB. ``Final score'' denotes averaging the
  evaluations at the last three checkpoints, not selecting the best checkpoint.}
  \label{tab:continuous_data_protocol}
  \small
  \setlength{\tabcolsep}{3.5pt}
  \resizebox{\textwidth}{!}{%
  \begin{tabular}{lcccc}
    \toprule
    Hyperparameter & Walker & Cheetah & AntMaze Large & Cube Single Play \\
    \midrule
    Benchmark & ExORL & ExORL & OGBench & OGBench \\
    Offline behavior data & RND & RND & play & play \\
    Representation-training transitions
      & $5\times10^6$ & $5\times10^6$ & $10^6$ & $10^6$ \\
    Zero-shot inference transitions
      & $10^5$ & $10^5$ & $10^5$ & $10^5$ \\
    Reward used for zero-shot inference
      & dense task reward & dense task reward & semi-sparse & semi-sparse \\
    Discount $\gamma$ & $0.98$ & $0.98$ & $0.99$ & $0.99$ \\
    Maximum evaluation horizon & $1000$ & $1000$ & $1000$ & $200$ \\
    Gradient steps & $10^6$ & $10^6$ & $10^6$ & $10^6$ \\
    Evaluation interval & $10^5$ & $10^5$ & $10^5$ & $10^5$ \\
    Episodes per task/checkpoint & $50$ & $50$ & $50$ & $50$ \\
    Evaluation metric & undiscounted return & undiscounted return
      & success rate & success rate \\
    Final reported checkpoints
      & $800/900/1000$k & $800/900/1000$k
      & $800/900/1000$k & $800/900/1000$k \\
    Independent random seeds & $8$ & $8$ & $8$ & $8$ \\
    Seed identifiers & \multicolumn{4}{c}{stored verbatim in every run manifest} \\
    Uncertainty summary & mean $\pm$ std. & mean $\pm$ std.
      & mean $\pm$ std. & mean $\pm$ std. \\
    \bottomrule
  \end{tabular}}
\end{table}

The ExORL training and zero-shot splits are generated from disjoint portions
of the same exploratory dataset.  OGBench uses its official independent train
and validation files.  Downstream rewards are never used to optimize the
reward-free representation.  They are first introduced when computing the
task coordinate and evaluating the resulting latent-conditioned actor.

\subsubsection{Network architecture}

Let $d$ denote the learned representation dimension.  Our left SUAF feature
$f(s,a)$ and forward TD predictor $h(s,a)$ use the upstream forward-network
architecture.  The right outcome feature $g(s)$ and reverse TD predictor
$v(s,a)$ use the upstream backward-network architecture.  The actor consumes
the concatenation of the observation and reward coordinate $\beta\in
\mathbb R^d$.  Two independently initialized left heads are used when
computing the actor score. The right representation is shared.

\begin{table}[H]
  \centering
  \caption{\textbf{Resolved network architecture.} A tuple lists hidden-layer widths. The output layer is not included. The twin count applies to the left
  feature and its target copy.}
  \label{tab:continuous_networks}
  \small
  \setlength{\tabcolsep}{3.5pt}
  \resizebox{\textwidth}{!}{%
  \begin{tabular}{lcccc}
    \toprule
    Hyperparameter & Walker & Cheetah & AntMaze Large & Cube Single Play \\
    \midrule
    Representation dimension $d$ & $50$ & $50$ & $128$ & $128$ \\
    Left feature $f$ / predictor $h$
      & $(1024,1024,1024)$ & $(1024,1024,1024)$
      & $(512,512,512,512)$ & $(512,512,512,512)$ \\
    Right feature $g$ / predictor $v$
      & $(256,256,256)$ & $(256,256,256)$
      & $(512,512,512,512)$ & $(512,512,512,512)$ \\
    Actor MLP
      & $(1024,1024,1024)$ & $(1024,1024,1024)$
      & $(512,512,512,512)$ & $(512,512,512,512)$ \\
    Activation & ReLU & ReLU & GELU & GELU \\
    Twin left heads & $2$ & $2$ & $2$ & $2$ \\
    Shared right head & yes & yes & yes & yes \\
    LayerNorm in $f,g,h,v$ & yes & yes & yes & yes \\
    LayerNorm in actor & no & no & no & no \\
    State-dependent actor std. & no & no & no & no \\
    Constant actor log-std. & $0$ & $0$ & $0$ & $0$ \\
    Actor final-layer init. scale & $10^{-2}$ & $10^{-2}$
      & $10^{-2}$ & $10^{-2}$ \\
    Actor tanh transformation & no & no & yes & yes \\
    Action clipping & environment bounds & environment bounds
      & $[-1,1]$ & $[-1,1]$ \\
    Image encoder & none & none & none & none \\
    \bottomrule
  \end{tabular}}
\end{table}

\subsubsection{Empirical action-centering}
\label{app:continuous_action_centering}

The offline datasets do not provide a callable behavior policy.  We therefore
construct a nonparametric empirical conditional action distribution once from
the reward-free training split.  We draw a fixed reference pool of
$N_{\rm ref}=32{,}768$ factual state--action rows.  States are standardized
coordinate-wise using training-set statistics.  For a query state $s$, let
$\mathcal N_K(s)$ contain the $K=16$ reference states with the smallest
squared Euclidean distance, and let $a_j$ be the factual action attached to
reference state $s_j$.  We define
$
 \widehat\pi_b(\mathrm da\mid s)
 =\frac1K\sum_{j\in\mathcal N_K(s)}\delta_{a_j}(\mathrm da),
$ with $K=16$
and action-center every raw left feature as
\begin{equation}
 f_{\rm c}(s,a)
 =f_{\rm raw}(s,a)
  -\frac1K\sum_{j\in\mathcal N_K(s)}f_{\rm raw}(s,a_j).
 \label{eq:continuous_f_centering}
\end{equation}

For a training query originating from episode $e$, reference rows from the
same original episode are masked before finding nearest neighbors.  This
cross-fitting rule prevents the factual action from being deterministically
included in its own baseline and avoids the finite-$K$ shrinkage caused by
reusing the factual action.  The neighbor search and the feature baseline are
recomputed for the current query state, but the reference pool and its
standardization statistics remain fixed throughout training.  No behavior
policy MLP is fitted.

Action-centering in \eqref{eq:continuous_f_centering} is used in the
feature loss, reverse-TD target, calibration covariance, whitening/SVD, and
all reported advantage errors.  During actor differentiation the subtracted
baseline is constant with respect to the proposed action at fixed $s$. Consequently it may be cached without changing the actor gradient.  We still
include it when logging numerical advantages, so the reported score is
centered and not an arbitrary state-dependent offset.

\begin{table}[H]
  \centering
  \caption{Hyperparameters of empirical action-centering, shared across the four continuous-control domains.}
  \label{tab:continuous_centering_hyperparameters}
  \small
  \begin{tabular}{ll}
    \toprule
    Hyperparameter & Value \\
    \midrule
    Behavior model & nonparametric empirical kNN \\
    Reference rows $N_{\rm ref}$ & $32{,}768$ \\
    Neighbors / action samples $K$ & $16$ \\
    State metric & squared Euclidean after coordinate standardization \\
    Neighbor weighting & uniform, $1/K$ \\
    Reference-pool sampling & uniform without replacement, fixed per seed \\
    Same-episode reference rows & excluded during representation training \\
    Factual action forcibly included & no \\
    Gradient through selected baseline actions & no \\
    \bottomrule
  \end{tabular}
\end{table}

\begin{figure}[H]
  \centering
  \includegraphics[width=\textwidth]{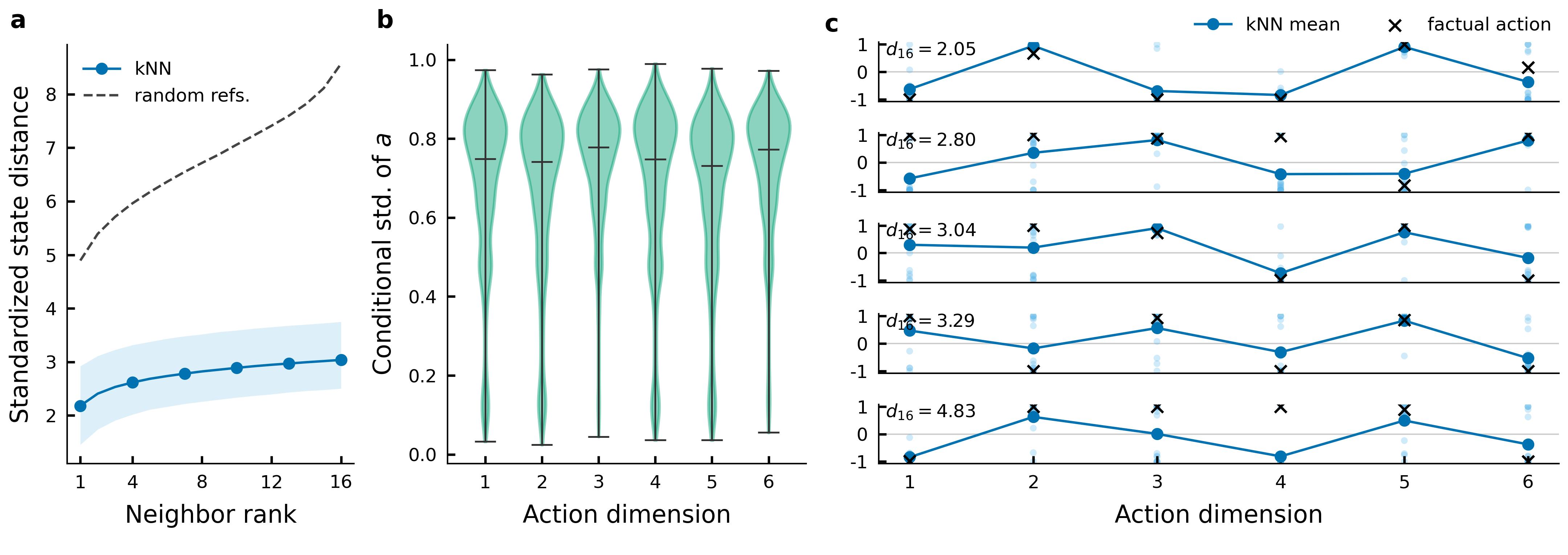}
  \caption{\textbf{Reserved diagnostic for empirical action-centering of Walker.} (a) standardized neighbor distance versus neighbor rank;
  (b) per-action-dimension conditional standard deviation; and
  (c) kNN action samples, their mean, and the factual action for representative query states.}\label{fig:continuous_centering_diagnostic_walker}
\end{figure}

\begin{figure}[H]
  \centering
  \includegraphics[width=\textwidth]{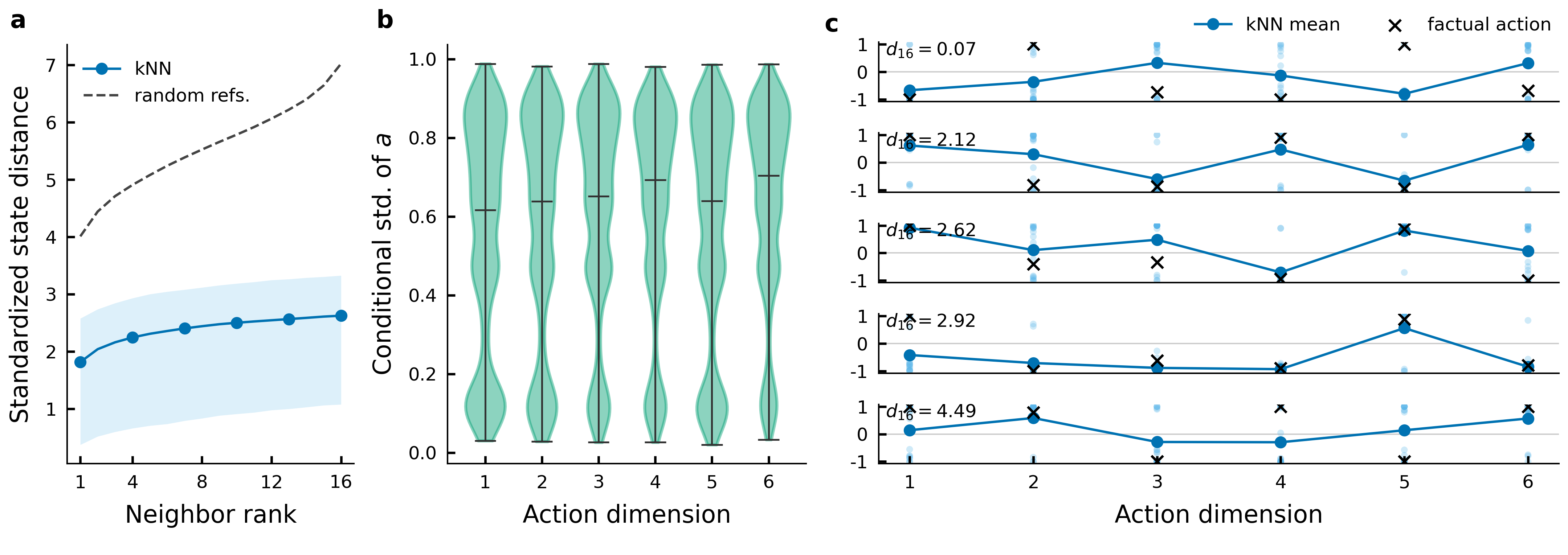}
  \caption{\textbf{Reserved diagnostic for empirical action-centering of Cheetah.} (a) standardized neighbor distance versus neighbor rank;
  (b) per-action-dimension conditional standard deviation; and
  (c) kNN action samples, their mean, and the factual action for representative query states.}\label{fig:continuous_centering_diagnostic_cheetah}
\end{figure}

\begin{figure}[H]
  \centering
  \includegraphics[width=\textwidth]{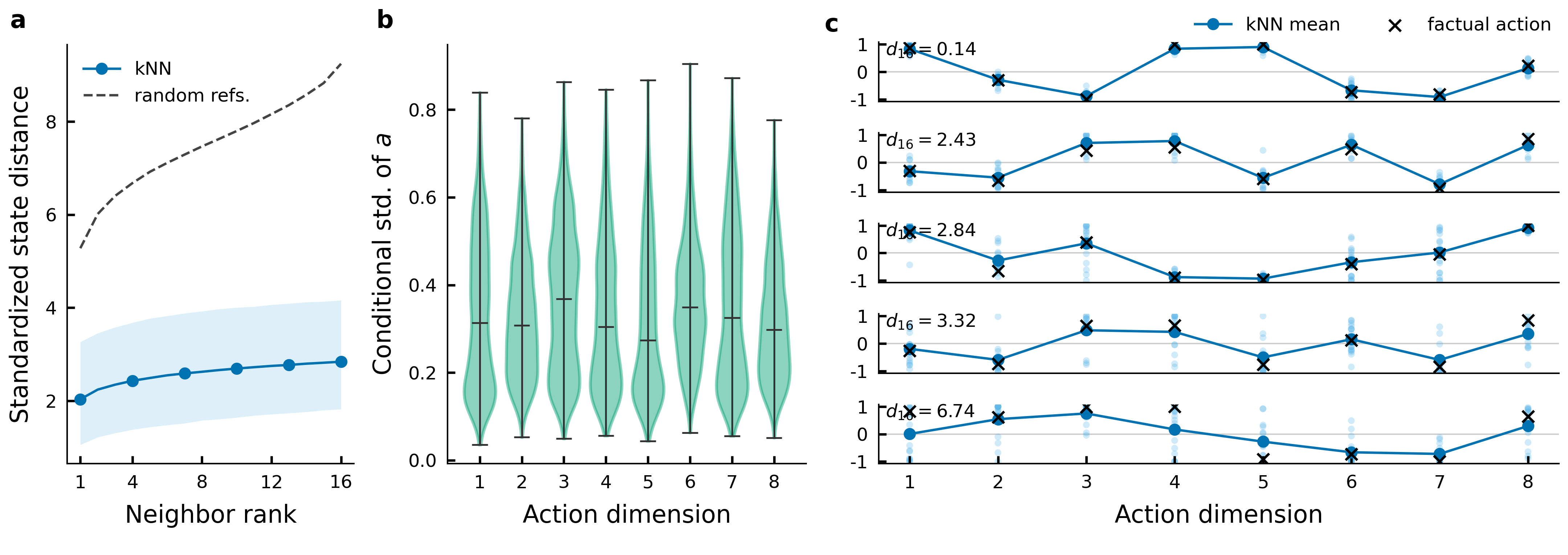}
  \caption{\textbf{Reserved diagnostic for empirical action-centering of Antmaze large.} (a) standardized neighbor distance versus neighbor rank;
  (b) per-action-dimension conditional standard deviation; and
  (c) kNN action samples, their mean, and the factual action for representative query states.}\label{fig:continuous_centering_diagnostic_antmaze_large}
\end{figure}

\begin{figure}[H]
  \centering
  \includegraphics[width=0.8\textwidth]{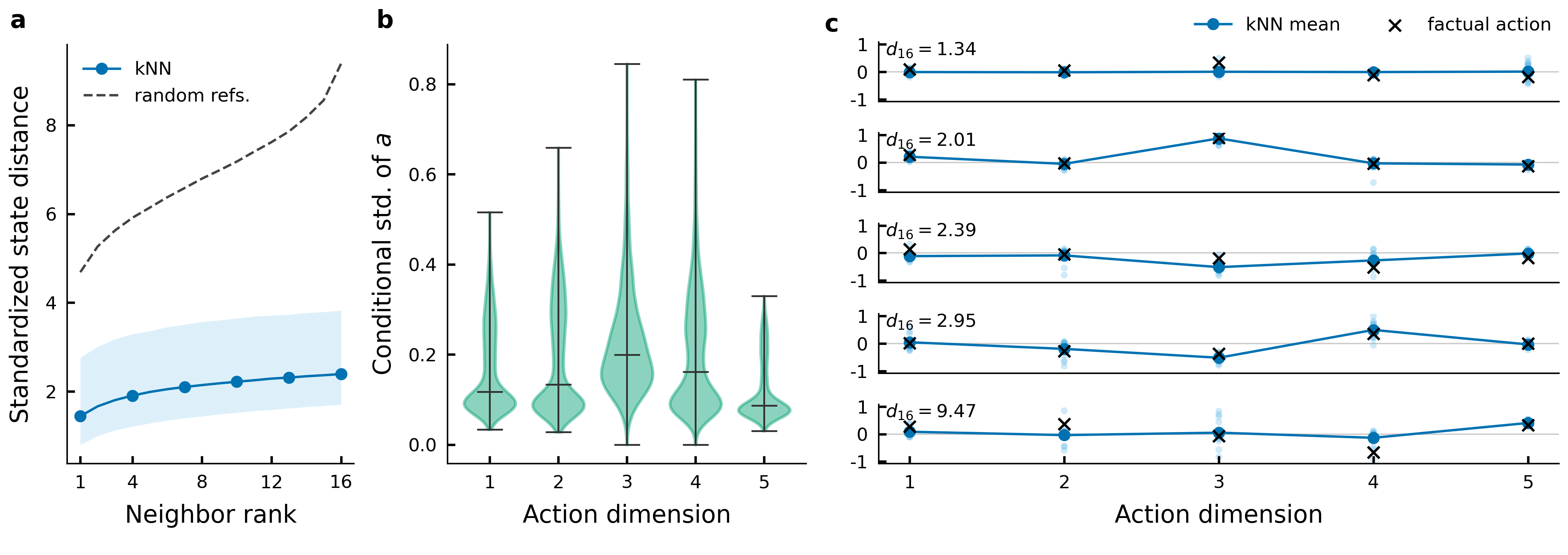}
  \caption{\textbf{Reserved diagnostic for empirical action-centering of Cube Single Play.} (a) standardized neighbor distance versus neighbor rank;
  (b) per-action-dimension conditional standard deviation; and
  (c) kNN action samples, their mean, and the factual action for representative query states.}\label{fig:continuous_centering_diagnostic_cube_single_play}
\end{figure}

\subsubsection{Representation learning and calibration}

For a positive-lag outcome $Y$ and $X=(S,A)$, the learned delayed component
uses the action-centered left feature and marginal-centered right feature.
The forward and reverse vector predictors are trained with one-step TD and
Polyak targets. After representation learning, an independent reward-free split
is used to estimate feature means and covariance matrices, whiten both sides,
and compute a $d\times d$ SVD.  The canonical modes are therefore
\begin{equation}
 \widehat f=f_{\rm c}T_f,\qquad
 \widehat g=(g_{\rm raw}-\widehat\mu_g)T_g,
 \qquad
 \widehat\Sigma=\operatorname{diag}(\widehat\sigma_1,\ldots,
 \widehat\sigma_d).
\end{equation}

\begin{table}[H]
  \centering
  \caption{\textbf{Optimization and representation hyperparameters.} Domain-specific regularization coefficients match the public One-Step FB settings.}
  \label{tab:continuous_optimization}
  \small
  \setlength{\tabcolsep}{3.1pt}
  \resizebox{\textwidth}{!}{%
  \begin{tabular}{lcccc}
    \toprule
    Hyperparameter & Walker & Cheetah & AntMaze Large & Cube Single Play \\
    \midrule
    Optimizer & Adam & Adam & Adam & Adam \\
    Adam $(\beta_1,\beta_2,\epsilon)$
      & $(0.9,0.999,10^{-8})$ & $(0.9,0.999,10^{-8})$
      & $(0.9,0.999,10^{-8})$ & $(0.9,0.999,10^{-8})$ \\
    Actor learning rate (One-Step FB aligned)
      & $10^{-4}$ & $10^{-4}$ & $10^{-4}$ & $10^{-4}$ \\
    Feature learning rate ($f,g$)
      & $10^{-5}$ & $10^{-5}$ & $10^{-5}$ & $10^{-5}$ \\
    TD-predictor learning rate ($h,v$)
      & $3\!\times\!10^{-4}$ & $3\!\times\!10^{-4}$
      & $3\!\times\!10^{-4}$ & $3\!\times\!10^{-4}$ \\
    Weight decay / AMSGrad & $0$/false & $0$/false & $0$/false & $0$/false \\
    Minibatch size & $1024$ & $1024$ & $1024$ & $1024$ \\
    Actor gradient clipping & none & none & none & none \\
    Feature/predictor gradient clipping
      & global norm $10$ & global norm $10$ & global norm $10$ & global norm $10$ \\
    Polyak target coefficient $\tau$ & $0.01$ & $0.01$ & $0.005$ & $0.005$ \\
    Representation outer iterations
      & $10{,}000$ & $10{,}000$ & $10{,}000$ & $10{,}000$ \\
    Forward-TD updates / outer iteration & $10$ & $10$ & $10$ & $10$ \\
    Reverse-TD updates / outer iteration & $10$ & $10$ & $10$ & $10$ \\
    $f$ updates / outer iteration & $1$ & $1$ & $1$ & $1$ \\
    $g$ updates / outer iteration & $1$ & $1$ & $1$ & $1$ \\
    Reverse-TD objective & \multicolumn{4}{c}{discounted boundary MSE} \\
    Nested ordering & sequential & sequential & sequential & sequential \\
    Covariance-anchor coefficient & $100$ & $100$ & $100$ & $100$ \\
    Orthonormality/covariance coefficient
      & $0.1$ & $1.0$ & $0$ & $0.3$ \\
    Forward-head aggregation for actor
      & mean & mean & min & min \\
    Latent/outcome mixing probability & $0.5$ & $0.5$ & $0.5$ & $0.5$ \\
    Normalize actor score loss by $\E|\widehat A|$ & yes & yes & yes & yes \\
    BC coefficient $\lambda_{\rm BC}$ & $0$ & $0$ & $0.03$ & $0.3$ \\
    Reward-weighting temperature $\tau_{\rm reward}$
      & $10$ & $3$ & $10$ & $300$ \\
    Positive-lag training pairs
      & $10^6$ & $10^6$ & $10^6$ & $10^6$ \\
    Diagnostic positive-lag pairs
      & $10^5$ & $10^5$ & $10^5$ & $10^5$ \\
    Calibration-fit snippets
      & $120{,}000$ & $120{,}000$ & $120{,}000$ & $120{,}000$ \\
    Calibration-selection snippets
      & $120{,}000$ & $120{,}000$ & $120{,}000$ & $120{,}000$ \\
    Calibration horizon
      & $500$ & $500$ & $512$ & $512$ \\
    Calibration minibatch size
      & $4096$ & $4096$ & $4096$ & $4096$ \\
    Calibration split & independent & independent & independent & independent \\
    Calibration arithmetic & float64 & float64 & float64 & float64 \\
    Ridge multiplier candidates
      & \multicolumn{4}{c}{$\{0,10^{-7},10^{-6},10^{-5},10^{-4},10^{-3},10^{-2},10^{-1}\}$} \\
    Shrinkage multiplier candidates
      & \multicolumn{4}{c}{$\{0,10^{-4},3\!\times\!10^{-4},10^{-3},3\!\times\!10^{-3},10^{-2},3\!\times\!10^{-2},10^{-1}\}$} \\
    \bottomrule
  \end{tabular}}
\end{table}

The covariance ridges for the two sides are selected jointly from the stated
grid by the reward-free held-out quadratic loss. The shrinkage multiplier is
then selected from its stated grid using the same held-out criterion.  Thus a
selected value may differ across seeds, but the selection rule and candidate
sets are fixed before inspecting downstream task performance.  We store the
complete sweep and selected values in each calibration artifact.

\subsubsection{Reward encoding and actor update}

During training, using a set of training rewards $(Y_i,r_i)_{i=1}^N$, reward embeddings follows,
\begin{equation}
 \beta_r=\sum_{i=1}^N r_i\widehat g(Y_i).
 \label{eq:continuous_reward_coordinate}
\end{equation}
Unlike the upstream implementation, our primary SUAF variant does not project
$\beta_r$ onto a fixed-radius sphere and does not multiply it by
$\widehat\Sigma$ before passing it to the actor.  Thus both its direction and
magnitude are available to the policy.  Singular values enter only the SUAF
advantage,
\begin{equation}
 \widehat A_r(s,a)
 =\widehat C_{\pi_b}r(s,a)
 +\frac{\gamma}{1-\gamma}
   \widehat f(s,a)^\top\widehat\Sigma\beta_r.
 \label{eq:continuous_suaf_advantage}
\end{equation}
For rewards represented as functions of the stored current state,
$\widehat C_{\pi_b}r=0$.  For genuinely action-dependent or next-state rewards,
the immediate contrast is estimated separately and retained. It is not folded
into the positive-lag factorization.

The actor is trained by reparameterized policy-gradient plus behavioral
cloning,
\begin{equation}
 \mathcal L_{\rm actor}
 =-\sg\!\left(\frac{1}{\E|\widehat A_r|}\right)
   \E\left[\widehat A_r
     \bigl(S,\pi_\phi(S,\beta_r)\bigr)\right]
 +\lambda_{\rm BC}
   \E\left[-\log\pi_\phi(A\mid S,\beta_r)\right].
 \label{eq:continuous_actor_loss}
\end{equation}
At evaluation time, each action is produced by one actor forward pass.  We do
not enumerate action candidates.  The upstream fixed-radius latent
normalization is retained as a separately labeled ablation; it is not mixed
with the raw-$\beta$ main result.



\end{document}